\documentclass[journal]{IEEEtran}
\usepackage{signalpreamble}
\usepackage{pgfplots}
\usepackage{tikz}
\usetikzlibrary{backgrounds , calc, shapes.geometric, arrows.meta, positioning}
\pgfplotsset{compat=1.18}

\usepackage{wrapfig}
\usepackage{standalone} 

\usepackage[labeled,resetlabels]{multibib}
\newcites{SR}{Supplementary References}

\usepackage{silence}
\usepackage{adjustbox}

\renewcommand{\thesection}{\docprefix\Roman{section}}
\renewcommand{\thesectiondis}{\docprefix\Roman{section}.}
\renewcommand{\theequation}{\docprefix\arabic{equation}}
\renewcommand{\theLemma}{\docprefix\arabic{Lemma}}

\renewcommand{\theTheorem}{\docprefix\arabic{Theorem}}

\renewcommand{\theAssumption}{\docprefix\arabic{Assumption}}

\renewcommand{\thetable}{\docprefix\arabic{table}}
\renewcommand{\thefigure}{\docprefix\arabic{figure}}

\def\docprefix{} 

\begin{document}


\setulcolor{orange}

\newcommand{\markc}{\brown{$\diamond$ }}

\newcommand{\Pdata}{\P_{\mathrm{data}}}
\newcommand{\Pret}{\P_r}
\newcommand{\Pfor}{\P_f}
\newcommand{\pdata}{p_{\mathrm{data}}}
\newcommand{\pret}{p_r}
\newcommand{\pfor}{p_f}

\newcommand{\trainset}{{\calD_{\mathrm{train}}}}
\newcommand{\calibset}{{\calD_{\mathrm{calib}}}}
\newcommand{\calibsize}{{\abs{\calibset}}}
\newcommand{\testset}{{\calD_{\mathrm{test}}}}
\newcommand{\ulset}{{\calD_{\mathrm{unlearn}}}}
\newcommand{\forgetset}{{\calD_f}}
\newcommand{\retainset}{{\calD_r}}
\newcommand{\Wforget}{\calW_{\mathrm{forget}}}
\newcommand{\Wretain}{\calW\backslash\calW_{\mathrm{forget}}}

\newcommand{\Xtest}{{X_{\mathrm{test}}}}
\newcommand{\Ytest}{{Y_{\mathrm{test}}}}
\newcommand{\Ztest}{{Z_{\mathrm{test}}}}
\newcommand{\CP}{\mathfrak{CP}}
\newcommand{\CR}{\mathsf{CR}}
\newcommand{\ncal}{n_{\mathrm{cal}}}
\newcommand{\nulset}{n_{\mathrm{unlearn}}}

\newcommand{\Ualgo}{\frakU}
\newcommand{\loss}{\mathsf{L}}
\newcommand{\losseff}{\ell_{\mathrm{eff}}}
\newcommand{\hqalpha}{\hat{q}_{\alpha}}

\NewDocumentCommand{\Leff}{s o}{%
        \IfBooleanTF{#1}
        {\widehat{L}_{\mathrm{eff}\IfValueT{#2}{,#2}}} 
        {L_{\mathrm{eff}\IfValueT{#2}{,#2}}} 
}

\NewDocumentCommand{\Lcov}{s o}{%
        \IfBooleanTF{#1}
        {\widehat{L}_{\mathrm{cov}\IfValueT{#2}{,#2}}} 
        {L_{\mathrm{cov}\IfValueT{#2}{,#2}}} 
}

\DeclareDocumentCommand \ECF { e{_} o } {%
        \mathsf{C}%
        \IfValueT{#1}{_{#1}}%
        \IfValueT{#2}{\parens*{#2}}%
}

\DeclareDocumentCommand \EmCF { e{_} o } {%
        \mathsf{M}%
        \IfValueT{#1}{_{#1}}%
        \IfValueT{#2}{\parens*{#2}}%
}

\DeclareDocumentCommand \H { e{_} o } {%
        \frakH%
        \IfValueT{#1}{_{#1}}%
        \IfValueT{#2}{\parens*{#2}}%
}

\newacronym{MU}{MU}{machine unlearning}
\newacronym{CP}{CP}{conformal prediction}
\newacronym{RT}{RT}{retrained}
\newacronym{cdf}{cdf}{cumulative distribution function}
\newacronym{pdf}{pdf}{probability density function}



\title{Conformal Unlearning: A New Paradigm for Unlearning in Conformal Predictors}

\author{Yahya Alkhatib, Muhammad Ahmar Jamal, and Wee Peng Tay,~\IEEEmembership{Senior Member,~IEEE}
\thanks{The authors are with the School of Electrical and Electronic Engineering, Nanyang Technological University, Singapore. Emails: \texttt{\{ya0001ib@e.ntu.edu.sg, JAMA0002@e.ntu.edu.sg, wptay@ntu.edu.sg\}}}
}
\markboth{}%
{}

\IEEEpubid{}

\maketitle

\begin{abstract}
Conformal unlearning aims to ensure that a trained conformal predictor miscovers data points with specific shared characteristics, such as those from a particular label class, associated with a specific user, or belonging to a defined cluster, while maintaining valid coverage on the remaining data. Existing machine unlearning methods, which typically approximate a model retrained from scratch after removing the data to be forgotten, face significant challenges when applied to conformal unlearning. These methods often lack rigorous, uncertainty-aware statistical measures to evaluate unlearning effectiveness and exhibit a mismatch between their degraded performance on forgotten data and the frequency with which that data are still correctly covered by conformal predictors—a phenomenon we term ``fake conformal unlearning.'' To address these limitations, we propose a new paradigm for conformal machine unlearning that provides finite-sample, uncertainty-aware guarantees on unlearning performance without relying on a retrained model as a reference. We formalize conformal unlearning to require high coverage on retained data and high miscoverage on forgotten data, introduce practical empirical metrics for evaluation, and present an algorithm that optimizes these conformal objectives. Extensive experiments on vision and text benchmarks demonstrate that the proposed approach effectively removes targeted information while preserving utility.
\end{abstract}


\glsresetall{}
\section{Introduction}\label{sec:intro}

\IEEEPARstart{A} robust framework for integrating uncertainty quantification into machine learning models is \gls{CP} \cite{vovk2005alrw, angelopoulos2022gentle, angelopoulos2025theoretical}. By providing rigorous guarantees on validity—ensuring that the true label of a test point is included in the prediction set with a user-specified probability—\gls{CP} offers a distribution-free approach that relies on minimal assumptions, such as data exchangeability. However, the increasing deployment of machine learning systems in dynamic environments has underscored the need for models to adapt to evolving knowledge and semantics.

In practice, entire categories of data may become obsolete, redefined, or classified as sensitive. For instance, in safety-critical applications such as content moderation, classification taxonomies are frequently updated to reflect newly restricted or redefined content categories. Similarly, in domains like e-commerce, recommendation systems, and inventory management, product categories may become obsolete or discontinued. These scenarios necessitate the ability to efficiently remove the influence of outdated or irrelevant data from trained models, a challenge addressed by the field of \emph{\gls{MU}} \cite{cao2015towards, bourtoule2021machine, warnecke2023machine}. Machine unlearning aims to eliminate the impact of specific data points from a model without the computational overhead of retraining from scratch, thereby ensuring model reliability and compliance with evolving requirements \cite{cao2015towards}. The need for both uncertainty quantification and selective forgetting is particularly pronounced in sensitive domains such as medical diagnosis \cite{lu2022fair, lambrou2011reliable}, where models must provide calibrated confidence while adapting to changes in clinical guidelines, patient data removal requests, or outdated treatment protocols. Despite these needs, the integration of unlearning capabilities into conformal predictors—enabling them to forget specific data groups while preserving valid coverage guarantees—remains an open research problem.

This work introduces the concept of \emph{conformal unlearning}, which extends traditional machine unlearning by targeting high miscoverage of the forgotten data within \gls{CP} sets. Unlike conventional unlearning, which focuses on point-estimate predictions, conformal unlearning aims to ensure that the prediction sets of the unlearned model exhibit high \emph{miscoverage} on the forget data, i.e., the true label is excluded from the prediction set with high probability, while maintaining valid \emph{coverage} on the retained data. This approach shifts the unlearning objective from point-estimate accuracy to a probabilistic framework, where a successfully unlearned model is characterized by high uncertainty about the forgotten data and high confidence about the retained data.

Traditional machine unlearning methods, which aim to approximate a \gls{RT} model, can be extended to conformal unlearning by first applying the unlearning procedure and subsequently employing \gls{CP} on the resulting model.
Existing machine unlearning techniques can be broadly classified into three categories:  
(i) Data-structure-based approaches that facilitate efficient partial retraining through specialized indexing or partitioning mechanisms \cite{ginart2019making, bourtoule2021machine};  
(ii) Gradient-influence and variational methods that adjust model parameters to negate the impact of the forgotten data \cite{warnecke2023machine, graves2021amnesiac, nguyen2020variational}; and  
(iii) Knowledge-transfer and noise-based strategies designed to eliminate sensitive information while maintaining model utility \cite{chundawat2023bad, chundawat2023zeroshot, foster2024information}. 
Certified unlearning approaches often utilize differential privacy (DP) \cite{dwork2014differential} or influence-function theory to provide formal guarantees on the extent of residual information leakage \cite{guo2020certified, sekhari2021remember, neel2021descent, koh2017understanding}. However, these methods predominantly focus on model parameters or point predictions, without addressing the behavior of prediction sets. This limitation highlights the necessity for a novel framework that explicitly targets the coverage properties of conformal predictors, as introduced in this work.

On the other hand, conformal unlearning does not extend to scenarios such as random or specific instance forgetting, where the forget data constitute a random subset of the training data and lack shared defining characteristics. In such cases, approximating a \gls{RT} model is a suitable strategy, as the forget data are inherently linked to the training set. Conversely, in conformal unlearning, the forget data are characterized by distinct, shared attributes that may not be fully represented within the training data. This distinction necessitates a fundamentally different methodological framework and evaluation criteria tailored to the unique objectives of conformal unlearning.

\begin{figure}[!tb]
\centering
\begin{subfigure}{0.5\textwidth}
\centering
\footnotesize
\begin{tikzpicture}

\begin{axis}[
width=0.85\linewidth, height=0.25\linewidth,
scale only axis,
grid=both,
xlabel={\# forgotten clusters},
ylabel={Percentage \%},
xmin=0, xmax=11,
ymin=45, ymax=60,
xtick={1,2,3,5,7,8,10},
tick label style={/pgf/number format/fixed},
clip=false,
legend to name=legend, 
legend columns=2,
legend style={font=\tiny, /tikz/every even column/.append style={column sep=0.25cm}},
legend cell align=left
]

\addplot+[
blue, thick, solid, mark=*, mark options={solid},
error bars/.cd, y dir=both, y explicit
] table[row sep=\\, x=a, y=mean] {
a   mean  \\
1   46.82\\
2   51.14\\
3   55.80\\
5   57.59\\
7   56.36\\
8   56.67\\
10  50.17\\
};
\addlegendentry{RT acc. on $\calT_f$}


\addlegendimage{magenta, thick, densely dashed, mark=pentagon*, mark options={solid}}
\addlegendentry{RT coverage on $\calT_f$}

\addlegendimage{green!50!black, thick, dotted, mark=triangle*, mark options={solid}}
\addlegendentry{Original acc. on $\calT_f$}

\addlegendimage{cyan!60!black, thick, dashdotdotted, mark=diamond*, mark options={solid}}
\addlegendentry{Original coverage on $\calT_f$}

\end{axis}

\begin{axis}[
width=0.85\linewidth, height=0.1\linewidth,
at={(0,0.30\linewidth)}, 
anchor=south west,
scale only axis,
grid=both,
xmin=0, xmax=11,
ymin=90, ymax=100,
ytick={90,95,100}, 
xtick=\empty,
clip=false
]

\addplot+[
cyan!60!black, thick, dashdotdotted, mark=diamond*, mark options={solid},
error bars/.cd, y dir=both, y explicit
] table[row sep=\\, x=a, y=mean] {
a   mean  \\
1   93\\
2   96\\
3   95\\
5   95\\
7   95\\
8   95\\
10  95\\
};

\addplot+[
green!50!black, thick, dotted, mark=triangle*, mark options={solid},
error bars/.cd, y dir=both, y explicit
] table[row sep=\\, x=a, y=mean] {
a   mean  \\
1   100\\
2   100\\
3   100\\
5   100\\
7   100\\
8   100\\
10  99.95\\
};

\addplot+[
magenta, thick, densely dashed, mark=pentagon*, mark options={solid},
error bars/.cd, y dir=both, y explicit
] table[row sep=\\, x=a, y=mean] {
a   mean  \\
1   95\\
2   93\\
3   95\\
5   95\\
7   94\\
8   95\\
10  94\\
};

\end{axis}

\node[anchor=south, yshift=1.0ex, xshift=2.5ex] at (current bounding box.north)
{\pgfplotslegendfromname{legend}};
\end{tikzpicture}
\end{subfigure}
\caption{Comparison of model accuracies (acc.) on the train forget data ($\calT_f$), and coverage over the forget data for the RT and original models using CIFAR100. 
}
\label{fig:fake_conformal_unlearning}
\end{figure}

\begin{table}[!tb]
\caption{Validation forgotten/retained accuracy for certified gradient clipping (PABI) and RT when forgetting 5 clusters in ImageNet100 (see \cref{sec:exp}). All values are in \%.}
\centering
\begin{tabular}{@{}l*{2}{c}@{}}
\toprule
Method & $\text{Acc}_{\text{val,for}}$ & $\text{Acc}_{\text{val,ret}}$ \\
\midrule
Certified grad. clip. (PABI) & 31.08 & 67.96 \\
Retraining (RT)              & 34.79 & 71.25 \\
\bottomrule
\end{tabular}
\label{tab:pabi-to-RT-accuracy-deviation}
\end{table}



\begin{figure}[!tb]
\centering
\begin{subfigure}[t]{0.13\textwidth}
\centering
\includegraphics[width=\linewidth]{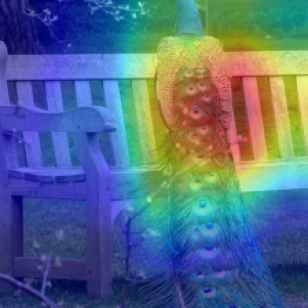}
\caption{PABI overlay}
\label{fig:certified_overlay}
\end{subfigure}\hfill
\begin{subfigure}[t]{0.13\textwidth}
\centering
\includegraphics[width=\linewidth]{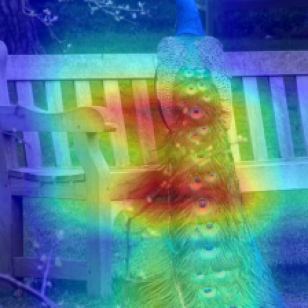}
\caption{RT overlay}
\label{fig:rt_overlay}
\end{subfigure}\hfill
\begin{subfigure}[t]{0.13\textwidth}
\centering
\includegraphics[width=\linewidth]{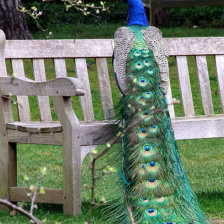}
\caption{Original image}
\label{fig:orig_image}
\end{subfigure}
\caption{Comparison of Grad-CAM overlay regions on a peacock image after cluster-wise forgetting. Certified unlearning PABI generates a saliency map considerably different to that of RT, despite certified indistinguishability.}
\label{fig:rt_pabi_cam_compare}
\end{figure}


Conventional metrics, such as accuracy on forgotten, retained, or test splits, primarily evaluate point-estimate predictions and fail to account for uncertainty. In the context of unlearning, particularly in scenarios like cluster-wise forgetting, these metrics often yield ambiguous results due to generalization effects. Specifically, the impact of unlearning on accuracy for forgotten or retained data can be inconsistent, leaving practitioners uncertain about whether genuine unlearning has been achieved. Moreover, these metrics are inherently inadequate for quantifying uncertainty, which is a critical aspect of conformal unlearning.

\Cref{fig:fake_conformal_unlearning} demonstrates that even when retraining leads to a significant reduction in accuracy on the training forget data, \gls{CP} at a significance level of $\alpha=0.05$ still achieves high coverage. This occurs despite the average size of the \gls{CP} sets being relatively small—approximately 20 out of 100 labels. This discrepancy underscores a critical issue: while accuracy on forget data may decline substantially, coverage remains largely unaffected. In other words, although the model's top-1 predictions are frequently incorrect on forget data (low accuracy), the true label often remains within the \gls{CP} sets (high coverage), indicating that the model retains substantial knowledge about the forgotten data.

This phenomenon, termed \emph{fake conformal unlearning}, persists even when accuracy drops are pronounced (exceeding $40\%$ across various numbers of forgotten clusters). It highlights a fundamental disconnect between accuracy-based metrics and the actual forgetting of conformal predictors \cite{cooper2025generativeaipolicy, shi2025rethinking}. This disconnect underscores the necessity for statistically rigorous metrics that explicitly capture uncertainty in the context of unlearning. Such metrics are indispensable for ensuring that machine unlearning is both interpretable and actionable in practice. While prior work, such as \cite{shi2025rethinking}, has made initial efforts to bridge traditional machine unlearning with \gls{CP}, their approach does not fully address the unique challenges posed by conformal unlearning as defined in this work (cf.\  \cref{sec:related-work}).

Next, we discuss in detail two critical limitations faced by traditional \gls{MU} that relies on approximating a \gls{RT} model, in the context of conformal unlearning:
\begin{enumerate}[label=\textbf{I\arabic*}]
\item \textbf{Dependence on \gls{RT} baselines for utility assessment}: 
Conventional \gls{MU} guarantees define unlearning as producing parameters indistinguishable from a model retrained without the forget data \cite{sekhari2021remember,guo2020certified,koloskova2025certifiednonconvex}. However, this parameter-space focus obscures empirical evaluation: models that are parameter-close to a \gls{RT} baseline may still exhibit divergent behavior on validation or held-out data. For instance, \Cref{tab:pabi-to-RT-accuracy-deviation} shows accuracy deviations between the \gls{RT} model and PABI \cite{koloskova2025certifiednonconvex}, an $(\epsilon, \delta)$-certified indistinguishable method, on validation data. While these deviations are bounded by $(\epsilon, \delta)$, they remain impractical to assess without referencing the \gls{RT} model. Similarly, \cref{fig:rt_pabi_cam_compare} illustrates perceptible differences in Grad-CAM overlays despite certified indistinguishability. Such reliance on costly retrained baselines for utility validation is infeasible at scale \cite{ginart2019making,warnecke2023machine,neel2021descent}. This underscores the need for conformal unlearning frameworks that (i) align with practitioner-specific forgetting objectives and (ii) enable transparent, model-agnostic evaluations without requiring expensive baselines \cite{thudi2022auditable}.

\item \textbf{Forgeability in parameter-space unlearning}: Parameter-space certification, as critiqued in \cite{thudi2022auditable}, is susceptible to forgeability, where indistinguishable or identical parameter vectors can arise from different training datasets. This undermines the validity of unlearning definitions based solely on parameter similarity to a \gls{RT} model. Certified approaches aiming for $(\epsilon,\delta)$-unlearning \cite{mu2025certifiedrewind,basaran2025certifiedsourcedata,chien2024certifiednoisySGD,koloskova2025certifiednonconvex,guo2020certified,ginart2019making} are particularly prone to this issue, as indistinguishability in parameter space does not guarantee behavioral consistency. This highlights the necessity of unlearning definitions that prioritize observable model behavior over proximity to a baseline in parameter space.
\end{enumerate}

These limitations arise from framing unlearning in terms of parameter-space proximity to a \gls{RT} model, rather than focusing on observable model behavior. While \gls{RT} models provide a useful reference for specific instance forgetting, their applicability to conformal unlearning is limited by these challenges.

\gls{CP} offers a natural resolution: by defining unlearning through coverage and miscoverage probabilities—which are directly observable and statistically quantifiable—we obtain objectives that are (i) independent of any baseline model, (ii) immune to forgeability concerns since they characterize prediction-set behavior rather than parameter values, and (iii) equipped with finite-sample guarantees under mild assumptions. Our main contributions are summarized as follows:
\begin{itemize}[leftmargin=10pt,topsep=0pt,partopsep=0pt,parsep=0pt,itemsep=0pt]
\item We propose a conformal, probabilistic definition of unlearning that quantifies conformal forgetting directly, without reference to any retrain-from-scratch baseline.
\item We introduce practical empirical metrics—Empirical Coverage Frequency (\emph{ECF}) at threshold $c$ and Empirical misCoverage Frequency (\emph{EmCF}) at threshold $d$—for evaluating uncertainty-aware unlearning. ECF measures the fraction of data points whose true label is covered by the prediction set of size at most $c$, while EmCF measures the fraction excluded—providing direct empirical counterparts to the theoretical coverage and miscoverage guarantees.
\item We develop a scalable unlearning algorithm that outputs an unlearned \gls{CP} set,  achieving strong forgetting of targeted data while preserving coverage on retained data.
\end{itemize}

The rest of this paper is organized as follows. \Cref{sec:preliminary} provides the necessary preliminaries and notation. In \cref{sec:theory}, we formalize the conformal unlearning framework, including its definitions, theoretical guarantees, and empirical metrics. \Cref{sec:method} details our proposed empirical method for implementing conformal unlearning. \Cref{sec:exp} presents the experimental setup and results, showcasing the effectiveness of our approach. \Cref{sec:different-scores} discusses the impact of using different conformity score functions during inference.  \Cref{sec:non-exch} extends the discussion to conformal unlearning beyond exchangeability. \Cref{sec:related-work} reviews related work in machine unlearning and highlights the distinctions of our framework. Finally, \Cref{sec:conclusion} summarizes the contributions and outlines future directions.

\section{Preliminaries and Notations}\label{sec:preliminary} 

Let $X\in\calX$ denote features and $Y\in\calY$ denote a label or response corresponding to $X$. We denote a dataset $\calD\sim p$ if $\calD$ consists of data points $(x,y)\in \calX\times\calY$ generated \gls{iid} from $p$. 

For a given model $f_{\theta_o}$, a machine learning model trained on the training set $\trainset \sim \pdata$ produces a model $f_{\theta_o}$ with parameters $\theta_o$. Let $\Pdata$ denote the probability measure corresponding to $\pdata$. We focus on \emph{conformal unlearning}: Let a target variable $W\in\calW$ encode the characteristics to be forgotten. Our objective is to forget the influence of data that are generated conditioned on $W\in\calW_{\mathrm{forget}}$, where $\calW_{\mathrm{forget}}\subset \calW$ defines the forget criteria. Let $\pi_f = \P(W \in \calW_{\mathrm{forget}}) = 1 - \pi_r$ so that \begin{align}\label{eq:pdata}
    \pdata = \pi_r \pret + \pi_f \pfor,    
\end{align}
where the retain and forget sets are drawn from the conditional \glspl{pdf}:
\begin{align}
\begin{aligned}\label{def:pret-pfor}
&\retainset \sim \pret(\cdot,\cdot) = p_{X,Y\mid W}(\cdot,\cdot\mid W\in\calW\backslash\calW_{\mathrm{forget}}), \\
&\forgetset \sim \pfor(\cdot,\cdot) = p_{X,Y\mid W}(\cdot,\cdot\mid W\in\calW_{\mathrm{forget}}),
\end{aligned}
\end{align}
respectively. We set $\ulset=\retainset\cup\forgetset \sim \pdata$.  Note that we do not require that $\ulset \subset \trainset$ although this is the typical case in practice. 

An important assumption used throughout this work is that $\pfor \ne \pret$, i.e., the distributions are distinguishable.
 
Examples of conformal unlearning include class-level forgetting ($W=Y$) and feature-, subspace-, or semantic-based criteria ($W=X$, $W=\Pi X$, or $W=h(X)$, where $\Pi$ is a subspace projector and $h$ is a feature transformation map). 

In the traditional \gls{MU} literature, a \gls{MU} algorithm $\Ualgo$ transforms the model parameters $\theta_o$ into $\theta_u$ by utilizing $(\retainset, \forgetset)$, and possibly other information so that the unlearned model $f_{\theta_u}$ approximates the \gls{RT} model. In conformal unlearning, as defined in \cref{sec:theory}, we depart from this perspective and instead focus on directly quantifying the forgetting and retaining performance of $f_{\theta_u}$ on $\forgetset$ and $\retainset$, respectively, without reference to the \gls{RT} model. Instead, conformal unlearning aims to ensure that the prediction sets produced by $f_{\theta_u}$ exhibit high miscoverage on $\forgetset$ while maintaining valid coverage on $\retainset$.

By letting $W=(X,Y)$ and setting $\Wforget$ to correspond to a specific subset of training data, we recover the standard specific instance forgetting setup \cite{cao2015towards, ginart2019making, warnecke2023machine, sekhari2021remember, guo2020certified}. However, specific instance forgetting does not align with our intended focus on conformal unlearning, which leverages shared characteristics to define the forget set. In many applications, practitioners seek to remove data based on common features or labels, e.g., removing all data from a deprecated class or associated with a particular user, and in the case of isolated instances, unlearning is vacuous since the model is generalizable leading to fake conformal unlearning. In addition, to evaluate the performance of specific instance forgetting, random forgetting \cite{foster2024fast, foster2024information, shi2025rethinking, peng2025adversarial, graves2021amnesiac} is often employed, where $\forgetset$ is a random sub-sample of $\trainset$, which violates the spirit of conformal unlearning. Therefore, in this work, we do not perform any random forgetting experiments.

Below, we provide a concise overview of split \gls{CP} \cite{angelopoulos2022gentle, shafer2008conformal, barber2023conformal}, along with miscoverage and efficiency losses, which serve as the foundation of our conformal unlearning framework. A high-level summary of the conformal unlearning framework is also presented, with detailed definitions and theoretical insights deferred to \cref{sec:theory}.

\paragraph{Split \gls{CP}}

The split \gls{CP} framework provides distribution-free prediction sets with guaranteed coverage. For a trained model $f_{\theta}$, a nonconformity score is defined as:
\begin{align}
s(X,Y;\theta) = \loss \parens*{f_{\theta}(X),Y},
\end{align}
where $\loss$ is a loss function, and smaller values indicate better conformity. In classification tasks, a common choice is \cite{sadinle2019least}:
\begin{align}\label{eq:non-conformity}
s(X,Y;\theta) \triangleq 1 - p_{\theta}(Y\mid X),
\end{align}
where $p_{\theta}$ represents the softmax probability output of the model $f_{\theta}$ \cite{sadinle2019least}. For regression problems, alternative score functions are typically used \cite{angelopoulos2022gentle}. 

Given an unseen test point $(X,Y)$ and a threshold $t \in \bbR$, a prediction set is constructed as follows: 
\begin{align}\label{predictionset}
\calC_{\theta, t} (X) = \set*{y \given s(X,y;\theta) \le t},
\end{align}
where the threshold $t$ determines the size of the prediction set.

In split \gls{CP}, the dataset is divided into a training set $\trainset$ for training the model $f_{\theta}$, a calibration set $\calibset$ of size $n$, and a test set $\testset$. Given a significance level $\alpha$, let $\hqalpha$ represent the $\ceil{(1-\alpha)(n+1)}/n$ quantile of the scores computed from $\calibset$. By setting $t = \hqalpha$, and assuming that $\calibset$ and $\testset$ are exchangeable \cite{angelopoulos2022gentle, shafer2008conformal, barber2023conformal}, the following \emph{coverage} guarantee is achieved for a test point $(X,Y)\in\testset$: 
\begin{align}\label{eq:quantile-coverage}
\P(Y \in \calC_{\theta, \hqalpha} (X)) = \P(s(X,Y;\theta) \le \hqalpha) \ge 1-\alpha,
\end{align}
where the probability accounts for the randomness in both the calibration set and the test point. In practice, the dataset is randomly partitioned into $\trainset$, $\calibset$, and $\testset$ to ensure exchangeability. The quantile-based \gls{CP} procedure described above with $t=\hqalpha$ is denoted as $\CP(\theta, s)$, which outputs the prediction set $\calC_{\theta} \triangleq \calC_{\theta, \hqalpha}$.

\paragraph{Miscoverage and Efficiency Losses}
A prediction set is a set-valued map $\calC: \calX \to 2^{\calY}$, where the coverage quantifies how often the true label $Y$ is included in $\calC$ for a given input $X$. To ensure reliable predictions, we aim to bound the error in coverage, referred to as \emph{miscoverage}, within a user-specified tolerance. For a prediction set $\calC(X)$, we define the miscoverage loss and the \emph{efficiency} loss as follows:
\begin{align}
\Lcov(\calC) \triangleq \P(Y \notin \calC(X)), 
\ \Leff(\calC) \triangleq \E[\losseff(\calC(X))],
\end{align}
where $\losseff$ quantifies the efficiency of the prediction set. In classification, the efficiency loss may correspond to the size (cardinality) of the prediction set, while in regression, it could represent the length of the prediction interval (e.g., its Lebesgue measure). In other contexts, it may measure size, volume, or hyper-volume. Intuitively, as the efficiency loss increases (i.e., larger prediction sets), the miscoverage decreases or remains unchanged, since larger sets are more likely to include the true label \cite{bai2022efficient}. We assume that $\losseff$ is non-decreasing with respect to set inclusion, i.e., if $\calC_1(X) \subseteq \calC_2(X)$, then $\losseff(\calC_1(X)) \le \losseff(\calC_2(X))$.

\paragraph{Overview of the Conformal Unlearning Framework}

\begin{figure}[!htb]
\centering
\includestandalone[
width=0.5\textwidth,
mode=buildnew      
]{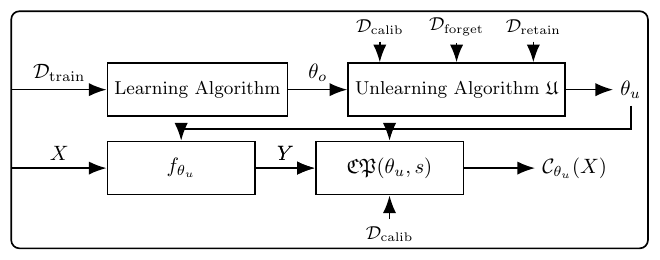}
\caption{The conformal unlearning framework.}
\label{fig:conformal-unlearning-flow-diagram}
\vspace{-5mm}
\end{figure}

Our conformal unlearning framework builds on split \gls{CP} to enable uncertainty-aware unlearning. Starting with a pretrained model $f_{\theta_o}$, we apply a \gls{MU} algorithm $\widebar{\Ualgo}$ to derive unlearned parameters $\theta_u = \widebar{\Ualgo}(\theta_o, \retainset, \forgetset)$. Using a calibration set $\calibset$, split \gls{CP} is then employed to construct prediction sets $\calC_{\theta_{u}, \hat{t}} (X)$ for points in $\ulset$, where $\hat{t}$ is a threshold found via an optimization formulation (detailed in \cref{sec:method}). The formal procedure is illustrated in \cref{fig:conformal-unlearning-flow-diagram} for the case $\hat{t}=\hqalpha$, while \cref{sec:method} discusses an empirical risk minimization approach to approximate this.

Importantly, the guarantee in \cref{eq:quantile-coverage} holds for any $\theta$, as it is induced by the calibration procedure rather than the specific model parameters. Our \gls{MU} objective (detailed in \cref{sec:theory}) leverages this property by promoting minimal coverage on $\forgetset$ (to achieve unlearning) while ensuring high, user-specified coverage on $\retainset$ (to preserve utility). We assume that, conditioned on $\trainset$, the sets $\calibset$ and $\ulset$ are exchangeable since testing is done on $\ulset$. 

This framework uses \gls{CP} both as an evaluation tool and as a training signal. Specifically, it ensures that the conformal set $\calC_{\theta_{u}, \hat{t}} (X)$ \emph{rarely} covers points from $\forgetset$, while maintaining the desired coverage on $\retainset$. Formal definitions are provided in \cref{sec:theory}.

\section{Conformal Machine Unlearning}\label{sec:theory}
In this section, we introduce the new notion of conformal unlearning for conformal predictors. We formalize the conditions under which conformal unlearning provides statistical guarantees. We propose empirical metrics to assess the performance of \gls{MU} algorithms, and present a practical conformal unlearning framework informed by our theoretical analysis.

\subsection{Definitions and Theory}

Consider a machine learning model $f_{\theta_o}$ and a quantile-based \gls{CP} procedure $\CP(\theta, s)$ as defined by \cref{predictionset}, where the probability measure in \cref{eq:quantile-coverage} is $\P = \Pdata$. For convenience, we use the notation $\P$ in place of $\Pdata$ throughout the rest of this paper whenever there is no confusion. We say that an unlearning algorithm $\Ualgo$ achieves \emph{conformal unlearning} for $\CP(\theta, s)$ if the unlearned model $f_{\theta_u}$ with parameters $\theta_u$ produces prediction sets that achieve high miscoverage on the forget set $\forgetset$ while maintaining valid coverage on the retain set $\retainset$ under $\CP(\theta_u, s)$.

Let $\Pret$ denote the probability measure under which $(X,Y) \sim \pret$ and $\calibset \sim \pdata$. Similarly, let $\Pfor$ denote the probability measure under which $(X,Y) \sim \pfor$ and $\calibset \sim \pdata$. We now present the formal definition of conformal unlearning.

\begin{Definition}[($\alpha$,$\beta$)-conformal unlearning]\label{def:conformalMU}
For $0 \leq \alpha \leq \beta \leq 1$, an unlearning algorithm $\Ualgo$ is said to be ($\alpha$,$\beta$)-conformal unlearning for a CP procedure $\CP(\theta, s)$ if the prediction sets $\calC_{\theta_{u}} (X)$ produced by $\CP(\theta_u, s)$ satisfy:
\begin{align}
\Pret(Y \in \calC_{\theta_{u}} (X)) & \geq 1 - \alpha, \label{eq:conformal-coverage-retain}\\
\Pfor(Y \notin \calC_{\theta_{u}} (X)) & \geq \beta. \label{eq:conformal-coverage-forget}
\end{align}
\end{Definition}

\Cref{def:conformalMU} places uncertainty sets at the core of unlearning, moving beyond point-estimate objectives such as misclassification \cite{graves2021amnesiac,hsu2025unseenresidualknowledge}. Intuitively, a high miscoverage indicates that the unlearned model demonstrates significant uncertainty about $\forgetset$ while maintaining confidence on $\retainset$. In sensitive applications, such as unlearning confidential or regulated information, achieving a large $\beta$ is crucial to minimize the risk of information leakage. In less critical contexts, a lower $\beta$ may suffice; for instance, if the goal is to forget a broad category, occasional inclusion in a prediction set may not be consequential. The coverage and miscoverage objectives in \cref{eq:conformal-coverage-retain,eq:conformal-coverage-forget} are independent of the underlying data distribution and the model's parameter space. Consequently, if the original model prior to unlearning already achieves uniformly high miscoverage rates on the forget data, no unlearning is required, as the model already exhibits significant uncertainty on that subset. In other words, without restricting the unlearning target to a specific parameter set, any model that satisfies the coverage and miscoverage objectives is deemed successful in unlearning the specified forget set. This ensures that forgeability \cite{thudi2022auditable} is \emph{not} a concern under conformal unlearning.

In \cref{def:conformalMU}, $\alpha$ represents the user-specified miscoverage rate inherent to \gls{CP}, which varies depending on the application. Once this tolerance level is defined, the unlearner's primary goal is to maximize the miscoverage on the left hand side of \cref{eq:conformal-coverage-forget}. We propose to do this by first \emph{parametrizing} the \gls{CP} procedure and then optimizing the unlearning algorithm $\Ualgo$ to maximize the miscoverage on $\forgetset$ and coverage on $\retainset$, subject to tradeoff constraints. A reconformalization step is finally performed. More details are provided in \cref{sec:method}.

To analyze the properties of conformal unlearning, we make the following fundamental assumption regarding the exchangeability of data points in $\calibset$ and $\ulset$. The more technical case where exchangeability does not hold is briefly discussed in \cref{sec:non-exch}.

\begin{Assumption}\label{assump:exchangeability}
The data points in $\calibset$ and $\ulset$ are exchangeable conditioned on $\trainset$.
\end{Assumption}

\begin{Lemma}\label{lem:conformal-coverage-retain-holds}
Suppose \cref{assump:exchangeability} holds and $\pi_r>0$. Then, \cref{eq:conformal-coverage-forget} implies \cref{eq:conformal-coverage-retain}.
\end{Lemma}
\begin{proof}
From \cref{assump:exchangeability} and \cref{eq:quantile-coverage}, we have for $(X,Y)\in \ulset$, $\P(Y \in \calC_{\theta_{u}} (X)) \geq 1 - \alpha$. From \cref{eq:conformal-coverage-forget}, we obtain 
\begin{align*}
\Pfor(Y \in \calC_{\theta_{u}} (X)) & \leq 1-\beta \leq 1- \alpha. 
\end{align*}
Suppose \cref{eq:conformal-coverage-retain} does not hold (i.e., $\Pret(Y \in \calC_{\theta_{u}} (X)) < 1-\alpha$). Then, 
\begin{align*}
\P(Y \in \calC_{\theta_{u}} (X))
&=\Pret(Y \in \calC_{\theta_{u}} (X)) \pi_r + \Pfor(Y \in \calC_{\theta_{u}} (X)) \pi_f \\
&< 1 - \alpha,
\end{align*}
a contradiction, and the proof is complete.
\end{proof}

On the other hand, suppose \cref{eq:conformal-coverage-retain} holds. In the context of conformal unlearning, if the conformity scores of the data samples in $\calibset$ are continuous and free of ties \cite{lei2018distribution} (a condition that can always be met by adding a small random perturbation), then letting $n=\calibsize$, $\tbeta=\Pfor(Y \notin \calC_{\theta_{u}} (X))$, and assuming $\pi_f > 0$, we have:
\begin{align*}
1 - \alpha + \ofrac{n+1} &\geq \P(Y \in \calC_{\theta_{u}} (X))\\
&\geq (1 - \alpha)(1-\pi_f) + (1-\tbeta)\pi_f \\ 
\implies \tbeta &\geq \alpha - \ofrac{(n +1)\pi_f}.
\end{align*} 
This inequality shows that \cref{eq:conformal-coverage-forget} cannot be guaranteed for $\beta \geq \alpha$ without a non-trivial unlearning algorithm. It highlights the necessity of designing effective MU algorithms to meet the requirements of \cref{def:conformalMU}. Furthermore, it suggests that $\beta$ cannot be arbitrarily high. \gls{CP} imposes constraints on the maximum achievable $\beta$, which depends on the likelihood of the forget set $\forgetset$.

\begin{Lemma}\label{lem:alpha-beta-tradeoff}
Suppose \cref{assump:exchangeability} and \cref{eq:conformal-coverage-forget} hold. Then,
\begin{align}\label{eq:alpha-beta-tradeoff}
\pi_f \beta \le \alpha \le \beta.
\end{align}
\end{Lemma}
\begin{proof}
From \cref{eq:quantile-coverage}, we have the marginal coverage guarantee: for $(X,Y)\in \ulset$,
\begin{align*} 
& 1- \alpha \leq \P( Y \in \calC_{\theta_{u}}(X) ) \\
&= \pi_r \Pret( Y \in \calC_{\theta_{u}}(X) )  + \pi_f \Pfor( Y \in \calC_{\theta_{u}}(X) ) \\
&\leq 1 - \pi_f  + \pi_f(1-\beta) = 1 - \pi_f \beta, 
\end{align*}
where the last inequality follows from \cref{eq:conformal-coverage-forget}. Rearranging the terms gives the desired result.
\end{proof}

Any $(\alpha,\beta)$-conformal unlearning algorithm with relatively small $\alpha$ and large $\beta$ has a statistically strong unlearning property. 
For a calibration set $\calibset$ exchangeable with $\ulset$, the worst $(\alpha, \beta)$-conformal unlearning method operating on $\calibset$ is given by $\alpha = \ofrac{n+1}$ since any smaller value of $\alpha$ leads to $\hqalpha = \infty$ and $\calC_{\theta_{u}} (X) = \calY $ for all $X$. Hence, the unlearning collapses, and all samples become covered with the trivial set. In that case, the forget set miscoverage probability in \cref{eq:conformal-coverage-forget} is $0$, and is excluded from \cref{def:conformalMU}. In another trivial case where $\alpha = 1$, then \cref{lem:alpha-beta-tradeoff} gives $\beta = 1$, which is expected.

Note that not all possible pairs $(\alpha, \beta)$ are achievable, depending on the given \gls{CP} procedure. Exploring the feasibility region for coverage and miscoverage pairs is out the scope of this work and presents an interesting direction for future research.

Throughout this work, we primarily assume that $\calibset$ and $\ulset$ are exchangeable, conditioned on $\trainset$. This assumption is critical for ensuring the validity of the coverage guarantees in \cref{eq:quantile-coverage}, \cref{eq:conformal-coverage-retain}, and \cref{eq:conformal-coverage-forget}. In practical scenarios, exchangeability can often be achieved in class-wise or group-wise unlearning by reserving validation points for each class or group during the training process. 

If exchangeability is violated, more general \gls{CP} frameworks, such as the non-exchangeable CP framework with coverage gap correction \cite{barber2023conformal}, can be employed. In such cases, the coverage gap must be explicitly incorporated into all relevant bounds, including \cref{eq:quantile-coverage}, \cref{eq:conformal-coverage-retain}, and \cref{eq:conformal-coverage-forget}. A brief discussion of the non-exchangeable setting is provided in \cref{sec:non-exch}.

\subsection{A Regression Example}\label{sec:regression-example}

Consider a regression model where the data $(X,Y) \in \calX \times \bbR^m$ and $\calX$ is a subset of a Euclidean space. The relationship between the input $X$ and the output $Y$ is modeled as $Y = f_{\theta}(X) + \varepsilon$, where $f_{\theta} : \calX \to \bbR^m$ denotes the regression function parameterized by $\theta \in \bbR^d$, and $\varepsilon$ represents the noise term. Here, $d$ specifies the dimensionality of the parameter space.
This is an example of feature-based cluster-wise forgetting. We assume that $\calX = \calX_R \cup \calX_F$, where $\calX_R$ and $\calX_F$ are \emph{disjoint} feature subspaces corresponding to the retain and forget groups, respectively.\footnote{For a slightly more complex model, we can impose this assumption on the semantic space obtained after projecting raw features through a deep neural network. The analysis remains similar in that case.}
We show that under some technical conditions, we can achieve perfect conformal unlearning (i.e., $\beta=1$) in this example. Suppose that $\theta_o$ is the model parameter learned from training data before the unlearning procedure.

Following \cite{vovk2005alrw,angelopoulos2022gentle}, we choose the score function as
\begin{align*}
s(X,Y; \theta) \triangleq \norm{Y - f_\theta(X)}.
\end{align*}
Let $n$ be the size of the calibration set $\calibset$. 
The conformal predictor associated with $\theta$ is
\begin{align}\label{eq:regression-cp-set}
\calC_{\theta} (X) \triangleq \set*{y \in \bbR^m \given \norm{y - f_\theta(X)} \le \hqalpha(\theta)},
\end{align}
where  $\hqalpha(\theta)$ is the $\ceil{(1-\alpha)(n+1)}/n$ quantile of the scores computed from $\calibset$.
From \cite{lei2018distribution}, we have
\begin{align}
1-\alpha \leq \P(Y \in \calC_{\theta} (X)) \leq 1 - \alpha + \ofrac{n+1}. \label{eq:regression-cp-coverage}
\end{align}

The following assumptions are imposed to facilitate the analysis. For clarity, we use $\nabla_{\theta}f_{\theta_o} \triangleq \nabla_{\theta}f_{\theta} \mid_{\theta = \theta_o}$ to denote the Jacobian of $f_{\theta}$ with respect to $\theta$, evaluated at $\theta = \theta_o$.

\begin{Assumption}\label{assum:regression-model}

\begin{enumerate}[label=(\roman*),ref={\theAssumption(\roman*)}]

\item\label[Assumption]{assum:continuous-variables} \emph{Continuous variables.}
$(X,Y)$ are continuous random variables, i.e., their joint distribution admits a probability density function.

\item\label[Assumption]{assum:bounded-noise} \emph{Bounded noise.}
We have $\norm{\varepsilon} \le \sigma$ almost surely ($\as$) for some $\sigma>0$.

\item\label[Assumption]{assum:smooth-theta} \emph{Smoothness with respect to $\theta$.}
The gradient of the parameterized model $f_{\theta}$ is assumed to be $L$-Lipschitz continuous with respect to $\theta$ in a neighborhood around $\theta_o$. Formally, for all $x \in \calX$, the following condition holds:
\begin{align*}
\|\nabla_{\theta}f_{\theta}(x) - \nabla_{\theta}f_{\theta_o}(x)\| \leq L \|\theta - \theta_o\|,
\end{align*}
where $L > 0$ is the Lipschitz constant. This assumption implies that the model can be locally approximated by its first-order Taylor expansion around $\theta_o$ with bounded error.

\item\label[Assumption]{assum:directional-separability} \emph{Directional separability in parameter space.}
There exists a unit vector $v \in \bbR^d$ (i.e., $\norm{v} = 1$) such that, for some constants $0 \leq a_R \leq a_F$, the following conditions hold:
\begin{align*}
&\sup_{x \in \calX_R} \|\nabla_{\theta}f_{\theta_o}(x)v\| \leq a_R, \\
&\inf_{x \in \calX_F} \|\nabla_{\theta}f_{\theta_o}(x)v\| \geq a_F.
\end{align*}
Denote $\Delta a = a_F - a_R > 0$. Assume that $(\Delta a)^2 \geq 8 \sigma L$. 
In other words, the model $f_{\theta_o}$ exhibits limited variation in the direction $v$ within the retain feature space $\calX_R$, while demonstrating significant variation in the same direction within the forget feature space $\calX_F$. Furthermore, the difference in variation between the two feature spaces is sufficiently large relative to the noise level $\sigma$ and the Lipschitz constant $L$.

\item\label[Assumption]{assum:small-df} \emph{Retained subpopulation mass.}
The proportion of the retained subpopulation satisfies 
\begin{align*}
\P(\calX_R) \geq 1 - \alpha + \ofrac{n+1}.
\end{align*}

\end{enumerate}
\end{Assumption}

\begin{Theorem}\label{theo:linear-cmu-exists}[Achievability of ($\alpha$,1)-conformal unlearning in regression.] 
Suppose \cref{assum:regression-model} holds with $\theta_o$ being the global minimizer of the population loss before unlearning. Then, there exists a step size $\gamma^* \in (\gamma_{-},\gamma_{+})$ with 
\begin{align}\label{eq:gamma-pm}
\gamma_{\pm} = \frac{\Delta a \pm \sqrt{(\Delta a)^2 - 8 \sigma L}}{2 L},
\end{align}
such that the unlearned parameters $\theta_u = \theta_o + \gamma^* v$ achieve ($\alpha$,$\beta$)-conformal unlearning with $\beta = 1$ for the \gls{CP} procedure defined by \cref{eq:regression-cp-set}.
\end{Theorem}
\begin{proof}
Let $\theta_{\gamma} = \theta_o+ \gamma v$, where $v$ is the unit vector from \cref{assum:directional-separability}. Then by \cref{assum:smooth-theta} and the Taylor expansion around $\theta_o$, we have $\as$,
\begin{align*}
Y - f_{\theta_o + \gamma v}(X) &= f_{\theta_o}(X) + \varepsilon - f_{\theta_o+\gamma v}(X)\\
&=  \varepsilon - \gamma \nabla_{\theta}f_{\theta_o}(X)v - r(X,\gamma).
\end{align*}
where the remainder $\|r(X,t)\| \le \ofrac{2}Lt^2$. 

Using the triangle inequality and \cref{assum:bounded-noise}, for any $X \in \calX_R$, we have
\begin{align}
\|Y - f_{\theta_o+ \gamma v}(X)\| 
&\le |\gamma|\|\nabla_{\theta}f_{\theta_o}(X)v\| + \sigma + \ofrac{2}L \gamma^2 \nn
&\leq |\gamma|a_R + \sigma + \ofrac{2}L \gamma^2 \triangleq r_R^{\max}(\gamma). \label{eq:bounded-RR}
\end{align}
Similarly, for any $X \in \calX_F$, we have
\begin{align}\label{eq:bounded-RF}
\|Y - f_{\theta_o + \gamma v}(X)\| \geq |\gamma|a_F - \sigma - \ofrac{2}L \gamma^2 \triangleq r_F^{\min}(\gamma).   
\end{align}
If there exists a step size $\gamma^*$ such that $r_R^{\max}(\gamma^*) < r_F^{\min}(\gamma^*)$, then we have 
\begin{align*}
&\P( s(X,Y; \theta_{\gamma^*}) \leq r_R^{\max}(\gamma^*) )
= \P( \set*{ (X,Y) \given X \in \calX_R }) \\
& \geq 1 - \alpha + \ofrac{n+1} \geq \P( s(X,Y; \theta_{\gamma^*}) \leq \hqalpha(\theta_{\gamma^*}) ), 
\end{align*}
where the first inequality follows from \cref{assum:small-df} and the second inequality from \cref{eq:regression-cp-coverage}. Hence, $ \hqalpha(\theta_{\gamma^*}) \leq r_R^{\max}(\gamma^*) < r_F^{\min}(\gamma^*) \as$ Therefore, $\P(Y \in \calC_{\theta_{\gamma^*}}(X) \given X \in \calX_F) = 0$, or $\beta = 1$, while the $1-\alpha$ coverage is retained on $X \in \calX_R$ by \cref{lem:conformal-coverage-retain-holds}. Thus, $\theta_u = \theta_{\gamma^*}$ achieves ($\alpha$,$1$)-conformal unlearning.

To find such a step size $\gamma^*$, letting $r_R^{\max}(\gamma) < r_F^{\min}(\gamma)$, we obtain
\begin{align}
& |\gamma|(a_F-a_R)-L\gamma^2 > 2\sigma \\
& L\gamma^2 - |\gamma|\Delta a + 2\sigma < 0.\label{eq:quadratice-form-t}
\end{align}
Taking $\gamma \geq 0$, the inequality in \cref{eq:quadratice-form-t} is satisfied for $\gamma \in (\gamma_{-},\gamma_{+})$, where $\gamma_{\pm}$ are defined in \cref{eq:gamma-pm}.
The proof is now complete by choosing $\gamma^* \in (\gamma_{-},\gamma_{+})$.
\end{proof}

Although \cref{assum:regression-model} and \cref{theo:linear-cmu-exists} establish theoretical conditions for achieving perfect conformal unlearning ($\beta=1$) in the regression example, these conditions may not always hold in practical scenarios. Consequently, empirical methodologies are essential to effectively optimize the unlearning process. The conformal unlearning framework introduced in \cref{sec:method} is specifically designed to address this need.


\subsection{Zero-Shot Conformal Unlearning}\label{sec:zero-shot-conformal-unlearning}

In the case of \emph{zero-shot} \gls{MU} (as defined in \cite{foster2024information}) where sampling from the same distribution as the forget set $\forgetset$ is not available, we are restricted to a calibration set whose samples are from the same underlying distribution as the retain set $\retainset$ (i.e., $\calibset \sim \pret$). We have the following result.

\begin{Proposition}\label{prop:worse-forget-conformity-scores}
Suppose $\calibset$ and $\retainset$ are exchangeable, and are independent of $\forgetset$. Let $(X_f, Y_f) \sim \Pfor$, and $(X_r, Y_r) \sim \Pret$.
An ($\alpha$,$\beta$)-conformal unlearning algorithm $\Ualgo$ based on $\calibset$ and $\retainset$ yields
\begin{align}\label{eq:worse-forget-conformity-scores}
\P(s(X_f, Y_f;\theta_u) \geq s(X_r, Y_r;\theta_u)) \geq \beta (1-\alpha).
\end{align}
\end{Proposition}
\begin{proof}
From \cref{eq:conformal-coverage-forget}, we have
\begin{align*}
\beta
&\leq \P(s(X_f, Y_f;\theta_u) > \hat q_\alpha)\\
&\leq \P(s(X_f, Y_f;\theta_u) \geq s(X_r, Y_r;\theta_u) \given  s(X_r, Y_r;\theta_u) \leq \hat q_\alpha )\\
&\leq \frac{ \P(s(X_f, Y_f;\theta_u) \geq s(X_r, Y_r;\theta_u)) }{\P(s(X_r, Y_r;\theta_u) \leq \hat q_\alpha) } \nn
&\leq \frac{ \P(s(X_f, Y_f;\theta_u) \geq s(X_r, Y_r;\theta_u)) }{1-\alpha}, 
\end{align*}
where the inequality follows from \cref{eq:conformal-coverage-forget}, the second inequality holds due to independence of $\forgetset$ from $\calibset,\retainset$, and the last inequality follows from \cref{eq:conformal-coverage-retain}. Therefore, the result holds.
\end{proof}

\Cref{prop:worse-forget-conformity-scores} establishes that for small $\alpha$ and large $\beta$, an ($\alpha$,$\beta$)-conformal unlearning algorithm demonstrates, on average, higher non-conformity on the forget data compared to the retained data. When the conformity scoring function $s$ corresponds to a loss function, this implies that the loss incurred on $\forgetset$ is, with high probability, greater than the loss on $\retainset$. This result aligns with the intuition articulated following \cref{def:conformalMU}.


\subsection{Efficiency-Aware Conditional Conformal Unlearning}\label{sec:efficient-conformal-unlearning}

\gls{CP} sets identify the most probable labels for a given test sample. However, excessively large prediction sets tend to lose their informativeness, which is undesirable for predictions on $\retainset$ but aligns with the objective of unlearning on $\forgetset$. To address this, we propose a refined version of \cref{def:conformalMU}, where the coverage and miscoverage guarantees are constrained to efficient (i.e., small) prediction sets.

\begin{Definition}[($c$,$d$)-efficient ($\alpha$,$\beta$)-conformal unlearning]\label{def:eff-conformalMU}
An unlearning algorithm $\Ualgo$ is said to be ($c$,$d$)-efficient ($\alpha$,$\beta$)-conformal unlearning for $0 \leq \alpha \leq \beta \leq 1$ and integers $c,d \in \set{0, \dots, |\calY|}$, if
\begin{align}
&\Pret(Y \in \calC_{\theta_{u}} (X) \given |\calC_{\theta_{u}} (X)| \le c)  \geq 1 - \alpha, \label{eq:efficient-conformal-retain}\\
&\Pfor(Y \notin \calC_{\theta_{u}} (X) \given |\calC_{\theta_{u}} (X)| \le d)  \geq \beta. \label{eq:efficient-conformal-forget}
\end{align}
\end{Definition}

The thresholds $c$ (for retained data) and $d$ (for forgotten data) define the maximum prediction-set sizes considered \emph{informative} by the unlearning framework. Prediction sets exceeding these thresholds are classified as ``inefficient'' and are \emph{excluded} from coverage calculations, as they are excessively broad and indicative of low model confidence \cite{shafer2008conformal}. The selection of $c$ and $d$ is context-dependent and reflects the unlearner’s tolerance for uncertainty in the prediction sets.

As an illustration, consider a 20-class document classification task. If the model produces a prediction set of size $10$ that includes a confidential label intended for forgetting, the set may be deemed too large to represent a significant information leak. Conversely, a smaller prediction set of size $5$ containing the same label would be more informative and thus indicative of incomplete forgetting. For retained labels, a prediction set of size $7$ may still provide sufficient specificity to be practically useful. Practitioners may therefore select $d=5$ for forgotten data and $c=7$ for retained data, reflecting their tolerance for uncertainty in each case. Evaluating multiple $(c,d)$ pairs can provide insights into the trade-offs between coverage and miscoverage. Notably, when $c=d=|\calY|$, \cref{def:eff-conformalMU} simplifies to the standard ($\alpha$,$\beta$)-conformal unlearning framework.

\begin{Corollary}\label{cor:efficient-cov-bound}
Under the same assumptions as \cref{lem:conformal-coverage-retain-holds}, suppose a MU algorithm $\Ualgo$ is ($\alpha$,$\beta$)-conformal unlearning. For $c, d \in \set{0, \dots, |\calY|}$, assume $\Pret(|\calC_{\theta_{u}} (X)|> c ) \leq \zeta_c$, and $\Pfor(|\calC_{\theta_{u}} (X)|> d ) \leq \eta_d$. Then, we have
\begin{align}
\Pret(Y \in \calC_{\theta_{u}} (X) \given |\calC_{\theta_{u}} (X)| \le c)
&\geq 1 - \alpha - \zeta_c, \label{eq:efficient-conformal-retain-bound}\\
\Pfor(Y \in \calC_{\theta_{u}} (X) \given |\calC_{\theta_{u}} (X)| \le d)
&\geq \beta - \eta_d. \label{eq:efficient-conformal-forget-bound}
\end{align}
\end{Corollary}
\begin{proof}
Define the following events:
\begin{align*}
A &= \set*{Y \in \calC_{\theta_{u}} (X)},\\
Q &= \set*{\abs{\calC_{\theta_{u}} (X)} \le c},\\
M &= \set*{\abs{\calC_{\theta_{u}} (X)} \le d}.
\end{align*}

From \cref{eq:conformal-coverage-retain}, we have
\begin{align}
1 - \alpha \le \Pret(A \given Q). \label{pABl}
\end{align} 
On the other hand, by the law of total probability, we have
\begin{align}
\Pret(A\given Q) &= \Pret(A\given Q)\Pret(Q) + \Pret(A\given Q\setcomp)\Pret(Q\setcomp) \nn
&\leq \Pret(A\given Q) + \Pret(Q\setcomp) \nn
&\leq \Pret(A\given Q) + \zeta_c. \label{pABu}
\end{align}
Combining \cref{pABl} and \cref{pABu}, we have
\begin{align}
\Pret(A\given Q) \geq 1 - \alpha - \zeta_c. \label{pABQ}
\end{align}

Furthermore, from \cref{eq:conformal-coverage-forget}, we have
\begin{align}
\beta \le \Pfor(A\setcomp). \label{pABn}
\end{align} 
Again, by the law of total probability, we have
\begin{align}
\Pfor(A\setcomp) &= \Pfor(A\setcomp, M) \Pfor(M) + \Pfor(A\setcomp, M\setcomp) \Pfor(M\setcomp) \nn
&\leq \Pfor(A\setcomp \given M) + \Pfor(M\setcomp) \nn
&\leq \Pfor(A\given M) + \eta_d. \label{pABk}
\end{align}
Combining \cref{pABn} and \cref{pABk}, we have
\begin{align}
\Pfor(A\setcomp \given M) \geq \beta - \eta_d. \label{pABM}
\end{align}

The proof is now complete.
\end{proof}

\Cref{cor:efficient-cov-bound} indicates that by having a sufficiently good model so that $\zeta_c$ and $\eta_d$ are small, we can achieve a good bound on the coverage of the retained points and the miscoverage of the forgotten points by efficient sets.


\section{Empirical Metrics and Optimization Framework}\label{sec:method}

In this section, we introduce two novel metrics, which serve as practical measures for evaluating coverage and miscoverage. Furthermore, we present the EFFiciency constrAined Conformal unlEarning (EFFACE) framework, a systematic approach designed to optimize these metrics and achieve effective unlearning.


\subsection{Empirical Conformal Unlearning Metrics}

Building on \cref{def:eff-conformalMU}, the objective is to optimize the left-hand sides of \cref{eq:efficient-conformal-retain,eq:efficient-conformal-forget}, thereby achieving reliable coverage on retained data and significant miscoverage on forgotten data, under the case where prediction sets are efficient. To approximate probabilities, we employ empirical frequencies \cite{shi2025rethinking}. Specifically, for a retained dataset $\retainset$ and a positive integer $c$, the \emph{Efficiently Covered Frequency} (ECF) at threshold $c$ for a prediction set $\calC(\cdot)$ is defined as:
\begin{align}\label{eq:ecf-retain}
\ECF_{c}[\retainset]  = \ofrac{|\calD_{r,c}|} \sum_{(x,y) \in \calD_{r,c}} \indicate*{y \in \calC (X)},
\end{align}
where $\calD_{r,c} = \set*{(x,y) \in \retainset \given  |\calC (X)| \le c}$ and $\indicate{}$ is the indicator function.

On the other hand, for a forget dataset $\forgetset$ and a positive integer $d$, we define the \emph{Efficiently Miscovered Frequency} (EmCF) at threshold $d$ of a prediction set $\calC(\cdot)$ as 
\begin{align}\label{eq:ecf-forget}
\EmCF_{d}[\forgetset]  = \ofrac{|\calD_{f,d}|} \sum_{(x,y) \in \calD_{f,d}} \indicate*{y \notin \calC (X)},
\end{align}
where $\calD_{f,d} = \set*{(x,y) \in \forgetset \given  |\calC (X)| \le d}$.

\subsection{Unlearning via Optimizing Conformal Sets}\label{subsec:framework}

For a \gls{CP} set $\calC_{\theta,t}$ and a dataset $\calN \in \set{\retainset, \forgetset}$, the conditional miscoverage and conditional efficiency losses are defined as follows:
\begin{align}
&\Lcov[\calN](\calC_{\theta,t}) \triangleq \P(Y \notin \calC_{\theta,t} (X) \given (X,Y)\in\calN), \label{eq:conditional-miscoverage}\\
&\Leff[\calN](\calC_{\theta,t}) \triangleq \E[\losseff(\calC_{\theta,t};(X,Y)) \given (X,Y)\in\calN]. \label{eq:conditional-efficiency-loss}
\end{align}
Their corresponding empirical versions are given by
\begin{align}\label{eq:empirical-miscoverage}
\Lcov*[\calN](\calC_{\theta,t}) & \triangleq \ofrac{|\calN|} \sum_{(x, y) \in \calN} \indicate{y \notin \calC_{\theta, t}(x)}, \\
\Leff*[\calN](\calC_{\theta,t}) & \triangleq \ofrac{|\calN|} \sum_{(x,y) \in \calN} \losseff(\calC_{\theta,t};(x,y)).
\end{align}

Inspired by the differentiable framework introduced in \cite{bai2022efficient}, we recast the conformal unlearning problem as a constrained empirical risk minimization (ERM) task. The primary objective is to maximize the miscoverage on the forget set $\forgetset$ while simultaneously minimizing the miscoverage on the retain set $\retainset$, thereby adhering to the principles of conformal unlearning. To ensure the efficiency of the prediction sets, constraints are imposed on the efficiency losses for both $\retainset$ and $\forgetset$. The resulting optimization problem is formulated as follows:
\begin{subequations}\label{problem:efface-erm}
\begin{align}
\min_{\theta, t}\ & 
\Lcov*[\retainset](\calC_{\theta,t}) - \Lcov*[\forgetset](\calC_{\theta,t}),\\
\ST\ 
&\Leff*[\retainset](\calC_{\theta,t}) \le c, \label{eq:retain-constraint}\\
&\Leff*[\forgetset](\calC_{\theta,t}) \le d. \label{eq:forget-constraint}
\end{align}
\end{subequations}
Excessively large prediction sets may fail to provide actionable insights for decision-makers. For forget data, however, the threshold $d$ can be set relatively high, as the primary objective is to ensure that coverage occurs only infrequently, potentially resulting in prediction sets that are less informative. In downstream applications, decision-makers may choose to reconformalize $t$ to re-establish marginal conformal validity across the data mixture. 

We refer to our proposed approach as EFFiciency constrAined Conformal unlEarning (EFFACE) and is presented in \cref{alg:framework-algorithm}. To address the non-differentiability of the indicator function, we employ a sigmoid hinge approximation, which facilitates gradient-based optimization. In that case, $\Lcov*[\calD](\calC_{\theta,t}) \approx \Lcov*[\calD]'(\calC_{\theta,t}) = \ofrac{|\calD|} \sum_{(x, y) \in \calD} \sigma\left(\kappa \cdot (s(x,y) - t\right)$. Moreover, in practice, the model might be prone to learning an easy threshold $t$ to minimize the objective in \cref{problem:efface-erm}. We found that choosing $\hat t = \hqalpha$ of the conformity scores of $\ulset$ at each epoch during unlearning yields better generalization empirically but requires the level $\alpha$ as an input to the algorithm. Additionally, a regularization term, $\gamma \norm{\theta_o - \theta_u}^2$, is incorporated into the objective function to mitigate excessive deviation of the unlearned model's parameters from the original model's parameters, thereby preserving utility. The steepness parameter of the sigmoid hinge, $\kappa$, and the regularization coefficient, $\gamma$, are treated as hyperparameters and are subject to fine-tuning. Following these relaxations, the aim is to solve the following updated minimization problem.
\begin{subequations}\label{problem:efface-erm-relaxed}
\begin{align}
\min_{\theta}\ & 
\Lcov*[\retainset]'(\calC_{\theta,\hat t}) - \Lcov*[\forgetset]'(\calC_{\theta,\hat t}) + \gamma \norm{\theta_o - \theta}^2 ,\\
\ST\ 
&\Leff*[\retainset](\calC_{\theta,\hat t}) \le c,\\
&\Leff*[\forgetset](\calC_{\theta,\hat t}) \le d.
\end{align}
\end{subequations}
The impact of these hyperparameters, as well as the constraints $c$ and $d$, is analyzed in the sensitivity analysis provided in \cref{app:sensitivity} of the supplementary material.

\begin{algorithm}[!tb]
\caption{EFFACE}
\label{alg:framework-algorithm}
\begin{algorithmic}[1] 
\Require Retained data $\retainset$, forget data $\forgetset$, calibration data $\calibset$ with $|\calibset|=n$, conformity scoring function $s(X,Y)$, size constraints $c, d$, steepness $\kappa$, regularization constant $\gamma$, miscoverage tolerance $\alpha$
\State Define $\calC _{\theta,\hat t}(x) = \{y : s(x,y) \le \hat t\}$
\State Solve \cref{problem:efface-erm-relaxed} to obtain $\theta_u$.

\State Compute $\hqalpha$ as the $\ceil{(1-\alpha)(n+1)}/n$ quantile of $\{s(X,Y;\theta_u) : (X,Y) \in \calibset\}$. 
\State Set $\calC _{\theta_u}(X) = \{y : s(X,y;\theta_u) \le \hqalpha\}$
\Ensure $\theta_u, \calC _{\theta_u}(X)$
\end{algorithmic}
\end{algorithm}


\subsection{Generalization Bounds}\label{subsec:generalization-bounds}

Let $(\widehat \theta,\widehat t)$ denote a solution obtained from the ERM problem in \cref{problem:efface-erm}. To quantify the generalization gap between the empirical and population-level metrics, we define the following conditional concentration terms:
\begin{align}\label{eq:varpsilons}
\begin{aligned}
\varepsilon_r &\triangleq\sup_{\theta,t}\abs*{\Lcov[\retainset](\calC_{\theta,t}) - \Lcov*[\retainset](\calC_{\theta,t})},\\
\varepsilon_f &\triangleq \sup_{\theta,t}\abs*{\Lcov[\forgetset](\calC_{\theta,t})-\Lcov*[\forgetset](\calC_{\theta,t})},\\
\varepsilon_{\mathrm{eff},r} &\triangleq \sup_{\theta,t}\abs*{\Leff[\retainset](\calC_{\theta,t})-\Leff*[\retainset](\calC_{\theta,t})},\\
\varepsilon_{\mathrm{eff},f} &\triangleq \sup_{\theta,t}\abs*{\Leff[\forgetset](\calC_{\theta,t})-\Leff*[\forgetset](\calC_{\theta,t})}.
\end{aligned}
\end{align}
These terms characterize the maximum deviation between the empirical and true values of the miscoverage and efficiency losses, conditioned on the retain and forget sets, respectively.

\begin{Proposition}\label{prop:efface}
A solution $(\widehat \theta,\widehat t)$ of \cref{problem:efface-erm} satisfies the following:
\begin{enumerate}[(a)]
\item\label[claim]{prop:efface-approx-eff} \textbf{Approximate conditional efficiencies.}
\begin{align}
&\Leff[\retainset](\calC_{\widehat\theta,\widehat t}) \le c + \varepsilon_{\mathrm{eff},r}, 
\Leff[\forgetset](\calC_{\widehat\theta,\widehat t}) \le d + \varepsilon_{\mathrm{eff},f}.
\end{align}

\item\label[claim]{prop:efface-near-opt} \textbf{Near-optimal conditional miscoverage \emph{difference}.}
Suppose $\varepsilon_{\mathrm{eff},r} < c$ and $\varepsilon_{\mathrm{eff},f} < d$. Let 
$\calU \triangleq \set{(\theta, t) \given \Leff[\retainset](\calC_{\theta,t}) \le c - \varepsilon_{\mathrm{eff},r},\ \Leff[\forgetset](\calC_{\theta,t}) \le d - \varepsilon_{\mathrm{eff},f}}$.
Then
\begin{align}
&\Lcov[\retainset](\calC_{\widehat\theta,\widehat t})-\Lcov[\forgetset](\calC_{\widehat\theta,\widehat t}) \nn
&\le \inf_{(\theta, t)\in \calU}\ \Big(\Lcov[\retainset](\calC_{\theta,t}) - \Lcov[\forgetset](\calC_{\theta,t})\Big) + 2\varepsilon_r + 2\varepsilon_f.
\end{align}

\end{enumerate}
\end{Proposition}

\begin{proof}
To prove \cref{prop:efface-approx-eff}, note that since $(\widehat\theta,\widehat t)$ is a feasible solution, $\Leff*[\retainset](\calC_{\widehat\theta,\widehat t}) \le c$. Therefore,
\begin{align*}
\Leff[\retainset](\calC_{\widehat\theta,\widehat t})
&= \Leff*[\retainset](\calC_{\widehat\theta,\widehat t})
+ \big( \Leff[\retainset]-\Leff*[\retainset]\big)(\calC_{\widehat\theta,\widehat t})\nonumber\\
&\le c+\varepsilon_{\mathrm{eff},r}.
\end{align*}
A similar proof holds for $\Leff[\forgetset](\calC_{\widehat\theta,\widehat t}) \le d+\varepsilon_{\mathrm{eff},f}$.

We next prove \cref{prop:efface-near-opt}. We have
\begin{align*}
&\Lcov[\retainset](\calC_{\widehat\theta,\widehat t}) -\Lcov[\forgetset](\calC_{\widehat\theta,\widehat t}) \\
&\le \Lcov*[\retainset](\calC_{\widehat\theta,\widehat t})-\Lcov*[\forgetset](\calC_{\widehat\theta,\widehat t}) + \varepsilon_r + \varepsilon_f\\
&\le \Lcov*[\retainset](\calC_{\theta,t})-\Lcov*[\forgetset](\calC_{\theta,t}) + \varepsilon_r + \varepsilon_f \\
&\leq \Lcov[\retainset](\calC_{\theta,t})-\Lcov[\forgetset](\calC_{\theta,t}) + 2\varepsilon_r + 2\varepsilon_f
\end{align*}
for any $(\theta,t)\in\calU$. The first and last inequalities follow from the definitions of $\varepsilon_r$ and $\varepsilon_f$, and the second inequality follows from the optimality of $(\widehat\theta,\widehat t)$ and the fact that $(\theta,t)$ is feasible for \cref{problem:efface-erm}.
Taking the infimum over $(\theta,t)\in\calU$ gives the claim.
\end{proof}

Concrete bounds for the terms in \cref{eq:varpsilons} under finite/VC/Rademacher classes are provided in App.~C of \cite{bai2022efficient}, leading to rates of order $\sqrt{\mathrm{Comp}(\calC)/\nulset}$, where $\mathrm{Comp}(\calC)$ is the complexity measure of the class $\calC$. Therefore, by having a sufficiently large $\nulset$, these terms can be made arbitrarily small. We refer the reader to \cite{bai2022efficient} for more details.


\section{Numerical Experiments}\label{sec:exp}

\textbf{Datasets and Models.} We conduct evaluations on CIFAR100 \cite{cifar100}, and a subset of Tiny-ImageNet comprising 100 classes, referred to as ImagenNet100 \cite{imagenet100}. In addition, we present results on the 20 Newsgroups dataset with 20 classes \cite{20newsgroups} in \cref{app:extra-empirics} of the supplementary material. The model before unlearning is ResNet18. Unless otherwise specified, all results are averaged over six random seeds for all baselines, except for the retrained (\gls{RT}) model and the certified method (PABI), where results are averaged over three random seeds due to their significant computational overhead. Notably, we observe minimal variability in their outcomes.

\textbf{Data Partitions.} The experimental setup involves six distinct data subsets: training forget/retain $(\calT_f,\calT_r)$, unlearning forget/retain $(\calD_f,\calD_r)$, and unseen forget/retain $(\calV_f,\calV_r)$. For ImagenNet100, the training dataset comprises 117k images, with an additional 13k images reserved for validation and final testing. Specifically, 6.5k images are allocated for validation by methods that require validation during the unlearning process, while the remaining 6.5k images are evenly divided into $\calV_f$ and $\calV_r$ for final evaluation. Additionally, a 4k calibration set $\calibset$ is extracted from the 5k test split to construct \gls{CP} sets and define label-based $\forgetset$ and $\retainset$ for the primary unlearning task. 

The proposed framework applies unlearning to $\forgetset$ and $\retainset$. When these sets are disjoint from the training data $(\calT_f,\calT_r)$, the scenario is referred to as \emph{out-sample unlearning} (\emph{Out}). Conversely, when $\forgetset$ and $\retainset$ are subsets of the training data $(\calT_f,\calT_r)$, the scenario is termed \emph{in-sample unlearning} (\emph{In}). Results for both scenarios are presented in the corresponding tables. Detailed information to ensure reproducibility is provided in \cref{app:reproduce} of the supplementary material.

In the experimental results detailed in \cref{sec:results}, we perform cluster-wise unlearning, which entails partitioning the training data into $k$ clusters within the embedding space using the $k$-means clustering algorithm, where $k$ corresponds to the number of classes. Each data point is assigned a pseudo-label based on its proximity to the nearest cluster centroid. The forget data are subsequently identified based on these pseudo-labels, ensuring that the data designated for unlearning share common high-level characteristics. This approach aligns with the foundational principles of conformal unlearning, as outlined in \cref{sec:preliminary}. Additional results encompassing both cluster-wise and label-wise unlearning are provided in \cref{app:extra-empirics} of the supplementary material.

\textbf{Training and Unlearning Procedures.} For CIFAR100, the training process employs stochastic gradient descent (SGD) over 50 epochs, with an initial learning rate of $0.1$ decaying linearly to $10^{-4}$, a momentum of $0.9$, and a weight decay of $5{\times}10^{-4}$. ImageNet100 follows a similar configuration, extended to 80 epochs. Text models are trained for 15 epochs with an initial learning rate of $0.01$. All experiments utilize a batch size of $256$ and two data-loading workers, with standard normalization and data augmentation techniques applied. The unlearning optimizer is configured to match the base training optimizer, maintaining the same momentum and weight decay, while employing a tuned learning rate and no learning rate scheduler across all methods. For RT and PABI, the original training algorithm and hyperparameter setup are used for fine-tuning. All experiments are conducted on four NVIDIA RTX A5000 GPUs, utilizing \texttt{nn.DataParallel} to ensure efficient parallelization across all methods.

\textbf{Baselines.} We evaluate our approach against several state-of-the-art unlearning methods, including $\nabla\tau$ \cite{trippa2024gradient}, SCRUB \cite{kurmanji2023towards}, SSD \cite{foster2024fast}, AMN \cite{graves2021amnesiac}, BADT \cite{chundawat2023bad}, UNSIR \cite{tarun2024fast}, and the RT baseline applied to $\calT_r$. Additionally, we compare against the certified unlearning method PABI \cite{koloskova2025certifiednonconvex}, which also operates exclusively on $\calT_r$. For implementation, we utilize the authors' publicly available code for methods from \cite{foster2024fast,chundawat2023bad} and re-implement PABI by introducing a dedicated function for gradient clipping steps followed by fine-tuning on $\calT_r$. To ensure a fair comparison, we perform grid-search hyperparameter tuning for each method. 
Finally, we perform conformalization on all methods using $\calibset$ to obtain valid unlearned \gls{CP} sets.
Further details on these baselines are provided in \cref{app:baselines} of the supplementary material.

\textbf{Evaluation Metrics.} The following metrics are utilized to assess the performance of the proposed framework: $\ECF_{\calD}(c)$, representing the efficiently covered frequency on retained subsets $(\calD_r, \calT_r, \calV_r)$; $\EmCF_{\calD}(d)$, denoting the efficiently miscovered frequency on forgotten subsets $(\calD_f, \calT_f, \calV_f)$, where $c = d$; and the harmonic mean $H$ of these six conformal metrics, defined as $H = n / \sum_i x_i^{-1}$, with $H = 0$ if any $x_i = 0$. Furthermore, we report the accuracy $A_{\calD}$ for each subset $\calD \in \{\calD_r, \calD_f, \calT_r, \calT_f, \calV_r, \calV_f\}$ (before the conformalization) to further illustrate the phenomenon of fake unlearning. Additional metrics include the Membership Inference Attack (MIA) Difference, calculated as the difference between the attacker's accuracy percentage and the majority-class ratio, and the unlearning time, denoted as Tsec, measured in seconds.

For the RT and PABI methods, which perform unlearning on the entirety of the retained dataset $\calT_r$, we omit results for the subsets $\calD_r \subset \calT_r$ and $\calD_f \subset \calT_f$. Complete results for ImagenNet100 are presented in the main text, while comprehensive results for both vision and text datasets are provided in \cref{app:extra-empirics} of the supplementary material.


\subsection{Results And Discussion}\label{sec:results}

\subsubsection{Coverage and miscoverage frequencies}
In \cref{tab:results_imagenet_clusters_f5_c50_a005_coverage,tab:results_imagenet_classes_f5_c50_a005_coverage}, the best, second, and third best scores are highlighted in \first{}, \second{}, and \third{}, respectively. Results from the original (OR) model (prior to unlearning) are included for reference. We use \colorbox{green!15}{green} to highlight results that appear favorable at first glance, and \colorbox{gray!20}{gray} to flag corresponding results from the same method—sometimes in a different table or scenario—that reveal poor performance or expose the green-highlighted result as misleading. When a method shows only green-highlighted results with no gray counterpart, the highlighting simply indicates genuinely strong performance worthy of note.


\begin{table*}[htbp]
\caption{ImageNet100, RepVGG-a2 \emph{cluster}-wise forgetting with $c=d=50$, $\alpha=0.05$, and 5 forgotten clusters. Coverage/miscoverage results.}
\label{tab:results_imagenet_clusters_f5_c50_a005_coverage}
\centering
\resizebox{\linewidth}{!}{%
\begin{tabular}{llccccccc}
\toprule
Split & Method
& $\ECF_{c}[\retainset]\uparrow$
& $\EmCF_{d}[\forgetset]\uparrow$
& $\ECF_{c}[\calT_r]\uparrow$
& $\EmCF_{d}[\calT_f]\uparrow$
& $\ECF_{c}[\calV_r]\uparrow$
& $\EmCF_{d}[\calV_f]\uparrow$
& $H\uparrow$ \\
\midrule

& OR
& $0.99 \pm 0.00$
& $0.00 \pm 0.00$
& $1.00 \pm 0.00$
& $0.00 \pm 0.00$
& $0.98 \pm 0.00$
& $0.02 \pm 0.00$
& $0.01 \pm 0.00$ \\

\midrule

\multirow{6}{*}{In}

& $\nabla\tau$
& $0.00 \pm 0.00$
& $0.00 \pm 0.00$
& $0.00 \pm 0.00$
& \cellcolor{gray!20}$0.00 \pm 0.00$
& $0.00 \pm 0.00$
& $0.00 \pm 0.00$
& \cellcolor{gray!20}$0.00 \pm 0.00$ \\

& SCRUB
& \first{$1.00 \pm 0.00$}
& \third{$0.33 \pm 0.09$}
& \first{$1.00 \pm 0.00$}
& \second{$0.32 \pm 0.09$}
& \second{$0.99 \pm 0.00$}
& \second{$0.35 \pm 0.08$}
& \second{$0.50 \pm 0.06$} \\

& SSD
& $0.99 \pm 0.00$
& $0.04 \pm 0.04$
& $0.99 \pm 0.00$
& \cellcolor{gray!20}$0.05 \pm 0.05$
& $0.98 \pm 0.00$
& $0.07 \pm 0.05$
& $0.10 \pm 0.05$ \\

& AMN
& \first{$1.00 \pm 0.00$}
& \cellcolor{green!15}\first{$1.00 \pm 0.00$}
& $0.99 \pm 0.00$
& \cellcolor{gray!20}\third{$0.24 \pm 0.02$}
& $0.98 \pm 0.00$
& \cellcolor{gray!20}\third{$0.21 \pm 0.02$}
& \third{$0.46 \pm 0.02$} \\

& BADT
& $0.99 \pm 0.00$
& $0.10 \pm 0.01$
& $0.99 \pm 0.00$
& $0.10 \pm 0.01$
& $0.98 \pm 0.00$
& $0.17 \pm 0.02$
& $0.21 \pm 0.01$ \\

& EFFACE
& \first{$1.00 \pm 0.00$}
& \cellcolor{green!15}\second{$0.85 \pm 0.02$}
& \first{$1.00 \pm 0.00$}
& \cellcolor{green!15}\first{$0.78 \pm 0.01$}
& \first{$1.00 \pm 0.00$}
& \cellcolor{green!15}\first{$0.79 \pm 0.00$}
& \first{$0.89 \pm 0.00$} \\

\midrule

\multirow{6}{*}{Out}

& $\nabla\tau$
& \first{$1.00 \pm 0.00$}
& $0.02 \pm 0.02$
& $0.99 \pm 0.00$
& $0.01 \pm 0.00$
& $0.98 \pm 0.00$
& $0.02 \pm 0.00$
& $0.03 \pm 0.01$ \\

& SCRUB
& \first{$1.00 \pm 0.00$}
& \second{$0.99 \pm 0.01$}
& \first{$1.00 \pm 0.00$}
& \second{$0.56 \pm 0.01$}
& \first{$0.99 \pm 0.00$}
& \second{$0.56 \pm 0.02$}
& \second{$0.79 \pm 0.01$} \\

& SSD
& $0.96 \pm 0.00$
& $0.06 \pm 0.00$
& \first{$1.00 \pm 0.00$}
& $0.00 \pm 0.00$
& $0.98 \pm 0.00$
& $0.02 \pm 0.00$
& $0.02 \pm 0.00$ \\

& AMN
& \first{$1.00 \pm 0.00$}
& \first{$1.00 \pm 0.00$}
& $0.99 \pm 0.00$
& \third{$0.12 \pm 0.02$}
& $0.98 \pm 0.00$
& \third{$0.14 \pm 0.02$}
& \third{$0.31 \pm 0.03$} \\

& BADT
& $0.95 \pm 0.00$
& $0.03 \pm 0.01$
& \first{$1.00 \pm 0.00$}
& $0.01 \pm 0.00$
& $0.98 \pm 0.00$
& $0.05 \pm 0.00$
& $0.05 \pm 0.00$ \\

& EFFACE
& $0.99 \pm 0.00$
& \third{$0.97 \pm 0.02$}
& \first{$1.00 \pm 0.00$}
& \first{$0.58 \pm 0.01$}
& \first{$0.99 \pm 0.00$}
& \first{$0.59 \pm 0.01$}
& \first{$0.80 \pm 0.01$} \\

\bottomrule
\end{tabular}%
}
\end{table*}


From \Cref{tab:results_imagenet_clusters_f5_c50_a005_coverage}, we observe that all methods achieve high retained coverage above the $1-\alpha=0.95$ threshold. However, $\nabla\tau$ fails to cover the retained data or miscover the forget data in the in-sample case. The reason is that the quantile of $\calibset$ becomes $\hqalpha=1.00$, which renders all prediction sets to be full size (trivial sets). When we set $c=d=50 < 100 = |\calY|$, there are no points with such set sizes, and hence the coverage and miscoverage frequencies are both 0. 

BADT exhibits limited effectiveness in achieving high miscoverage levels on the forget subsets. This method relies on the Kullback-Leibler (KL) divergence for unlearning, which appears insufficient in the context of cluster-wise forgetting. The semantic overlap between forget and retain points in the feature space complicates the enforcement of distinct KL divergence values between these groups, thereby limiting the method's efficacy. In contrast, SCRUB incorporates an additional fine-tuning cross-entropy term over the retained data, supplementing the KL divergence-based objectives. This enhancement improves its performance relative to BADT, with out-sample results surpassing in-sample results—likely due to the KL divergence's greater effectiveness in distinguishing unseen data from training data. SSD, however, fails to demonstrate significant forgetting in both in-sample and out-sample scenarios, with miscoverage levels converging to $\beta \approx \alpha$ despite extensive hyperparameter tuning. This outcome may stem from the similarity of parameter importance scores between forget and retain data in the cluster forgetting case, which undermines the method's ability to differentiate between the two. AMN achieves near-perfect miscoverage on $\calD_f$ ($\approx 1.00$), but this performance does not generalize to $\calT_f$ ($\approx 0.12$) or $\calV_f$ ($\approx 0.14$). This overfitting behavior persists even when in-sample data are utilized for unlearning. Nevertheless, AMN consistently maintains retained coverage above the target threshold of $1-\alpha = 0.95$.

In contrast, EFFACE consistently satisfies coverage above $0.95$ on the retained subsets and high miscoverage on the forget subsets, achieving a substantial margin in $H$ over the next best method, especially in the in-sample case (difference in $H$ $\ge 0.39$). Moreover, it demonstrates consistent generalizability from $\forgetset$ to $\calT_f$ and $\calV_f$.

\Cref{tab:results_imagenet_classes_f5_c50_a005_coverage} presents the results for the class-wise forgetting scenario. All methods achieve better conformal unlearning performance compared to cluster-wise forgetting, likely due to the clearer separation between the forget and retain data (belonging to distinct classes) along the model's decision boundaries. Note that in this case $\nabla \tau$, SSD, and BADT suffer a big drop in unlearning performance when moving from the in-sample case to the out-sample case. EFFACE consistently demonstrates superior performance compared to all competing methods across all data subsets and in both in-sample and out-sample scenarios. The significant margin achieved by EFFACE ($\approx 0.19$) underscores its robustness and efficacy in simultaneously achieving high retained coverage and substantial forget miscoverage.


\begin{table*}[htbp] 
\caption{ImageNet100, RepVGG-a2 \emph{class}-wise forgetting with $c=d=50$, $\alpha=0.05$, and 5 forgotten classes. Coverage/miscoverage results.}
\label{tab:results_imagenet_classes_f5_c50_a005_coverage}
\centering
\resizebox{\linewidth}{!}{%
\begin{tabular}{llccccccc}
\toprule
Split & Method
& $\ECF_{c}[\retainset]\uparrow$
& $\EmCF_{d}[\forgetset]\uparrow$
& $\ECF_{c}[\calT_r]\uparrow$
& $\EmCF_{d}[\calT_f]\uparrow$
& $\ECF_{c}[\calV_r]\uparrow$
& $\EmCF_{d}[\calV_f]\uparrow$
& $H\uparrow$ \\
\midrule

& OR
& $1.00 \pm 0.00$
& $0.01 \pm 0.00$
& $1.00 \pm 0.00$
& $0.00 \pm 0.00$
& $0.98 \pm 0.00$
& $0.01 \pm 0.00$
& $0.01 \pm 0.00$ \\

\midrule

\multirow{7}{*}{In} 

& $\nabla\tau$
& $0.83 \pm 0.37$
& $0.46 \pm 0.24$
& $0.83 \pm 0.37$
& \cellcolor{green!15}\third{$0.40 \pm 0.22$}
& $0.83 \pm 0.37$
& $0.38 \pm 0.21$
& $0.55 \pm 0.12$ \\

& SCRUB
& $0.86 \pm 0.19$
& \third{$0.76 \pm 0.27$}
& $0.85 \pm 0.21$
& \second{$0.78 \pm 0.25$}
& $0.84 \pm 0.21$
& \second{$0.80 \pm 0.23$}
& \second{$0.81 \pm 0.10$} \\

& SSD
& $0.99 \pm 0.00$
& $0.47 \pm 0.03$
& \second{$0.99 \pm 0.00$}
& \cellcolor{green!15}$0.40 \pm 0.03$
& \second{$0.99 \pm 0.00$}
& \third{$0.39 \pm 0.02$}
& $0.59 \pm 0.02$ \\

& AMN
& \first{$1.00 \pm 0.00$}
& \first{$1.00 \pm 0.00$}
& \second{$0.99 \pm 0.00$}
& $0.37 \pm 0.01$
& \second{$0.99 \pm 0.00$}
& $0.36 \pm 0.02$
& \third{$0.63 \pm 0.01$} \\

& BADT
& $0.99 \pm 0.00$
& $0.13 \pm 0.01$
& \second{$0.99 \pm 0.00$}
& \cellcolor{green!15}$0.14 \pm 0.01$
& $0.98 \pm 0.00$
& $0.19 \pm 0.02$
& $0.26 \pm 0.01$ \\

& UNSIR
& \first{$1.00 \pm 0.00$}
& $0.13 \pm 0.01$
& \second{$0.99 \pm 0.00$}
& $0.14 \pm 0.01$
& $0.98 \pm 0.00$
& $0.14 \pm 0.01$
& $0.24 \pm 0.01$ \\

& EFFACE
& \first{$1.00 \pm 0.00$}
& \second{$0.99 \pm 0.00$}
& \first{$1.00 \pm 0.00$}
& \first{$1.00 \pm 0.00$}
& \first{$1.00 \pm 0.00$}
& \first{$1.00 \pm 0.00$}
& \first{$1.00 \pm 0.00$} \\

\midrule

\multirow{7}{*}{Out} 

& $\nabla\tau$
& \first{$1.00 \pm 0.00$}
& $0.14 \pm 0.05$
& $0.99 \pm 0.00$
& \cellcolor{gray!20}$0.04 \pm 0.02$
& $0.98 \pm 0.00$
& $0.05 \pm 0.02$
& $0.11 \pm 0.03$ \\

& SCRUB
& \first{$1.00 \pm 0.00$}
& \third{$0.97 \pm 0.03$}
& \first{$1.00 \pm 0.00$}
& \second{$0.65 \pm 0.01$}
& \second{$0.99 \pm 0.00$}
& \second{$0.70 \pm 0.01$}
& \second{$0.86 \pm 0.01$} \\

& SSD
& $0.96 \pm 0.00$
& $0.05 \pm 0.00$
& \first{$1.00 \pm 0.00$}
& \cellcolor{gray!20}$0.00 \pm 0.00$
& $0.98 \pm 0.00$
& $0.01 \pm 0.00$
& $0.02 \pm 0.00$ \\

& AMN
& \first{$1.00 \pm 0.00$}
& \first{$1.00 \pm 0.00$}
& $0.99 \pm 0.00$
& $0.14 \pm 0.02$
& $0.98 \pm 0.00$
& \third{$0.15 \pm 0.03$}
& \third{$0.33 \pm 0.03$} \\

& BADT
& $0.96 \pm 0.00$
& $0.12 \pm 0.01$
& \first{$1.00 \pm 0.00$}
& \cellcolor{gray!20}$0.02 \pm 0.00$
& $0.98 \pm 0.00$
& $0.03 \pm 0.00$
& $0.06 \pm 0.00$ \\

& UNSIR
& \first{$1.00 \pm 0.00$}
& $0.19 \pm 0.03$
& $0.99 \pm 0.00$
& \third{$0.15 \pm 0.01$}
& $0.98 \pm 0.00$
& $0.13 \pm 0.01$
& $0.26 \pm 0.01$ \\

& EFFACE
& $0.99 \pm 0.00$
& \first{$1.00 \pm 0.00$}
& \first{$1.00 \pm 0.00$}
& \first{$0.79 \pm 0.01$}
& \first{$1.00 \pm 0.00$}
& \first{$0.82 \pm 0.00$}
& \first{$0.92 \pm 0.00$} \\

\bottomrule
\end{tabular}%
}
\end{table*}



\begin{table*}[htbp] 
\caption{ImageNet100, RepVGG-a2 \emph{cluster}-wise forgetting with $c=d=50$, $\alpha=0.05$, and 5 forgotten clusters. Accuracy results in \%.}
\label{tab:results_imagenet_clusters_f5_c50_a005_accuracy}
\centering
\begin{tabular}{llcccccc}
\toprule
Split & Method
& $A(\calD_r)\uparrow$
& $A(\calD_f)\downarrow$
& $A(\calT_r)\uparrow$
& $A(\calT_f)\downarrow$
& $A(\calV_r)\uparrow$
& $A(\calV_f)\downarrow$ \\
\midrule

& OR
& $96.13 \pm 0.00$
& $94.97 \pm 0.00$
& $96.16 \pm 0.00$
& $93.32 \pm 0.00$
& $91.52 \pm 0.00$
& $88.35 \pm 0.00$ \\

\midrule

\multirow{6}{*}{In}

& $\nabla\tau$
& $90.57 \pm 7.09$
& $24.15 \pm 4.35$
& $81.91 \pm 5.59$
& \cellcolor{gray!20}$26.31 \pm 4.87$
& $79.76 \pm 5.11$
& $26.50 \pm 4.73$ \\

& SCRUB
& $97.21 \pm 2.42 $
& $43.45 \pm 12.73$
& $93.25 \pm 4.88 $
& $43.27 \pm 12.23$
& $89.36 \pm 3.80 $
& $42.07 \pm 10.54$ \\

& SSD
& $95.45 \pm 0.46$
& $87.40 \pm 6.10$
& $95.49 \pm 0.48$
& $84.92 \pm 7.21$
& $90.89 \pm 0.42$
& $80.50 \pm 6.42$ \\

& AMN
& $99.98 \pm 0.00$
& $0.00  \pm 0.00$
& $92.06 \pm 0.06$
& \cellcolor{gray!20}$28.15 \pm 1.39$
& $88.85 \pm 0.13$
& $27.02 \pm 1.36$ \\

& BADT
& $94.51 \pm 0.06$
& $50.40 \pm 4.07$
& $92.88 \pm 0.15$
& \cellcolor{gray!20}$52.83 \pm 4.59$
& $89.67 \pm 0.15$
& $50.00 \pm 3.83$ \\

& EFFACE
& $93.73 \pm 0.14$
& $1.37  \pm 0.09$
& $89.60 \pm 0.15$
& $5.61  \pm 0.35$
& $85.64 \pm 0.13$
& $6.63  \pm 0.39$ \\

\midrule

\multirow{6}{*}{Out}

& $\nabla\tau$
& $98.87 \pm 0.30$
& $72.33 \pm 2.59$
& $89.77 \pm 0.40$
& $86.38 \pm 0.80$
& $87.33 \pm 0.30$
& $81.96 \pm 0.76$ \\

& SCRUB
& $99.82 \pm 0.08$
& $0.31 \pm 0.70 $
& $91.84 \pm 0.15$
& $12.31 \pm 0.36$
& $88.03 \pm 0.25$
& $11.61 \pm 0.58$ \\

& SSD
& $87.33 \pm 0.00$
& $75.47 \pm 0.00$
& $96.16 \pm 0.00$
& $93.32 \pm 0.00$
& $91.53 \pm 0.01$
& $88.19 \pm 0.11$ \\

& AMN
& $100.00\pm 0.00$
& $0.00  \pm 0.00$
& $90.43 \pm 0.29$
& $47.52 \pm 2.75$
& $87.51 \pm 0.40$
& $47.65 \pm 1.75$ \\

& BADT
& $87.45 \pm 0.21$
& $65.41 \pm 2.09$
& $94.68 \pm 0.05$
& $82.23 \pm 0.55$
& $90.39 \pm 0.09$
& $77.51 \pm 0.55$ \\

& EFFACE
& $96.90 \pm 0.23$
& $2.20  \pm 1.69$
& $89.28 \pm 0.23$
& $17.42 \pm 1.50$
& $85.63 \pm 0.41$
& $17.35 \pm 1.25$ \\

\bottomrule
\end{tabular}%
\end{table*}



\begin{table*}[htpb]
\caption{ImageNet100, ResNet18 \emph{cluster}-wise forgetting with $c=d=100$, $\alpha=0.05$, and 5 forgotten clusters. Coverage/miscoverage results.}
\label{tab:EFFACE_vs_PABI_RT}
\centering
\begin{tabular}{llccccc}
\toprule
Split & Method
& $\ECF_{c}[\calT_r]\uparrow$
& $\EmCF_{d}[\calT_f]\uparrow$
& $\ECF_{c}[\calV_r]\uparrow$
& $\EmCF_{d}[\calV_f]\uparrow$
& $H\uparrow$ \\
\midrule

& RT
& $1.00 \pm 0.00$
& $0.09 \pm 0.01$
& $0.96 \pm 0.00$
& $0.09 \pm 0.01$
& $0.19 \pm 0.03$ \\

& PABI
& $1.00 \pm 0.00$
& $0.08 \pm 0.01$
& $0.96 \pm 0.00$
& $0.08 \pm 0.01$
& $0.15 \pm 0.01$ \\
\midrule

\multirow{1}{*}{In}

& EFFACE
& $1.00 \pm 0.00$
& $0.29 \pm 0.02$
& $0.97 \pm 0.00$
& $0.33 \pm 0.01$
& $0.56 \pm 0.01$ \\


\multirow{1}{*}{Out}

& EFFACE
& $1.00 \pm 0.00$
& $0.33 \pm 0.03$
& $0.97 \pm 0.00$
& $0.37 \pm 0.03$
& $0.61 \pm 0.02$ \\

\bottomrule
\end{tabular}%
\end{table*}


\subsubsection{Fake conformal unlearning}
\Cref{tab:results_imagenet_clusters_f5_c50_a005_accuracy} illustrates the phenomenon of \emph{fake conformal unlearning} (cf.\ \cref{sec:intro}), where several methods exhibit substantial accuracy degradation on the training-forgotten split $\calT_f$—for instance, $\nabla\tau$ (over $60\%$), AMN (over $60\%$), and BADT (over $50\%$)—yet fail to achieve the desired miscoverage rates on the same data, remaining below the target significance level $\alpha$. This discrepancy indicates that, despite the observed accuracy drop, the conformal prediction sets $\calC_{\theta_{u}} (X)$ frequently include the true label, even when constrained to small prediction set sizes ($c \leq 50$). Consequently, these samples are \emph{covered} rather than \emph{miscovered}, undermining the objective of effective unlearning.

The root cause of this inconsistency lies in the fundamentally different objectives of accuracy and conformal coverage. Accuracy penalizes any top-1 prediction error, whereas conformal coverage only requires the true label to be included within the prediction set, irrespective of its rank. As a result, methods that merely expand prediction sets—or fail to sufficiently reduce their size—can exhibit significant accuracy degradation without achieving genuine forgetting. For example, while the accuracy drop on $\calT_f$ is comparable between $\nabla\tau$ and EFFACE, their efficiency-aware miscoverage rates differ markedly ($0.00$ vs.\ $0.78$, respectively; cf.\ \cref{tab:results_imagenet_clusters_f5_c50_a005_coverage}). Notably, in \cref{tab:EFFACE_vs_PABI_RT}, both the retrained model (RT) and the certified unlearning method (PABI) also exhibit signs of fake conformal unlearning, as evidenced by their negligible miscoverage rates on forget data ($\beta \approx \alpha = 0.05$). These findings underscore the importance of adopting global, coverage-based criteria to enable uncertainty-aware evaluation.

These results reinforce the argument presented in \cref{sec:intro}: \textbf{accuracy alone is an insufficient metric for evaluating effective conformal unlearning} and may instead serve as an indicator of \emph{fake conformal unlearning}. In contrast, EFFACE demonstrates a consistent alignment between reductions in accuracy on forget data and corresponding increases in miscoverage, ensuring that true labels are systematically excluded from $\calC_{\theta_{u}} (X)$ at the specified prediction set size. Simultaneously, EFFACE maintains high retained coverage ($\geq 1-\alpha$), thereby satisfying the requirements of \cref{def:eff-conformalMU}. This alignment between coverage and miscoverage highlights that EFFACE does not indiscriminately degrade logits but instead strategically adjusts prediction sets to ensure that forgotten concepts are effectively unsupported, while retained concepts remain reliably covered.

When unlearning is performed using proxy out-sample data $\calD_f$, EFFACE effectively reduces coverage on $\calT_f$, as expected in the context of conformal unlearning. Since conformal unlearning targets data with shared characteristics, $\calD_f$ and $\calT_f$ are likely to occupy similar regions in the feature space. Consequently, shifting decision boundaries to miscover $\calD_f$ naturally impacts $\calT_f$. The observed increase in miscoverage, coupled with the corresponding drop in accuracy on both forgotten splits, demonstrates the intended effect of conformal unlearning.


\begin{table}[htbp]
\caption{ImageNet100, RepVGG-a2 out-sample \emph{cluster}-wise forgetting with $c=d=50$, $\alpha=0.05$, and 5 forgotten clusters. MIA scores (percentage) and time efficiency (seconds) results.}
\label{tab:results_imagenet_clusters_f5_c50_a005_miatime}
\centering
\begin{tabular}{lcc}
\toprule
Method
& MIA Diff.$\downarrow$
& Tsec$\downarrow$ \\
\midrule

$\nabla\tau$
& $0.06 \pm 0.05$
& $84.46 \pm 1.51$ \\

SCRUB
& $0.09 \pm 0.04$
& $98.18 \pm 0.32$ \\

SSD
& $0.04 \pm 0.02$
& $569.70 \pm 0.96$ \\

AMN
& $0.06 \pm 0.02$
& $299.85 \pm 2.51$ \\

BADT
& $0.09 \pm 0.04$
& $41.54 \pm 0.83$ \\

EFFACE
& $0.05 \pm 0.03$
& $340.64 \pm 1.12$ \\

\bottomrule
\end{tabular}
\end{table}


\subsubsection{MIA and unlearning time}
In \cref{tab:results_imagenet_clusters_f5_c50_a005_miatime}, an optimal MIA Diff value approaches zero, indicating robust privacy preservation. All methods exhibit low MIA scores, suggesting limited vulnerability to membership inference attacks. EFFACE achieves a similarly low MIA score, demonstrating its effectiveness against MIAs. Specifically, an adversary employing the MIA technique outlined in \cref{app:reproduce} would face significant difficulty in distinguishing between forgotten data points used during pretraining ($\calT_f$) and unseen data points ($\calV_f$).
In terms of unlearning time efficiency, EFFACE is at the slower side (but not the slowest) due to the quantile calculation step at each iteration which requires a full forward-pass of the data to find the scores and their quantile.



\begin{figure*}[htpb]
\centering

\begin{minipage}[c]{0.00\linewidth}
\hspace{\linewidth} 
\end{minipage}
\hfill
\begin{minipage}[t]{4cm}
\centering
\textbf{$\alpha=0.05,\; |\forgetset|=5$}
\end{minipage}
\hfill
\begin{minipage}[t]{4cm}
\centering
\textbf{$c=d=100,\; |\forgetset|=5$}
\end{minipage}
\hfill
\begin{minipage}[t]{4cm}
\centering
\textbf{$\alpha=0.05,\; c=d=100$}
\end{minipage}

\vspace{0.2cm}


\begin{minipage}[c]{0.03\linewidth}
\centering
{\rotatebox{90}{\textbf{In-Sample}}}
\end{minipage}
\hfill
\begin{minipage}[t]{0.30\linewidth}
\centering
\begin{tikzpicture}
\begin{axis}[
width=\linewidth, height=0.75\linewidth,
grid=both,
xlabel={$c,d$},
ylabel={$H$},
xmin=5, xmax=105,
ymin=0.0, ymax=1.0,
xtick={20,40,60,80,100},
ytick={0,0.2,0.4,0.6,0.8,1.0},
legend to name=sharedlegend,
legend columns=4,
legend style={font=\small, /tikz/every even column/.append style={column sep=0.3cm}},
]

\addplot+[magenta, thick, solid, mark=pentagon*, error bars/.cd, y dir=both, y explicit] table[row sep=\\, x=c, y=mean, y error=err] {
c   mean  err \\
10  0.26  0.00 \\
20  0.26  0.00 \\
40  0.26  0.00 \\
60  0.26  0.00 \\
80  0.26  0.00 \\
100 0.26  0.00 \\
};
\addlegendentry{BADT}

\addplot+[violet, thick, solid, mark=x, error bars/.cd, y dir=both, y explicit] table[row sep=\\, x=c, y=mean, y error=err] {
c   mean  err \\
10  0.24  0.00 \\
20  0.24  0.00 \\
40  0.24  0.00 \\
60  0.24  0.00 \\
80  0.24  0.00 \\
100 0.25  0.00 \\
};
\addlegendentry{UNSIR}

\addplot+[blue, thick, solid, mark=*, error bars/.cd, y dir=both, y explicit] table[row sep=\\, x=c, y=mean, y error=err] {
c   mean  err \\
10  0.57  0.00 \\
20  0.56  0.00 \\
40  0.55  0.00 \\
60  0.55  0.00 \\
80  0.55  0.00 \\
100 0.59  0.00 \\
};
\addlegendentry{$\nabla \tau$}

\addplot+[red, thick, dashed, mark=square*, error bars/.cd, y dir=both, y explicit] table[row sep=\\, x=c, y=mean, y error=err] {
c   mean  err \\
10  0.72  0.00 \\
20  0.76  0.00 \\
40  0.81  0.00 \\
60  0.82  0.00 \\
80  0.82  0.00 \\
100 0.64  0.00 \\
};
\addlegendentry{SCRUB}

\addplot+[orange, thick, dashdotdotted, mark=diamond*, error bars/.cd, y dir=both, y explicit] table[row sep=\\, x=c, y=mean, y error=err] {
c   mean  err \\
10  0.48  0.00 \\
20  0.57  0.00 \\
40  0.59  0.00 \\
60  0.58  0.00 \\
80  0.58  0.00 \\
100 0.59  0.00 \\
};
\addlegendentry{SSD}

\addplot+[green!60!black, thick, dotted, mark=triangle*, error bars/.cd, y dir=both, y explicit] table[row sep=\\, x=c, y=mean, y error=err] {
c   mean  err \\
10  0.80  0.00 \\
20  0.67  0.00 \\
40  0.63  0.00 \\
60  0.63  0.00 \\
80  0.63  0.00 \\
100 0.63  0.00 \\
};
\addlegendentry{AMN}

\addplot+[black, thick, solid, mark=star, error bars/.cd, y dir=both, y explicit] table[row sep=\\, x=c, y=mean, y error=err] {
c   mean  err \\
10  1.00  0.00 \\
20  1.00  0.00 \\
40  1.00  0.00 \\
60  1.00  0.00 \\
80  1.00  0.00 \\
100 0.97  0.00 \\
};
\addlegendentry{EFFACE}

\end{axis}
\end{tikzpicture}
\end{minipage}
\hfill
\begin{minipage}[t]{0.30\linewidth}
\centering
\begin{tikzpicture}
\begin{axis}[
width=\linewidth, height=0.75\linewidth,
grid=both,
xlabel={$\alpha$},
ylabel={$H$},
xmin=0.04, xmax=0.26,
ymin=0.0, ymax=1.0,
xtick={0.05,0.1,0.15,0.2,0.25},
ytick={0,0.2,0.4,0.6,0.8,1.0},
tick label style={/pgf/number format/fixed},
]

\addplot+[magenta, thick, solid, mark=pentagon*, error bars/.cd, y dir=both, y explicit] table[row sep=\\, x=a, y=mean, y error=err] {
a    mean  err \\
0.05 0.26  0.00 \\
0.10 0.66  0.00 \\
0.15 0.84  0.00 \\
0.20 0.92  0.00 \\
0.25 0.92  0.00 \\
};

\addplot+[violet, thick, solid, mark=x, error bars/.cd, y dir=both, y explicit] table[row sep=\\, x=a, y=mean, y error=err] {
a    mean  err \\
0.05 0.25  0.00 \\
0.10 0.43  0.00 \\
0.15 0.53  0.00 \\
0.20 0.60  0.00 \\
0.25 0.66  0.00 \\
};

\addplot+[blue, thick, solid, mark=*, error bars/.cd, y dir=both, y explicit] table[row sep=\\, x=a, y=mean, y error=err] {
a    mean  err \\
0.05 0.59  0.00 \\
0.10 0.72  0.00 \\
0.15 0.78  0.00 \\
0.20 0.81  0.00 \\
0.25 0.83  0.00 \\
};

\addplot+[red, thick, dashed, mark=square*, error bars/.cd, y dir=both, y explicit] table[row sep=\\, x=a, y=mean, y error=err] {
a    mean  err \\
0.05 0.64  0.00 \\
0.10 0.81  0.00 \\
0.15 0.87  0.00 \\
0.20 0.89  0.00 \\
0.25 0.89  0.00 \\
};

\addplot+[orange, thick, dashdotdotted, mark=diamond*, error bars/.cd, y dir=both, y explicit] table[row sep=\\, x=a, y=mean, y error=err] {
a    mean  err \\
0.05 0.59  0.00 \\
0.10 0.75  0.00 \\
0.15 0.78  0.00 \\
0.20 0.80  0.00 \\
0.25 0.80  0.00 \\
};

\addplot+[green!60!black, thick, dotted, mark=triangle*, error bars/.cd, y dir=both, y explicit] table[row sep=\\, x=a, y=mean, y error=err] {
a    mean  err \\
0.05 0.63  0.00 \\
0.10 0.85  0.00 \\
0.15 0.94  0.00 \\
0.20 0.96  0.00 \\
0.25 0.95  0.00 \\
};

\addplot+[black, thick, solid, mark=star, error bars/.cd, y dir=both, y explicit] table[row sep=\\, x=a, y=mean, y error=err] {
a    mean  err \\
0.05 0.96  0.00 \\
0.10 0.99  0.00 \\
0.15 0.99  0.00 \\
0.20 0.97  0.00 \\
0.25 0.95  0.00 \\
};

\end{axis}
\end{tikzpicture}
\end{minipage}
\hfill
\begin{minipage}[t]{0.30\linewidth}
\centering
\begin{tikzpicture}
\begin{axis}[
width=\linewidth, height=0.75\linewidth,
grid=both,
xlabel={$|\forgetset|$},
ylabel={$H$},
xmin=0, xmax=55,
ymin=0.0, ymax=1.0,
xtick={10,20,30,50},
ytick={0,0.2,0.4,0.6,0.8,1.0},
]

\addplot+[magenta, thick, solid, mark=pentagon*, error bars/.cd, y dir=both, y explicit] table[row sep=\\, x=f, y=mean, y error=err] {
f    mean  err \\
5    0.26  0.00 \\
10   0.20  0.00 \\
20   0.16  0.00 \\
30   0.13  0.00 \\
50   0.09  0.00 \\
};

\addplot+[violet, thick, solid, mark=x, error bars/.cd, y dir=both, y explicit] table[row sep=\\, x=f, y=mean, y error=err] {
f    mean  err \\
5    0.25  0.00 \\
10   0.17  0.00 \\
20   0.15  0.00 \\
30   0.13  0.00 \\
50   0.09  0.00 \\
};

\addplot+[blue, thick, solid, mark=*, error bars/.cd, y dir=both, y explicit] table[row sep=\\, x=f, y=mean, y error=err] {
f    mean  err \\
5    0.59  0.00 \\
10   0.36  0.00 \\
20   0.20  0.00 \\
30   0.15  0.00 \\
50   0.09  0.00 \\
};

\addplot+[red, thick, dashed, mark=square*, error bars/.cd, y dir=both, y explicit] table[row sep=\\, x=f, y=mean, y error=err] {
f    mean  err \\
5    0.64  0.00 \\
10   0.35  0.00 \\
20   0.18  0.00 \\
30   0.21  0.00 \\
50   0.14  0.00 \\
};

\addplot+[orange, thick, dashdotdotted, mark=diamond*, error bars/.cd, y dir=both, y explicit] table[row sep=\\, x=f, y=mean, y error=err] {
f    mean  err \\
5    0.59  0.00 \\
10   0.34  0.00 \\
20   0.01  0.00 \\
30   0.01  0.00 \\
50   0.01  0.00 \\
};

\addplot+[green!60!black, thick, dotted, mark=triangle*, error bars/.cd, y dir=both, y explicit] table[row sep=\\, x=f, y=mean, y error=err] {
f    mean  err \\
5    0.63  0.00 \\
10   0.45  0.00 \\
20   0.36  0.00 \\
30   0.29  0.00 \\
50   0.20  0.00 \\
};

\addplot+[black, thick, solid, mark=star, error bars/.cd, y dir=both, y explicit] table[row sep=\\, x=f, y=mean, y error=err] {
f    mean  err \\
5    0.97  0.00 \\
10   0.63  0.00 \\
20   0.38  0.00 \\
30   0.25  0.00 \\
50   0.14  0.00 \\
};

\end{axis}
\end{tikzpicture}
\end{minipage}



\begin{minipage}[c]{0.03\linewidth}
\centering
{\rotatebox{90}{\textbf{Out-Sample}}}
\end{minipage}
\hfill
\begin{minipage}[t]{0.30\linewidth}
\centering
\begin{tikzpicture}
\begin{axis}[
width=\linewidth, height=0.75\linewidth,
grid=both,
xlabel={$c,d$},
ylabel={$H$},
xmin=5, xmax=105,
ymin=0.0, ymax=1.0,
xtick={20,40,60,80,100},
ytick={0,0.2,0.4,0.6,0.8,1.0},
]

\addplot+[magenta, thick, solid, mark=pentagon*, error bars/.cd, y dir=both, y explicit] table[row sep=\\, x=c, y=mean, y error=err] {
c   mean  err \\
10  0.06  0.00 \\
20  0.06  0.00 \\
40  0.06  0.00 \\
60  0.06  0.00 \\
80  0.06  0.00 \\
100 0.06  0.00 \\
};

\addplot+[violet, thick, solid, mark=x, error bars/.cd, y dir=both, y explicit] table[row sep=\\, x=c, y=mean, y error=err] {
c   mean  err \\
10  0.27  0.00 \\
20  0.26  0.00 \\
40  0.26  0.00 \\
60  0.26  0.00 \\
80  0.26  0.00 \\
100 0.27  0.00 \\
};

\addplot+[blue, thick, solid, mark=*, error bars/.cd, y dir=both, y explicit] table[row sep=\\, x=c, y=mean, y error=err] {
c   mean  err \\
10  0.11  0.00 \\
20  0.11  0.00 \\
40  0.11  0.00 \\
60  0.11  0.00 \\
80  0.11  0.00 \\
100 0.12  0.00 \\
};

\addplot+[red, thick, dashed, mark=square*, error bars/.cd, y dir=both, y explicit] table[row sep=\\, x=c, y=mean, y error=err] {
c   mean  err \\
10  0.79  0.00 \\
20  0.82  0.00 \\
40  0.85  0.00 \\
60  0.86  0.00 \\
80  0.85  0.00 \\
100 0.85  0.00 \\
};

\addplot+[orange, thick, dashdotdotted, mark=diamond*, error bars/.cd, y dir=both, y explicit] table[row sep=\\, x=c, y=mean, y error=err] {
c   mean  err \\
10  0.02  0.00 \\
20  0.02  0.00 \\
40  0.02  0.00 \\
60  0.02  0.00 \\
80  0.02  0.00 \\
100 0.02  0.00 \\
};

\addplot+[green!60!black, thick, dotted, mark=triangle*, error bars/.cd, y dir=both, y explicit] table[row sep=\\, x=c, y=mean, y error=err] {
c   mean  err \\
10  0.34  0.00 \\
20  0.34  0.00 \\
40  0.33  0.00 \\
60  0.33  0.00 \\
80  0.33  0.00 \\
100 0.33  0.00 \\
};

\addplot+[black, thick, solid, mark=star, error bars/.cd, y dir=both, y explicit] table[row sep=\\, x=c, y=mean, y error=err] {
c   mean  err \\
10  0.95  0.00 \\
20  0.93  0.00 \\
40  0.93  0.00 \\
60  0.92  0.00 \\
80  0.92  0.00 \\
100 0.92  0.00 \\
};

\end{axis}
\end{tikzpicture}
\end{minipage}
\hfill
\begin{minipage}[t]{0.30\linewidth}
\centering
\begin{tikzpicture}
\begin{axis}[
width=\linewidth, height=0.75\linewidth,
grid=both,
xlabel={$\alpha$},
ylabel={$H$},
xmin=0.04, xmax=0.26,
ymin=0.0, ymax=1.0,
xtick={0.05,0.1,0.15,0.2,0.25},
ytick={0,0.2,0.4,0.6,0.8,1.0},
tick label style={/pgf/number format/fixed},
]

\addplot+[magenta, thick, solid, mark=pentagon*, error bars/.cd, y dir=both, y explicit] table[row sep=\\, x=a, y=mean, y error=err] {
a    mean  err \\
0.05 0.06  0.00 \\
0.10 0.20  0.00 \\
0.15 0.37  0.00 \\
0.20 0.52  0.00 \\
0.25 0.62  0.00 \\
};

\addplot+[violet, thick, solid, mark=x, error bars/.cd, y dir=both, y explicit] table[row sep=\\, x=a, y=mean, y error=err] {
a    mean  err \\
0.05 0.27  0.00 \\
0.10 0.41  0.00 \\
0.15 0.53  0.00 \\
0.20 0.63  0.00 \\
0.25 0.70  0.00 \\
};

\addplot+[blue, thick, solid, mark=*, error bars/.cd, y dir=both, y explicit] table[row sep=\\, x=a, y=mean, y error=err] {
a    mean  err \\
0.05 0.12  0.00 \\
0.10 0.27  0.00 \\
0.15 0.38  0.00 \\
0.20 0.46  0.00 \\
0.25 0.53  0.00 \\
};

\addplot+[red, thick, dashed, mark=square*, error bars/.cd, y dir=both, y explicit] table[row sep=\\, x=a, y=mean, y error=err] {
a    mean  err \\
0.05 0.85  0.00 \\
0.10 0.91  0.00 \\
0.15 0.93  0.00 \\
0.20 0.93  0.00 \\
0.25 0.92  0.00 \\
};

\addplot+[orange, thick, dashdotdotted, mark=diamond*, error bars/.cd, y dir=both, y explicit] table[row sep=\\, x=a, y=mean, y error=err] {
a    mean  err \\
0.05 0.02  0.00 \\
0.10 0.05  0.00 \\
0.15 0.10  0.00 \\
0.20 0.16  0.00 \\
0.25 0.25  0.00 \\
};

\addplot+[green!60!black, thick, dotted, mark=triangle*, error bars/.cd, y dir=both, y explicit] table[row sep=\\, x=a, y=mean, y error=err] {
a    mean  err \\
0.05 0.33  0.00 \\
0.10 0.61  0.00 \\
0.15 0.76  0.00 \\
0.20 0.84  0.00 \\
0.25 0.88  0.00 \\
};

\addplot+[black, thick, solid, mark=star, error bars/.cd, y dir=both, y explicit] table[row sep=\\, x=a, y=mean, y error=err] {
a    mean  err \\
0.05 0.92  0.00 \\
0.10 0.94  0.00 \\
0.15 0.95  0.00 \\
0.20 0.94  0.00 \\
0.25 0.93  0.00 \\
};

\end{axis}
\end{tikzpicture}
\end{minipage}
\hfill
\begin{minipage}[t]{0.30\linewidth}
\centering
\begin{tikzpicture}
\begin{axis}[
width=\linewidth, height=0.75\linewidth,
grid=both,
xlabel={$|\forgetset|$},
ylabel={$H$},
xmin=0, xmax=55,
ymin=0.0, ymax=1.0,
xtick={10,20,30,50},
ytick={0,0.2,0.4,0.6,0.8,1.0},
]

\addplot+[magenta, thick, solid, mark=pentagon*, error bars/.cd, y dir=both, y explicit] table[row sep=\\, x=f, y=mean, y error=err] {
f    mean  err \\
5    0.06  0.00 \\
10   0.05  0.00 \\
20   0.09  0.00 \\
30   0.08  0.00 \\
50   0.10  0.00 \\
};

\addplot+[violet, thick, solid, mark=x, error bars/.cd, y dir=both, y explicit] table[row sep=\\, x=f, y=mean, y error=err] {
f    mean  err \\
5    0.27  0.00 \\
10   0.18  0.00 \\
20   0.08  0.00 \\
30   0.11  0.00 \\
50   0.10  0.00 \\
};

\addplot+[blue, thick, solid, mark=*, error bars/.cd, y dir=both, y explicit] table[row sep=\\, x=f, y=mean, y error=err] {
f    mean  err \\
5    0.12  0.00 \\
10   0.10  0.00 \\
20   0.09  0.00 \\
30   0.10  0.00 \\
50   0.07  0.00 \\
};

\addplot+[red, thick, dashed, mark=square*, error bars/.cd, y dir=both, y explicit] table[row sep=\\, x=f, y=mean, y error=err] {
f    mean  err \\
5    0.85  0.00 \\
10   0.62  0.00 \\
20   0.22  0.00 \\
30   0.28  0.00 \\
50   0.07  0.00 \\
};

\addplot+[orange, thick, dashdotdotted, mark=diamond*, error bars/.cd, y dir=both, y explicit] table[row sep=\\, x=f, y=mean, y error=err] {
f    mean  err \\
5    0.42  0.00 \\
10   0.01  0.00 \\
20   0.01  0.00 \\
30   0.02  0.00 \\
50   0.02  0.00 \\
};

\addplot+[green!60!black, thick, dotted, mark=triangle*, error bars/.cd, y dir=both, y explicit] table[row sep=\\, x=f, y=mean, y error=err] {
f    mean  err \\
5    0.33  0.00 \\
10   0.25  0.00 \\
20   0.22  0.00 \\
30   0.16  0.00 \\
50   0.11  0.00 \\
};

\addplot+[black, thick, solid, mark=star, error bars/.cd, y dir=both, y explicit] table[row sep=\\, x=f, y=mean, y error=err] {
f    mean  err \\
5    0.92  0.00 \\
10   0.64  0.00 \\
20   0.34  0.00 \\
30   0.24  0.00 \\
50   0.15  0.00 \\
};

\end{axis}
\end{tikzpicture}
\end{minipage}

\begin{minipage}{\linewidth}\centering
\pgfplotslegendfromname{sharedlegend}
\end{minipage}

\caption{ImageNet100: 5 classes forgetting. (Top): In-sample results. (Bottom): Out-sample results. (Left): $H$ vs. $c=d$. (Middle): $H$ vs. $\alpha$. (Right): $H$ vs. $|\forgetset|$.}
\label{fig:H_vs_c_alpha_imagenet_class}

\end{figure*}



\subsubsection{Sensitivity analysis}
In \cref{fig:H_vs_c_alpha_imagenet_class}, we examine the effects of varying each parameter while holding others fixed: the critical set sizes $c,d$ (left), the miscoverage tolerance $\alpha$ (middle), and the number of forgotten classes $|\forgetset|$ (right). When we vary the number of forgotten classes, the size of the forgotten set increases, which relatively increases $\pi_f$. By the bound in \cref{lem:alpha-beta-tradeoff}, the maximum possible value of $\beta$ decreases accordingly. That justifies the drop in $H$ as the size of $|\forgetset|$ increases, which holds for all methods. When $\alpha$ increases, the conformal predictor has more tolerance of having miscovered points while still marginally satisfying \cref{eq:quantile-coverage}. Meanwhile, the larger $\alpha$ is, the looser the bound on $\beta$ becomes by \cref{lem:alpha-beta-tradeoff}. Therefore, the level of miscoverage on the forget sets increases, hence $H$ increases. All methods show the same tendancy. The effect of the critical set sizes $c$ and $d$ is more subtle and depends on both the model and the unlearning method. When the original model performs very well—such as RepVGG-A2 on ImageNet100, which achieves 90+ accuracy on test data—most methods produce substantially small prediction sets (average size $\leq 10$). In such cases, varying $c$ and $d$ has little effect on coverage and miscoverage, since the metrics at the full set size already capture the total coverage and miscoverage levels. EFFACE produces larger prediction sets, particularly on the forget data ($\approx 30$), but since its $H$ is already very high (0.95+), the metrics remain stable across different thresholds. However, this stability does not hold universally. For instance, in \cref{tab:results_imagenet_clusters_f5_c50_a005_coverage}, setting $c = d = 50$ reveals that $\nabla\tau$ collapses in terms of coverage and miscoverage, exposing deficiencies that would remain hidden if only full set sizes were tested. This underscores the importance of thorough conformal unlearning evaluation: in practice, unlearners should test across a range of $c$ and $d$ values. We note that EFFACE consistently outperforms other methods when $\alpha$ is small (e.g., $0.05$), which is the regime of practical interest and the most common setting in the literature.


\begin{table}[htpb]
\centering
\caption{ImagenNet100, ResNet18 10 clusters forgetting with $c,d=100$ and $\alpha=0.05$.}
\label{tab:EFFACE_vs_CPU}
\begin{tabular}{lcc}
\toprule
Metric & CPU & EFFACE \\
\midrule
$\ECF_{c}[\calT_r]\uparrow$   & 1.00 $\pm$ 0.00
        & 1.00 $\pm$ 0.00 \\
$\EmCF_{d}[\calT_f]\uparrow$  & 0.02 $\pm$ 0.00 
        & 0.17 $\pm$ 0.01 \\
$\ECF_{c}[\calV_r]\uparrow$   & 0.96 $\pm$ 0.00
        & 0.98 $\pm$ 0.00\\
$\EmCF_{d}[\calV_f]\uparrow$  & 0.07 $\pm$ 0.01 
        & 0.27 $\pm$ 0.02\\
\addlinespace
$\CR(\calV_r)$                & 0.04 $\pm$ 0.00 
        & 0.03 $\pm$ 0.00 \\
$\CR(\calV_f)$                & 0.02 $\pm$ 0.00 
        & 0.01 $\pm$ 0.00 \\
\addlinespace
$H\uparrow$                   & 0.06 $\pm$ 0.01 
        & 0.34 $\pm$ 0.02 \\
\bottomrule
\end{tabular}%
\end{table}

\subsubsection{Comparison with CR}
\Cref{tab:EFFACE_vs_CPU} compares EFFACE with the CPU (fine-tuning) variant \cite{shi2025rethinking} and the CR metric (cf.\ \cref{CR}). EFFACE consistently meets the desired $\ECF$/$\EmCF$ targets, achieving a significant $H$ improvement ($\ge 0.28$) over CPU. In contrast, the CR metric can be misleading: smaller prediction sets reduce the denominator, potentially inflating $\CR(\calV_f)$ even when true labels are frequently covered (e.g., EFFACE's $\EmCF(\calV_f) = 0.27 > 0.07 =$ CPU's $\EmCF(\calV_f)$). 
Similarly, $\CR(\calV_r)$ may appear disproportionately small in many-class settings despite high retained coverage (e.g., $\ECF(\calV_r) \approx 0.98$). 
These results verify that \emph{CR does not reliably address fake conformal unlearning}, particularly in scenarios with large label spaces (cf.\  \cref{sec:related-work} for a detailed discussion).

We refer the reader to more numerical results in \cref{app:extra-empirics}, including results on CIFAR100 and  20NewsGroups and targeted \emph{class}-wise forgetting. We discuss the limitations of our framework in \cref{app:limitation} and conduct a sensitivity analysis in \cref{app:sensitivity}.


\section{Different Conformity Score Function During Inference}\label{sec:different-scores}

EFFACE employs a CP procedure $\CP(\theta, s)$ to facilitate unlearning. However, a downstream practitioner may opt to utilize an alternative nonconformity score function $s'$. This raises a pertinent question: does achieving $(\alpha,\beta)$-conformal unlearning with respect to $s$ provide any guarantees when evaluated under $s'$?

Throughout this section, we assume that there exists an unlearning algorithm that achieves $(\alpha,\beta)$-conformal unlearning with respect to $s$. We provide sufficient conditions under which the coverage and miscoverage guarantees exhibit controlled degradation when using $s'$. Specifically, if $\Ualgo$ satisfies $(\alpha,\beta)$-conformal unlearning with respect to $s$, then it also satisfies $(\alpha',\beta')$-conformal unlearning with respect to $s'$, where $\alpha'$ and $\beta'$ remain close to $\alpha$ and $\beta$, respectively. We provide explicit bounds to quantify the extent of this degradation. We denote the \gls{CP} prediction set constructed using $s'$ as $\calC_{\theta_u}'(\cdot)$ and the $\ceil{(1-\alpha)(n+1)}/n$ quantile of the $s'$ scores on $\calibset$ as $\hqalpha'$. To avoid clutter, we write $s$ and $s'$ instead of $s(X,Y)$ and $s'(X,Y)$ for $(X,Y)\sim \pdata$. For a score function $s$, let $F_s(\cdot) = \P(s(X,Y) \leq \cdot)$ denote its \gls{cdf}, $F_{r,s}(\cdot ) = \Pret(s(X,Y) \leq \cdot)$, and $F_{f,s}(\cdot ) = \Pfor(s(X,Y) \leq \cdot)$.

\begin{Lemma}\label{lem:strictly-increasing-s}
Suppose $s'=g(s)$, for a strictly increasing function $g(\cdot)$. Then, $\alpha'=\alpha$, and $\beta'=\beta$. I.e., conformal unlearning guarantees are invariant under strictly increasing transformations of the conformity score.
\end{Lemma}
\begin{proof}
Using the score function $s'$, the coverage probability on $\retainset$ is
\begin{align*}
\Pret(Y \in \calC_{\theta_{u}}'(X)) &= \Pret(s'(X,Y) \le \hqalpha' ) \\
&=\Pret(g(s(X,Y)) \le \hat q'_{\alpha} )\\ 
&= \Pret(g(s(X,Y)) \le g(\hqalpha) )\\
&= \Pret(s(X,Y) \le \hqalpha )\\
&= \Pret(Y \in \calC_{\theta_{u}}(X)) \ge 1-\alpha,
\end{align*}
where the third and fourth equalities follow since $g(\cdot)$ is strictly increasing.
A similar proof yields 
\begin{align*}
    \Pfor(Y \notin \calC_{\theta_{u}}'(X)) \ge \beta.
\end{align*}
\end{proof}

The sufficient condition in \cref{lem:strictly-increasing-s} is distribution-free but limited to strictly monotonic transformations. To handle more general perturbations, we impose distributional assumptions. The following propositions quantify how bounded deviations between $s$ and $s'$ affect the coverage guarantees.

\begin{Proposition}\label{prop:bounded-point-wise-s}
Suppose $|s'-g(s)| \le c\ \as$, where $g(\cdot)$ is a strictly increasing function and $c \ge 0$ is a constant. In addition, suppose that $F_{s'}(\cdot)$ is $L$-Lipschitz. Then,
\begin{align}
\alpha' &\le \alpha + \frac{2Lc}{\pi_r},\\
\beta' &\ge \beta - \frac{2Lc}{\pi_f}.
\end{align}
\end{Proposition}
\begin{proof}
Define a new score function $s'' = g(s)$. By \cref{lem:strictly-increasing-s}, the unlearning guarantees for $s''$ remain valid with the same $\alpha$ and $\beta$. Given the point-wise bounded difference between $s''$ and $s'$, we have $F_{s''}(\sigma-c) \le F_{s'}(\sigma) \le F_{s''}(\sigma+c)$ for all $\sigma$. Moreover, it follows that $\hqalpha' - c \le \hqalpha'' \le \hqalpha' + c\ \as$. With the expectation taken over $\calibset$, we have 
\begin{align*}
&\E F_{r,s'}(\hqalpha' + 2c ) - \E F_{r,s'}(\hqalpha' ) \\
&= \Pret(\hqalpha' < s' \le \hqalpha' + 2c) \\ 
&\le \frac{\P(\hqalpha' < s' \le \hqalpha' + 2c)}{\pi_r} \\
&= \frac{\E[ F_{s'}(\hqalpha' + 2c) - F_{s'}(\hqalpha') ]}{\pi_r} \\
&\le \frac{2Lc}{\pi_r},
\end{align*}
where the last inequality follows from the $L$-Lipschitz assumption on $F_{s'}$.
Therefore,
\begin{align*}
\E F_{r,s'}(\hqalpha') 
&\geq \E F_{r,s'}(\hqalpha' + 2c ) - \frac{2Lc}{\pi_r} \\
&= \Pret(s' \le \hqalpha' + 2c) - \frac{2Lc}{\pi_r} \\
&\geq \Pret(s'' \le \hqalpha' + c) - \frac{2Lc}{\pi_r} \\
&\geq \Pret(s'' \le \hqalpha'') - \frac{2Lc}{\pi_r}  \\
&\geq 1 - \alpha - \frac{2Lc}{\pi_r},
\end{align*}
where the last inequality follows from the conformal coverage guarantee for $s''$ (i.e., \cref{eq:conformal-coverage-retain} and \cref{lem:strictly-increasing-s}).

Similarly, we have
\begin{align*}
& \E F_{f,s'}(\hqalpha' ) - \E F_{f,s'}(\hqalpha' - 2c ) 
\le \frac{2Lc}{\pi_f}
\end{align*}
and 
\begin{align*}
\E F_{f,s'}(\hqalpha' ) 
&\leq \E F_{f,s'}(\hqalpha' - 2c ) + \frac{2Lc}{\pi_f} \\
&= \Pfor(s' \le \hqalpha' - 2c) + \frac{2Lc}{\pi_f} \\
&\leq \Pfor(s'' \le \hqalpha' - c) + \frac{2Lc}{\pi_f} \\
&\leq \Pfor(s'' \le \hqalpha'') + \frac{2Lc}{\pi_f} \\
&\leq 1 - \parens*{\beta - \frac{2Lc}{\pi_f}},
\end{align*}
where the last inequality follows from the conformal miscoverage guarantee for $s''$ (i.e., \cref{eq:conformal-coverage-forget} and \cref{lem:strictly-increasing-s}).
The proof is now complete.
\end{proof}


\begin{Proposition}\label{prop:bounded-exp-s}
Suppose the following:
\begin{enumerate}[i)]
\item $F_{\ell,s}(\cdot)$ is $L$-Lipschitz for $\ell\in\set{r,f}$.
\item $\E\abs{s'-s} \leq c$, for a constant $c \ge 0$.
\item $\E\abs{s} \le m$, for a constant $m \ge 0$.
\end{enumerate}
Let $n = |\calibset|$. Then,
\begin{align}
\alpha' &\le \alpha + 2\sqrt{Lc} + L(c + 2m), \label{bxs-alpha}\\
\beta' &\ge \beta - 2\sqrt{Lc} - L(c + 2m). \label{bxs-beta}
\end{align}
\end{Proposition}

To prove \cref{prop:bounded-exp-s}, we first show a preliminary lemma.

\begin{Lemma}\label{lem:cdf-difference}
Under the assumptions of \cref{prop:bounded-exp-s}, we have for any $t$ and $\epsilon > 0$,
\begin{align}
F_{\ell,s}(t-\epsilon) - \frac{c}{\epsilon} \leq F_{\ell,s'}(t) \leq F_{\ell,s}(t+\epsilon) + \frac{c}{\epsilon},
\end{align}
for $\ell \in \set{r,f}$.
\end{Lemma}
\begin{proof}
We have for any $t$ and $\epsilon > 0$,
\begin{align*}
F_{\ell,s'}(t) 
&= \P_{\ell}(s' \le t)\\
&\le \P_{\ell}(s \le t + \epsilon) + \P_{\ell}(|s'-s| > \epsilon)\\
&\le F_{\ell,s}(t + \epsilon) + \frac{c}{\epsilon},
\end{align*}
where the last inequality follows from the Markov inequality.
A similar proof yields
\begin{align*}
F_{\ell,s'}(t) \ge F_{\ell,s}(t - \epsilon) - \frac{c}{\epsilon},
\end{align*}
and the proof is complete.
\end{proof}

We are now ready to prove \cref{prop:bounded-exp-s}.
\begin{proof}[Proof of \cref{prop:bounded-exp-s}]
We prove the bound \cref{bxs-alpha}; the proof of \cref{bxs-beta} is similar. For any $\epsilon > 0$,
\begin{align*}
&\Pret(s' \leq \hqalpha') \\
&= \E \Pret(s' \leq \hqalpha' \given \calibset) \\
&\geq \E \Pret(s \leq \hqalpha' - \epsilon \given \calibset) - \frac{c}{\epsilon}, \\
&= \E[F_{r,s}(\hqalpha' - \epsilon) - F_{r,s}(\hqalpha) + F_{r,s}(\hqalpha) ] - \frac{c}{\epsilon}, \\
&\geq -L (\epsilon + \E\abs*{\hqalpha' - \hqalpha}) + \E F_{r,s}(\hqalpha) - \frac{c}{\epsilon} \\
&\geq -L (\epsilon + c + 2\E|s|) + \E F_{r,s}(\hqalpha) - \frac{c}{\epsilon}\\
&\geq 1 - (\alpha + L(\epsilon + c + 2m) + \frac{c}{\epsilon}),
\end{align*}
where the first inequality follows from \cref{lem:cdf-difference}, and the last inequality follows from the conformal coverage guarantee for $s$ (i.e., \cref{eq:conformal-coverage-retain}). Maximizing the right-hand side over $\epsilon > 0$ gives the desired bound \cref{bxs-alpha}, and the proof is complete.
\end{proof}

By leveraging \cref{lem:strictly-increasing-s}, we observe that the conclusions of \cref{prop:bounded-exp-s} remain valid when $s$ is substituted with $g(s)$, provided that $g(\cdot)$ is a strictly increasing function with bounded derivatives. The results in this section demonstrate that $(\alpha,\beta)$-conformal unlearning exhibits robustness to bounded perturbations in the conformity score function. Specifically, when an alternative score $s'$ is employed during inference and is statistically close to the original score $s$, the coverage and miscoverage guarantees degrade in a controlled manner. Consequently, the unlearning guarantees retain their statistical significance and interpretability, even when different conformity scores are utilized in downstream applications.


\section{Conformal Unlearning Beyond Exchangeability}\label{sec:non-exch}

The work \cite{barber2023conformal} provides a framework to relax the exchangeability requirement and still obtain meaningful coverage bounds for \gls{CP}. Specifically, when split \gls{CP} is applied to a calibration set $\calibset$ that is \emph{not} exchangeable with $\testset$, let $\calibsize = n$ and $\calibset = \{Z_1, \dots, Z_n\}$. The coverage bound becomes
\begin{align}
\P(Y \in \calC_{\theta} (X)) \ge 1- \alpha - \sum_{i=1}^{n}\widetilde \omega_i \cdot d_{TV}\big(s(\boldsymbol{Z}), s(\boldsymbol{Z^i})\big),
\end{align}
where $\widetilde \omega_i = \frac{\omega_i}{\omega_1 + \dots + \omega_n + 1}$ for a set of user-specified weights $\{\omega_i\}_{i=1}^{n}$, $d_{TV}(P,Q)$ denotes the total variation distance between distributions $P$ and $Q$, $s(\cdot)$ is the nonconformity score function, $\boldsymbol{Z} = \{Z_1, \dots, Z_n, Z\}$, and $\boldsymbol{Z^i}$ is the set $\boldsymbol{Z}$ with the $i$th entry swapped with $Z$, i.e., $\boldsymbol{Z^i} = \{Z_1, \dots, Z_{i-1}, Z, Z_{i+1}, \dots, Z_n, Z_i\}$. Intuitively, the weights $\omega_i$ can be chosen so that the samples $Z_i$ more similar to the test sample $Z$ receive higher weight. For further details on non-exchangeable \gls{CP}, see \cite{barber2023conformal}.

The primary concern in this context is the impact of the gap correction on our results, particularly in \cref{lem:conformal-coverage-retain-holds}. To address this, we define the correction gap for a test sample $Z$ as $g(Z)$. Under the assumption of non-exchangeability, the coverage guarantee is adjusted as follows:
\begin{align}\label{eq:quantile-nonexchangeable-coverage}
\P(Y \in \calC_{\theta} (X)) \ge 1 - \alpha - g(Z).
\end{align}

It follows that for \cref{lem:conformal-coverage-retain-holds} to hold, we have to include the coverage gap into \cref{eq:conformal-coverage-retain} and \cref{eq:conformal-coverage-forget}.
\begin{Proposition}\label{prop:coverage-gap-included}
In the conformal unlearning scenario, suppose $\pi_r > 0$. Moreover, suppose $g(Z) > 0$ (otherwise $\ulset$ and $\calibset$ will be exchangeable). Then, \cref{eq:conformal-coverage-forget} implies $\cref{eq:conformal-coverage-retain}$ if $\beta' \ge \alpha' \ge \alpha + g(Z)$, where $\alpha'$ and $\beta'$ are to replace $\alpha$ and $\beta$ in \cref{eq:conformal-coverage-retain} and \cref{eq:conformal-coverage-forget}, respectively.
\end{Proposition}
\begin{proof}\label{proof:proof-of-proposition2}
From \cref{eq:quantile-nonexchangeable-coverage}, we have that $\P(Y \in \calC_{\theta_{u}} (X)) \ge 1 - \alpha - g(Z)$. From \cref{eq:conformal-coverage-forget}, we obtain 
\begin{align*}
\Pfor(Y \in \calC_{\theta_{u}} (X) ) & \le 1-\beta' \le 1- \alpha'. 
\end{align*}
Suppose \cref{eq:conformal-coverage-retain} does not hold (i.e., $\Pret(Y \in \calC_{\theta_{u}} (X) ) < 1-\alpha'$). Then, 
\begin{align*}
\P(Y \in \calC_{\theta_{u}} (X)) 
&=\pi_r \Pret(Y \in \calC_{\theta_{u}} (X) )
+\,\pi_f \Pfor(Y \in \calC_{\theta_{u}} (X) )\nn
&< 1 - \alpha',
\end{align*}
a contradiction if $\alpha' \ge \alpha + g(Z)$. Since $\beta' \ge \alpha'$ by definition, the proposition holds. 
\end{proof}

Even in scenarios where exchangeability is difficult to assume, conformal unlearning remains a robust framework, provided that minor corrections are incorporated into the coverage and miscoverage bounds. This robustness ensures that the framework can accommodate practical situations where strict exchangeability is not guaranteed, while still delivering meaningful statistical guarantees. As demonstrated in \cref{prop:coverage-gap-included}, the inclusion of a coverage gap correction enables the adaptation of conformal unlearning to non-exchangeable settings, thereby extending its applicability to real-world data distributions.

The lack of exchangeability often arises due to various types of distributional shifts. For instance, \cite{podkopaev2021label} address shifts in label distributions, \cite{tibshirani2019covariateshift} focus on covariate shifts in the input data, and \cite{chernozhukov2018timeseries} examine dependencies in time-series data. In general, as discussed earlier, the challenge lies in appropriately selecting weights to account for these distributional shifts. These weights are used to adjust the conformity scores of $\calibset$ when determining the $(1-\alpha)$-th quantile. Developing practical conformal unlearning methods that effectively handle diverse types of distributional shifts represents an important avenue for future research, which is beyond the scope of this paper.

\begin{figure}[htpb]
\centering
\begin{minipage}[t]{0.49\linewidth}
\centering
\begin{tikzpicture}
\begin{axis}[
width=\linewidth, height=0.7\linewidth,
grid=both,
xlabel={mean shift},
ylabel={value},
title={std scaling $=2$},
xmin=0.05, xmax=0.75,
ymin=0.0, ymax=1.05,
xtick={0.1,0.3,0.5,0.7},
tick label style={/pgf/number format/fixed},
clip=false,
legend to name=sharedlegend1,
legend columns=3,
legend style={font=\small, /tikz/every even column/.append style={column sep=0.6cm}},
]

\addplot+[
blue, thick, solid, mark=*,
error bars/.cd, y dir=both, y explicit
] table[row sep=\\, x=m, y=mean, y error=err] {
m     mean  err \\
0.1   0.59  0.00 \\
0.3   0.62  0.00 \\
0.5   0.61  0.00 \\
0.7   0.61  0.00 \\
};
\addlegendentry{$H$}

\addplot+[
red, thick, dashed, mark=square*,
error bars/.cd, y dir=both, y explicit
] table[row sep=\\, x=m, y=mean, y error=err] {
m     mean  err \\
0.1   0.97  0.00 \\
0.3   0.97  0.00 \\
0.5   0.97  0.00 \\
0.7   0.97  0.00 \\
};
\addlegendentry{$\ECF(\calV _r)$}

\addplot+[
green, thick, dotted, mark=triangle*,
error bars/.cd, y dir=both, y explicit
] table[row sep=\\, x=m, y=mean, y error=err] {
m     mean  err \\
0.1   0.39  0.00 \\
0.3   0.43  0.00 \\
0.5   0.41  0.00 \\
0.7   0.41  0.00 \\
};
\addlegendentry{$\EmCF(\calV _f)$}

\end{axis}
\end{tikzpicture}
\end{minipage}
\hfill
\begin{minipage}[t]{0.49\linewidth}
\centering
\begin{tikzpicture}
\begin{axis}[
width=\linewidth, height=0.7\linewidth,
grid=both,
xlabel={std scaling},
ylabel={value},
title={mean shift $=0.5$},
xmin=0.45, xmax=2.05,
ymin=0.0, ymax=1.05,
xtick={0.5,1.0,1.5,2.0},
tick label style={/pgf/number format/fixed},
clip=false
]

\addplot+[
blue, thick, solid, mark=*,
error bars/.cd, y dir=both, y explicit
] table[row sep=\\, x=s, y=mean, y error=err] {
s     mean  err \\
0.5   0.43  0.00 \\
1.0   0.61  0.00 \\
1.5   0.64  0.00 \\
2.0   0.61  0.00 \\
};

\addplot+[
red, thick, dashed, mark=square*,
error bars/.cd, y dir=both, y explicit
] table[row sep=\\, x=s, y=mean, y error=err] {
s     mean  err \\
0.5   0.96  0.00 \\
1.0   0.97  0.00 \\
1.5   0.97  0.00 \\
2.0   0.97  0.00 \\
};

\addplot+[
green, thick, dotted, mark=triangle*,
error bars/.cd, y dir=both, y explicit
] table[row sep=\\, x=s, y=mean, y error=err] {
s     mean  err \\
0.5   0.22  0.00 \\
1.0   0.41  0.00 \\
1.5   0.46  0.00 \\
2.0   0.41  0.00 \\
};

\end{axis}
\end{tikzpicture}
\end{minipage}

\begin{minipage}{\linewidth}\centering
\pgfplotslegendfromname{sharedlegend1}
\end{minipage}

\caption{\footnotesize CIFAR100, RepVGG-a2 5-class forgetting with $\alpha=0.05$ and $c=d=100$ (EFFACE). The shifts are applied after normalization. Left: metrics vs mean shift. Right: metrics vs std scaling.}
\label{fig:cifar100_efface_shift_mean_std}
\end{figure}

To illustrate the impact of affine transformations, we evaluate EFFACE under mean and standard deviation shifts applied simultaneously on $\retainset$ and $\forgetset$. \Cref{fig:cifar100_efface_shift_mean_std} shows that $\ECF_{\calV_r}$, $\EmCF_{\calV_f}$, and $H$ exhibit minimal variation under the considered shifts (with an exception when std scaling $< 1.0$, where compression of 
the training distribution causes a train-validation mismatch: the model learns on data with reduced variance but is evaluated on validation data with the original, larger variance, leading to degraded overall performance). Furthermore, during the experiments, we observed no significant change in the accuracy across $\calV_r$ and $\calV_f$, and thus these results are omitted for brevity. These findings suggest that EFFACE demonstrates resilience to small affine transformations, further underscoring its robustness in practical applications.


\section{Related Work}\label{sec:related-work}
Existing machine unlearning methods face significant challenges when applied to \gls{CP}, including (I1) reliance on retrained model baselines for evaluation and (I2) vulnerability to forgeability in parameter-space definitions. Below, we review relevant literature and highlight these limitations.

Most unlearning approaches aim to approximate a model retrained from scratch without the forget data \cite{cao2015towards, guo2020certified, ginart2019making, chien2024certifiednoisySGD, koloskova2025certifiednonconvex}. These methods often assume strong convexity \cite{sekhari2021remember, allouah2025utility, thudi2022unrolling} or employ practical mechanisms such as gradient-influence subtraction \cite{warnecke2023machine, graves2021amnesiac}, Bayesian updates \cite{nguyen2020variational}, teacher–student transfer \cite{chundawat2023bad}, noise-based unlearning \cite{chundawat2023zeroshot}, and information-theoretic objectives \cite{foster2024information, xu2025machine}. Other strategies include KL-divergence–based forgetting \cite{kurmanji2023towards}, selective gradient dampening \cite{trippa2024gradient, foster2024fast}, and adversarial mixup \cite{peng2025adversarial}. Recent work addresses "residual knowledge" by penalizing prediction deviations on neighboring samples of the forget data \cite{hsu2025unseenresidualknowledge}, but the objective remains to approximate a retrained model.

Certified unlearning methods, such as \cite{koloskova2025certifiednonconvex}, relax this goal by approximating any certifying model trained without the forget data. However, these approaches are susceptible to forgeability \cite{thudi2022auditable}, as even the original model's parameters can satisfy such definitions.

The first conformal-prediction–based evaluation for unlearning was proposed by \cite{shi2025rethinking}, which introduced the metric
\begin{align}\label{CR}
\CR({\calD}) \triangleq \frac{\sum_{(x,y)\in \calD} \indicate*{y\in \calC (X)}}{\sum_{(x,y)\in \calD} \abs{\calC (X)}},
\end{align}
targeting low $\CR$ on $\forgetset$ and high $\CR$ on $\retainset$. However, $\CR$ has notable limitations: (i) it may underestimate coverage on forget data due to large denominators, (ii) it can overestimate coverage on retained data with small prediction sets, and (iii) it is less interpretable in many-class settings due to uniformly small values. Empirically, $\CR$ sometimes fails to distinguish between forget and retain sets \cite{shi2025rethinking}. In contrast, our proposed metrics, $\ECF$ and $\EmCF$, directly measure coverage and miscoverage rates, aligning with the theoretical framework of $(\alpha,\beta)$-conformal unlearning (\cref{def:conformalMU}).

Methodologically, the CPU procedure in \cite{shi2025rethinking} updates conformity scores only for forget data, whereas EFFACE optimizes the miscoverage gap between forget and retain sets while constraining prediction set sizes. By grounding unlearning in explicit coverage and miscoverage targets, EFFACE addresses both I1 and I2.

Finally, $\nabla \tau$ \cite{trippa2024gradient} uses external data to induce forgetting by matching entropy losses between validation and training forget sets. However, it does not leverage conformity scores or target conformal objectives, leading to performance differences highlighted in our results.

By defining forget data through shared characteristics, conformal unlearning also mitigates the \emph{residual knowledge} problem identified in prior unlearning methods \cite{hsu2025unseenresidualknowledge}. Residual knowledge arises when an unlearned model's predictions align with those of a \gls{RT} model on the exact forget data points but deviate on slightly perturbed samples in their neighborhood, indicating that latent information about the forget data persists. In conformal unlearning, the shared characteristics defining the forget data ensure that small perturbations are likely to remain within the forget set, naturally extending the unlearning process to the local neighborhood of the forget data. 

This intuition can be formalized under mild regularity conditions. Suppose the scoring function $s$ is $M$-Lipschitz continuous with respect to its first argument. For any $X \sim \pdata$ and a perturbed input $X' = X + \delta$, where $\|\delta\| \leq \epsilon$ for some small $\epsilon > 0$, it follows that
\begin{align*}
|s(X,Y) - s(X',Y)| \leq M\|X - X'\| \leq M\epsilon \ \as
\end{align*}
By \cref{prop:bounded-point-wise-s}, if the \gls{cdf} of $s$ is $L$-Lipschitz, the perturbed scores $s(X',Y)$ satisfy \cref{eq:conformal-coverage-retain} with $\alpha' \leq \alpha + \frac{2LM\epsilon}{\pi_r}$ and \cref{eq:conformal-coverage-forget} with $\beta' \geq \beta - \frac{2LM\epsilon}{\pi_f}$. Therefore, for scoring functions that exhibit sufficient smoothness and for small perturbations, the guarantees of conformal unlearning degrade in a controlled manner within the neighborhood of the forget data. This directly mitigates the issue of residual knowledge by ensuring that the unlearning guarantees extend to local perturbations of the forget data.



\section{Conclusion}\label{sec:conclusion}
We have introduced a novel perspective on \gls{MU} by anchoring it in the framework of \gls{CP}, enabling a rigorous unlearning notion that is universal and unlinked to retrained baselines. By defining conformal \gls{MU} and corresponding empirical metrics, we offer a principled approach to evaluate unlearning effectiveness through the exclusion of forget data and the retention of coverage over retained data. This framework ensures statistical reliability for unlearning while preserving performance on retained data.
The conformal approach is inherently versatile, with potential extensions to regression tasks, graph neural networks, and natural language models.
Future work could explore tighter theoretical guarantees, adaptive methods tailored to diverse model architectures, and broader metrics to capture various dimensions of data influence.


\bibliographystyle{IEEEtran}
\bibliography{papers,books}

\pagebreak

\def\docprefix{S} 

\appendices   

\setcounter{equation}{0}
\setcounter{figure}{0}
\setcounter{table}{0}
\setcounter{section}{0}

\renewcommand{\theequation}{S\arabic{equation}}
\renewcommand{\thefigure}{S\arabic{figure}}
\renewcommand{\thetable}{S\arabic{table}}
\renewcommand{\thesection}{S\Roman{section}}
\renewcommand{\thesectiondis}{S\Roman{section}.}
\setcounter{Lemma}{0}
\renewcommand{\theLemma}{S\arabic{Lemma}}
\setcounter{Theorem}{0}
\renewcommand{\theTheorem}{S\arabic{Theorem}}

\begin{center}
\textbf{\Large Supplementary Material - Conformal Unlearning: A New Paradigm for Unlearning in Conformal Predictors}
\end{center}

\newcommand{\Lalgo}{\frakA}

\section{How Can Conformal Machine Unlearning Be Generalized?}\label[Appendix]{app:how-to-generalize}

We have focused on machine unlearning for clustering and classification tasks. However, our work establishes a foundational framework for a new paradigm of machine unlearning based on rigorous quantification of conformal prediction uncertainty. This paradigm naturally extends to regression tasks through appropriate non-conformity score functions, as demonstrated in foundational conformal prediction literature (cf.\ \cite{angelopoulos2022gentle, shafer2008conformal}). Furthermore, the approach generalizes to any domain where conformal prediction has been developed with suitable handling of its theoretical foundations. For instance, conformal prediction has been successfully applied to graph neural networks \citeSR{zargarbashi2023uncertainty, sadinle2019least, huang2023uncertainty, clarkson2022distribution}, natural language processing \citeSR{cha2025towards, quach2024conformal, liu2025rethinking, zhang2024right}, and other emerging application areas. Consequently, our paradigm is immediately applicable to contemporary machine unlearning research across these domains. Extending the framework to these additional areas represents a promising direction for future work.\footnote{The code repository of this work can be found here: \url{https://github.com/Y-kht/efface_official}.}


\section{More on the Datasets and Baselines}

\subsection{Datasets}\label[Appendix]{app:datasets} 

\textbf{CIFAR100} \cite{cifar100} is a carefully curated, labeled subset of the 80 Million Tiny Images dataset developed by Alex Krizhevsky, Vinod Nair, and Geoffrey Hinton . It comprises 60 000 color images of size 32×32 pixels, evenly distributed across 100 distinct object classes . Each class contains exactly 600 images, which are split into 500 samples for training and 100 for testing. These 100 classes are further organized into 20 higher-level “superclasses,” enabling both fine-grained and coarse-grained classification experiments. Every image carries two annotations: a fine label denoting its specific class and a coarse label indicating its superclass. The small 32×32 resolution makes CIFAR100 computationally efficient for prototyping convolutional networks and other vision models. Its perfectly balanced class distribution and hierarchical labelling have established CIFAR100 as a standard benchmark in the computer-vision community. The dataset shares its file‐format conventions (Python “pickled” batches, MATLAB files, or binary versions) with CIFAR-10, where each batch bundles image data and labels together \cite{cifar100}. CIFAR100 is also natively supported in major ML libraries like TensorFlow Datasets and PyTorch’s torchvision for seamless integration into research pipelines. In our experiments, we load CIFAR100 using torchvision's datasets library.

\textbf{ImageNet100} \cite{imagenet100} is a compact subset of ILSVRC 2012, containing 100 classes randomly sampled from the original 1,000; it was assembled by the Kaggle user \texttt{ambityga} and released in August 2021 as “A Sample of ImageNet Classes.” The included categories are listed in a \texttt{Labels.json} file, and the dataset is widely used as a smaller, more manageable proxy for ImageNet in research, experimentation, and teaching.

\textbf{20NewsGroups Dataset} \cite{20newsgroups} comprises roughly 20,000 English posts nearly evenly distributed across 20 topics, originally collected by Ken Lang for the 1995 ''Newsweeder'' study and now a staple benchmark for text classification and clustering. Documents are plain text (headers plus message body), and popular distributions include the scikit-learn version and a SetFit release on Hugging Face that stores \texttt{text}, integer \texttt{label} (0–19), and \texttt{label\_text}, providing convenient modern access while preserving the dataset’s original structure.


\subsection{On The Baselines}\label[Appendix]{app:baselines} 

Here we summarize the objectives of each of the baselines and how they achieve unlearning. This section is not meant to be comprehensive but to give a clearer idea about the methods we compare with.

$\nabla \tau$ \cite{trippa2024gradient} introduces a new loss objective that focuses on pushing the loss of the data meant to be forgotten from the training data to become larger than the loss of some validation data (carrying the same unlearned labels). They merge it with the original objective of minimizing the loss on the retained data. The new loss is then given by
\begin{align*}
    L = \alpha \parens*{\mathrm{ReLU}(L_{\calD_v}-L_{\forgetset})} + (1-\alpha)L_{\retainset},
\end{align*}
where $\alpha$ controls how much emphases should be given to the retained versus forgotten data. In our framework, however, non of the subsets $\calD_v$, $\calD_f$, and $\calD_r$ is seen during training. We assume that this causes the performance of $\nabla \tau$ to drop.

SCRUB \cite{kurmanji2023towards} builds the loss function on the KL-divergence of the unlearning model to a teacher that was trained on the full training data (both $\calD _r$ and $\calD_f$). This is nothing but the original base model trained on $\trainset$. They add one more regular loss term to be minimized over the retained data to maintain performance on those points. The final loss that should be minimized becomes
\begin{align*}
    L &= \frac{\alpha}{N_r}\sum\limits_{x_r\in\calD_r}d_{KL}(x_r;\omega^u) + \frac{\gamma}{N_r}\sum\limits_{(x_r,y_r)\in\calD_r}l(f(x_r;\omega^u),y_r) \\
    &-  \ofrac{N_f}\sum\limits_{x_f\in\calD_f}d_{KL}(x_f;\omega^u),
\end{align*}
where $N_r$ is the number of data points to retain, $N_f$ the number of data points to forget, and $\alpha$ and $\gamma$ control the importance of the terms of retaining. Notice that SCRUB tries to make the distributions of the unlearning model and the base model converge to each other on the retained data and diverge from each other on the forgotten data. Note that when $\ulset$ is a proxy set not used in pretraining, then the base model is less capable of correctly classifying $\retainset$, which leaves the unlearned model with high variance. 

SSD \cite{foster2024fast} uses synaptic dampening of the parameters (weights) of the model which are ''specialized'' for $\calD _f$. SSD compares the ''importances'' of the weights using the first-order derivative property of the Fisher Information Matrix (FIM) and decides whether to dampen a weight if it is more specialized for $\calD _f$ than for other training data, as follows.

\begin{align*}
    []_{\calD}
    &= \E[-\evalat*{\frac{\delta^2 \ln p(\calD\mid \theta)}
                             {\delta \theta^2}}_{\theta^{*}_{D}}],\\
    []_{\calD}
    &= \E[\parens{\frac{\delta \ln p(\calD\mid \theta)}{\delta \theta}}
                           \parens{\frac{\delta \ln p(\calD\mid \theta)}{\delta \theta}}^{T}
                             |_{\theta^{*}_{D}}].
\end{align*}

\begin{align*}
    \beta
    &= \min\parens{\lambda \frac{[]_{\calD,i}}{[]_{\calD_{f,i}}},1},\\
    \theta_i
    &=
    \begin{cases}
    \beta\,\theta_i, & \text{ if }[]_{\calD_{f,i}} > \alpha\,[]_{\calD,i},\\
    \theta_i,         & \text{ if }[]_{\calD_{f,i}} \le \alpha\,[]_{\calD,i},
    \end{cases}
    \;  \forall\,i\in[0,\lvert\theta\rvert].
\end{align*}

Generally, they assume that the training data importances can be calculated before training and then the importances of the parameters will be compared between $\calD _f$ and $\trainset$. In our framework, $\forgetset$ might not be part of $\trainset$. Hence, it is not straightforward to argue for a stable relation between the importances. That seems to be the reason why SSD fails to unlearn when $\ulset$ is a proxy set.

AMN \cite{graves2021amnesiac} randomly relabels the data to be forgotten. It replaces the classes to be forgotten with new random labels over the whole training set and then retrains the model for a few iterations over the newly labeled data. However, in our case we feed $\forgetset$ and $\retainset$ to the unlearning algorithm rather than the full training set. Since $\forgetset$ in our framework might not be used for training or is just a smaller subset compared to the fraction of forgotten data in the training set, we find that AMN overfits to $\ulset$ and does not perform well on the other subsets of data.

BADT \cite{chundawat2023bad} introduces a bad teacher initialized with random noise which induces forgetting by minimizing the KL-divergence between its distribution and that of the unlearned model (student) on the forgotten data. On the other hand, BADT minimizes the divergence between the distribution of the base model and that of the student on the retained data. The objective of BADT is given below.
\begin{align*}
    L(x,l_u) = (1-l_u)\calK \calL (T_s(x) || S(x)) + l_u \calK \calL (T_d(x) || S(x)),
\end{align*}
where $l_u$ is the label to be forgotten, $x$ is a sample point, $T_s$ is the base model, $T_d$ is the bad teacher model, and $\calK \calL (P,Q)$ is the KL-divergence between the $P$ and $Q$. 
Note that BADT is initially proposed for label-wise forgetting but can be used in the cluster-wise case.

UNSIR \cite{tarun2024fast} constructs noisy data by maximizing the loss on the noisy samples that carry the label to be forgotten. Then, it feeds the loss-maximizing noise to the model along with some retrained data in an impair-repair fashion. UNSIR depends on the label to be forgotten to construct the loss-maximizing noise. Hence, it is not suitable for targeted cluster-wise forgetting. That is why we do not include its results in those scenarios.

PABI \cite{koloskova2025certifiednonconvex} is an ($\epsilon$,$\delta$)-unlearning certified method that attempts to approximate a model trained without the forgotten data in the parameters space. In particular, the define ($\epsilon$,$\delta$)-unlearning as follows.
\begin{Definition}[($\epsilon$,$\delta$)-unlearning {\cite[Def. 2.1]{koloskova2025certifiednonconvex}}]
    Let $\epsilon \ge 0$, $\delta \in [0, 1]$. We say that $\Ualgo$ is ($\epsilon$, $\delta$)-unlearning algorithm for $\Lalgo$ if there exists a certifying algorithm $\tilde \Lalgo$, such that for any forget and initial datasets $\forgetset \in \calD$ and any observation $\theta \in \bbR ^d$,
    \begin{align*}
        \P(\Ualgo(\Lalgo(\calD),\calD,\forgetset) = \theta) \le e^{\epsilon} \P(\tilde \Lalgo(\calD \setminus \forgetset ) = \theta) &+ \delta,\\
        \P(\tilde \Lalgo(\calD \setminus \forgetset ) = \theta) \le e^{\epsilon} \P(\Ualgo(\Lalgo(\calD),\calD,\forgetset) = \theta) &+ \delta. 
    \end{align*}
\end{Definition}
Note that the certifying algorithm $\tilde \Lalgo$ might not be the original training algorithm. It is just a training algorithm that is not trained on the forgotten data. Hence, its result might be a model very different that a model trained from scratch using the same original training algorithm but only on the retained data (RT). PABI induces this unlearning by adding noise to the gradients during training and clipping the weights. The exact approach is as follows.
\begin{align*}
    x_0 &= \prod_{C_0}(\hat x),\\
    x_{t+1} &= x_t - \gamma(\prod_{C_1} (g_t) + \lambda x_t) + \varsigma_{t+1},
\end{align*}
where $\varsigma_{t+1} \sim \calN(0, \sigma^2 I^d)$ is Gaussian noise, and $\prod_{C_0}$ ,$\prod_{C_1}$
are the clipping operators of radii $C_0, C_1 > 0$, respectively. The PABI method employs a combination of noisy updates to induce forgetting of $\forgetset$, followed by fine-tuning on $\retainset$. In our experiments, the number of noisy update steps was minimal (typically one), resulting in a fine-tuning phase with a number of epochs equivalent to that of retraining from scratch. We observe that PABI does not achieve convergence to the same validation accuracy in fewer epochs relative to RT.


\section{Further Experimental Details}

\subsection{Reproducibility Details}\label[Appendix]{app:reproduce} 

\paragraph{Environment.}
Four NVIDIA RTX A5000 GPUs; PyTorch with \texttt{nn.DataParallel}; batch size $256$; dataloader workers $2$; no memory pinning. All vision inputs are normalized with the standard dataset statistics; text tokenization follows the BERTa-Distill’s pipeline from the \texttt{transformers} library.

\paragraph{Training recipes.}
\textbf{CIFAR100:} ResNet18, SGD $50$ epochs, initial lr $0.1$ with linear decay to $10^{-4}$, momentum $0.9$, wd $5{\times}10^{-4}$, cross-entropy, no early stopping.  
\textbf{ImageNet100 (100 labels):} ResNet18, same as CIFAR100 but $80$ epochs.  
\textbf{20NewsGroups:} BERTa-Distill, $15$ epochs, initial lr $0.01$, otherwise as above. The RepVGG-A2 models for CIFAR100\footnote{Provided here: https://github.com/chenyaofo/pytorch-cifar-models} and ImageNet\footnote{Provided here: https://github.com/DingXiaoH/RepVGG/blob/main/repvgg.py} were loaded with pretrained weights.
All tables/plots are averaged over 6 random seeds except the experiments on 20NewsGroups and News which are averaged over 3 seeds. 

\paragraph{Partitions and calibration.}
\textbf{CIFAR100:} from the 50k train split, use 45k for training, 5k held out: 2.5k as $\calibset$ (for quantile estimation during unlearning) and 2.5k split into unseen $\calV_f,\calV_r$. The 10k test split yields an 8k testing calibration set and label-based $\forgetset,\retainset$ for unlearning. Baselines that need validation use $\calibset$.  
\textbf{Other datasets:} analogous retain/forget and calibration partitions: $90\%$ of the train split is used for training, the remaining $10\%$ of the train split is used for validation, if any baseline uses a validation split ($5\%$) and $\calV_r$ and $\calV_f$ ($5\%$), and $80\%$ of the test split is used for testing (reconformalization) calibration $\calibset$ and $20\%$ is used for $\retainset$ and $\forgetset$.

\paragraph{Unlearning optimization.}
For EFFACE, we retain SGD with the base momentum and weight decay; we tune only the learning rate (grid-search). No scheduler during unlearning. For other baselines, we keep the same optimizers used in authors' repositories.

\paragraph{Baselines.}
$\nabla\tau$ \cite{trippa2024gradient}, SCRUB \cite{kurmanji2023towards}, SSD \cite{foster2024fast}, AMN \cite{graves2021amnesiac}, BADT \cite{chundawat2023bad}, UNSIR \cite{tarun2024fast}; PABI \cite{koloskova2025certifiednonconvex}; plus RT on $\calT_r$. We use authors’ repositories (from \cite{foster2024fast,chundawat2023bad}); we grid-search around released settings to keep compute comparable only for CIFAR100 and ImageNet100, and we use those same settings for 20NewsGroups. Our evaluation applies unlearning on $\forgetset/\retainset$. We implement PABI on our own.

\paragraph{MIA evaluation.}
For each sample we extract: loss, entropy, prediction margin, logit $\ell_2$-norm, and top-\emph{k} probabilities (dynamic $k$). We train a RandomForest attacker with stratified 10-fold cross-validation and report \emph{Adversarial Advantage} (attacker accuracy minus majority-class ratio). We include MIA Diff in all tables.

\paragraph{Metrics (formal).}
$A_{\calD}$, $\ECF_{\calD}(c)$ for $\calD\in\{\calD_r,\calT_r,\calV_r\}$, $\EmCF_{\calD}(d)$ for $\calD\in\{\calD_f,\calT_f,\calV_f\}$ with $c{=}d$, harmonic mean $H$ over the six conformal metrics (defined as $H=\frac{n}{\sum_i x_i^{-1}}$, with $H{=}0$ if any $x_i{=}0$), MIA Diff, and Tsec. Implementation details for $\ECF/\EmCF$ and calibration protocols follow main text theory.

\paragraph{Code and Reproducibility.}
Scripts for dataset preparation, partition seeds, hyperparameter grids, and exact command lines are provided in the accompanying repository (to be made public upon publication). We fix seeds for data splits and model initialization to ensure full reproducibility. The hyperparameters used in our experiments are detailed below.

\begin{itemize}
    \item \textbf{EFFACE:} $\kappa = 5.0$, $\gamma = 0.0$, $\rho = 0.0$ in all experiments. Number of epochs is 20, except for CIFAR100 cluster-wise forgetting (35 epochs). Learning rate is 0.04 for all experiments, except for ImageNet100 RepVGG-A2 in-sample forgetting ($\mathrm{lr} = 0.02$) and \cref{tab:EFFACE_vs_CPU} ($\mathrm{lr} = 0.08$).
    
    \item \textbf{$\nabla\tau$:} Split ratio $= 0.5$ and $\mathrm{lr} = 1 \times 10^{-4}$ in all experiments.
    
    \item \textbf{SCRUB:} Epochs $= 10$, $\gamma = 2.0$, $\beta = 0.1$, msteps $= 3$, sstart $= 10$, kd-T $= 4$, and $\mathrm{lr} = 0.01$ in all experiments, except for ImageNet100 in-sample class-wise forgetting ($\mathrm{lr} = 0.04$).
    
    \item \textbf{AMN:} $\mathrm{lr} = 1 \times 10^{-3}$ and epochs $= 8$ in all experiments.
    
    \item \textbf{SSD:} $\lambda = 1$ in all experiments. $\alpha = 20$ for in-sample experiments and $\alpha = 50000$ for out-of-sample experiments (larger values cause SSD to collapse in the out-of-sample setting).
    
    \item \textbf{UNSIR:} 150 noise-generating epochs, 5 impair epochs, and 10 repair epochs, with $\mathrm{lr} = 1 \times 10^{-4}$.
    
    \item \textbf{BADT:} KL temperature $= 1.0$, $\mathrm{lr} = 0.03$, and epochs $= 10$ in all experiments.
    
    \item \textbf{PABI} (used in \cref{tab:EFFACE_vs_PABI_RT}): Constant noise scheduler, $\epsilon = 1.0$, $\delta = 1 \times 10^{-5}$, initial model clipping constant $= 0.01$, gradient clipping constant $= 10.0$, maximum learning rate $= 0.001$, $\lambda = 500$, and initial $\sigma = 0.0$.
\end{itemize}

\subsection{More Results}\label[Appendix]{app:extra-empirics}
In this appendix we show more results on CIFAR100 and 20NewsGroups and both class-wise unlearning as well as cluster-wise unlearning. For image datasets, the model before unlearning is RepVGG-a2, while for text datasets, it is Berta-distill. Note that cluster-wise unlearning is performed by first clustering all the data points in the embedding space using $k$-means. That is, we take the high representation of the points produced by a pre-trained model (the same model used as the original model prior to unlearning) and perform $k$-means in that space. We use $k=|\calY|$ in k-means, resembling the number of labels in that dataset. Then, we pick a specified number of clusters to forget.

In all the tables in this appendix, we follow the same convention as in the main text by highlighting the best, second best, and third best $\EmCF$ and $H$ results with \first{}, \second{}, and \third{}, respectively. Note that by \cref{lem:alpha-beta-tradeoff},  $\beta$ is subject to the bound $\beta \le \frac{\alpha}{\pi_f}$. E.g., if $\alpha = 0.05$ (as is the case in the following tables), and $\pi_f \approx 0.25$, then $\beta \le 0.25$, marginally over the forget data subsets.


\begin{table*}[htbp]
\caption{CIFAR100, RepVGG-a2 \emph{cluster}-wise forgetting with $c=d=50$, $\alpha=0.05$, and 5 forgotten clusters. Coverage/miscoverage results.}
\label{tab:results_cifar100_clusters_f5_c50_a005_coverage}
\centering
\resizebox{\linewidth}{!}{%
\begin{tabular}{llccccccc}
\toprule
Split & Method
& $\ECF_{c}[\retainset]\uparrow$
& $\EmCF_{d}[\forgetset]\uparrow$
& $\ECF_{c}[\calT_r]\uparrow$
& $\EmCF_{d}[\calT_f]\uparrow$
& $\ECF_{c}[\calV_r]\uparrow$
& $\EmCF_{d}[\calV_f]\uparrow$
& $H\uparrow$ \\
\midrule

& OR
& $1.00 \pm 0.00$
& $0.00 \pm 0.00$
& $1.00 \pm 0.00$
& $0.00 \pm 0.00$
& $0.95 \pm 0.00$
& $0.06 \pm 0.00$
& $0.00 \pm 0.00$ \\

\midrule

\multirow{6}{*}{In}

& $\nabla\tau$
& $1.00 \pm 0.00$
& \first{$1.00 \pm 0.00$}
& \first{$1.00 \pm 0.00$}
& \first{$1.00 \pm 0.00$}
& \first{$0.98 \pm 0.00$}
& \second{$0.85 \pm 0.01$}
& \cellcolor{green!20}\first{$0.97 \pm 0.00$} \\

& SCRUB
& $1.00 \pm 0.00$
& $0.05 \pm 0.02$
& $0.99 \pm 0.00$
& $0.05 \pm 0.01$
& $0.95 \pm 0.00$
& $0.08 \pm 0.01$
& $0.11 \pm 0.02$ \\

& SSD
& $1.00 \pm 0.00$
& $0.95 \pm 0.04$
& \first{$1.00 \pm 0.00$}
& \third{$0.87 \pm 0.02$}
& \second{$0.97 \pm 0.00$}
& \third{$0.63 \pm 0.01$}
& \third{$0.88 \pm 0.01$} \\

& AMN
& $1.00 \pm 0.00$
& \cellcolor{green!20}\third{$0.97 \pm 0.01$}
& \first{$1.00 \pm 0.00$}
& \cellcolor{gray!20}$0.19 \pm 0.02$
& $0.95 \pm 0.00$
& \cellcolor{gray!20}$0.17 \pm 0.02$
& $0.39 \pm 0.02$ \\

& BADT
& $1.00 \pm 0.00$
& $0.02 \pm 0.02$
& \first{$1.00 \pm 0.00$}
& $0.02 \pm 0.01$
& $0.95 \pm 0.00$
& $0.16 \pm 0.01$
& $0.04 \pm 0.02$ \\

& EFFACE
& $1.00 \pm 0.00$
& \first{$1.00 \pm 0.00$}
& \first{$1.00 \pm 0.00$}
& \first{$1.00 \pm 0.00$}
& \second{$0.97 \pm 0.00$}
& \first{$0.86 \pm 0.01$}
& \first{$0.97 \pm 0.00$} \\

\midrule

\multirow{6}{*}{Out}

& $\nabla\tau$
& \first{$1.00 \pm 0.00$}
& $0.24 \pm 0.16$
& \first{$1.00 \pm 0.00$}
& $0.03 \pm 0.06$
& \third{$0.95 \pm 0.00$}
& $0.08 \pm 0.05$
& \cellcolor{gray!20}$0.11 \pm 0.14$ \\

& SCRUB
& \first{$1.00 \pm 0.00$}
& $0.12 \pm 0.05$
& $0.96 \pm 0.01$
& \third{$0.14 \pm 0.05$}
& $0.94 \pm 0.01$
& \third{$0.14 \pm 0.07$}
& \third{$0.23 \pm 0.05$} \\

& SSD
& $0.96 \pm 0.00$
& \third{$0.45 \pm 0.02$}
& \first{$1.00 \pm 0.00$}
& \second{$0.36 \pm 0.02$}
& \second{$0.96 \pm 0.00$}
& \second{$0.43 \pm 0.03$}
& \second{$0.58 \pm 0.01$} \\

& AMN
& \first{$1.00 \pm 0.00$}
& \second{$0.92 \pm 0.03$}
& \first{$1.00 \pm 0.00$}
& $0.00 \pm 0.00$
& \third{$0.95 \pm 0.00$}
& $0.04 \pm 0.01$
& $0.02 \pm 0.01$ \\

& BADT
& $0.94 \pm 0.00$
& $0.00 \pm 0.00$
& \first{$1.00 \pm 0.00$}
& $0.00 \pm 0.00$
& \third{$0.95 \pm 0.00$}
& $0.01 \pm 0.00$
& $0.00 \pm 0.00$ \\

& EFFACE
& $0.99 \pm 0.00$
& \first{$1.00 \pm 0.00$}
& \first{$1.00 \pm 0.00$}
& \first{$0.82 \pm 0.06$}
& \first{$0.97 \pm 0.00$}
& \first{$0.78 \pm 0.03$}
& \first{$0.92 \pm 0.02$} \\

\bottomrule
\end{tabular}%
}
\end{table*}

\begin{table*}[htbp]
\caption{CIFAR100, RepVGG-a2 \emph{class}-wise forgetting with $c=d=50$, $\alpha=0.05$, and 5 forgotten classes. Coverage/miscoverage results.}
\label{tab:results_cifar100_class_f5_c50_a005_coverage}
\centering
\resizebox{\linewidth}{!}{%
\begin{tabular}{llccccccc}
\toprule
Split & Method
& $\ECF_{c}[\retainset]\uparrow$
& $\EmCF_{d}[\forgetset]\uparrow$
& $\ECF_{c}[\calT_r]\uparrow$
& $\EmCF_{d}[\calT_f]\uparrow$
& $\ECF_{c}[\calV_r]\uparrow$
& $\EmCF_{d}[\calV_f]\uparrow$
& $H\uparrow$ \\
\midrule

& OR
& $1.00 \pm 0.00$
& $0.00 \pm 0.00$
& $1.00 \pm 0.00$
& $0.00 \pm 0.00$
& $0.95 \pm 0.00$
& $0.05 \pm 0.00$
& $0.00 \pm 0.00$ \\

\midrule

\multirow{7}{*}{In}

& $\nabla\tau$
& $1.00 \pm 0.00$
& \first{$1.00 \pm 0.00$}
& \first{$1.00 \pm 0.00$}
& \first{$1.00 \pm 0.00$}
& \first{$0.99 \pm 0.00$}
& \first{$1.00 \pm 0.00$}
& \cellcolor{green!20}\first{$1.00 \pm 0.00$} \\

& SCRUB
& $1.00 \pm 0.00$
& $0.04 \pm 0.04$
& $0.99 \pm 0.01$
& $0.05 \pm 0.04$
& $0.95 \pm 0.00$
& $0.15 \pm 0.03$
& $0.12 \pm 0.05$ \\

& SSD
& $1.00 \pm 0.00$
& $0.93 \pm 0.03$
& \first{$1.00 \pm 0.00$}
& \second{$0.90 \pm 0.01$}
& \third{$0.97 \pm 0.00$}
& \second{$0.94 \pm 0.01$}
& \second{$0.96 \pm 0.01$} \\

& AMN
& $1.00 \pm 0.00$
& \cellcolor{green!20}\second{$0.97 \pm 0.01$}
& \first{$1.00 \pm 0.00$}
& \cellcolor{gray!20}$0.19 \pm 0.03$
& $0.96 \pm 0.00$
& \cellcolor{gray!20}$0.32 \pm 0.04$
& $0.48 \pm 0.03$ \\

& BADT
& $1.00 \pm 0.00$
& $0.00 \pm 0.00$
& \first{$1.00 \pm 0.00$}
& $0.00 \pm 0.00$
& $0.96 \pm 0.00$
& $0.28 \pm 0.01$
& $0.00 \pm 0.00$ \\

& UNSIR
& $1.00 \pm 0.00$
& $0.01 \pm 0.00$
& \first{$1.00 \pm 0.00$}
& $0.01 \pm 0.00$
& $0.95 \pm 0.00$
& $0.06 \pm 0.01$
& $0.02 \pm 0.01$ \\

& EFFACE
& $1.00 \pm 0.00$
& \second{$0.98 \pm 0.01$}
& \first{$1.00 \pm 0.00$}
& \third{$0.89 \pm 0.02$}
& \second{$0.98 \pm 0.00$}
& \third{$0.92 \pm 0.01$}
& \second{$0.96 \pm 0.00$} \\

\midrule

\multirow{7}{*}{Out}

& $\nabla\tau$
& \first{$1.00 \pm 0.00$}
& $0.40 \pm 0.24$
& $1.00 \pm 0.00$
& \third{$0.09 \pm 0.11$}
& \second{$0.96 \pm 0.00$}
& \third{$0.36 \pm 0.15$}
& \cellcolor{gray!20}\third{$0.31 \pm 0.21$} \\

& SCRUB
& \first{$1.00 \pm 0.00$}
& $0.33 \pm 0.03$
& $1.00 \pm 0.00$
& $0.05 \pm 0.02$
& \second{$0.96 \pm 0.00$}
& $0.16 \pm 0.01$
& $0.18 \pm 0.04$ \\

& SSD
& $0.97 \pm 0.00$
& \third{$0.46 \pm 0.02$}
& $1.00 \pm 0.00$
& \second{$0.36 \pm 0.00$}
& \second{$0.96 \pm 0.00$}
& \second{$0.45 \pm 0.01$}
& \second{$0.59 \pm 0.01$} \\

& AMN
& \first{$1.00 \pm 0.00$}
& \second{$0.95 \pm 0.02$}
& $1.00 \pm 0.00$
& $0.04 \pm 0.02$
& \second{$0.96 \pm 0.00$}
& $0.27 \pm 0.02$
& $0.19 \pm 0.05$ \\

& BADT
& $0.95 \pm 0.00$
& $0.04 \pm 0.00$
& $1.00 \pm 0.00$
& $0.00 \pm 0.00$
& $0.95 \pm 0.00$
& $0.10 \pm 0.00$
& $0.00 \pm 0.00$ \\

& UNSIR
& \first{$1.00 \pm 0.00$}
& $0.02 \pm 0.02$
& $1.00 \pm 0.00$
& $0.01 \pm 0.00$
& $0.95 \pm 0.00$
& $0.04 \pm 0.01$
& $0.04 \pm 0.01$ \\

& EFFACE
& $0.99 \pm 0.00$
& \first{$0.98 \pm 0.01$}
& $1.00 \pm 0.00$
& \first{$0.68 \pm 0.04$}
& \first{$0.97 \pm 0.00$}
& \first{$0.79 \pm 0.03$}
& \first{$0.88 \pm 0.01$} \\

\bottomrule
\end{tabular}%
}
\end{table*}



\begin{figure*}[htpb]
\centering

\begin{minipage}[c]{0.00\linewidth}
\hspace{\linewidth} 
\end{minipage}
\hfill
\begin{minipage}[t]{4cm}
\centering
\textbf{$\alpha=0.05,\; |\forgetset|=5$}
\end{minipage}
\hfill
\begin{minipage}[t]{4cm}
\centering
\textbf{$c=d=100,\; |\forgetset|=5$}
\end{minipage}
\hfill
\begin{minipage}[t]{4cm}
\centering
\textbf{$\alpha=0.05,\; c=d=100$}
\end{minipage}

\vspace{0.2cm}


\begin{minipage}[c]{0.03\linewidth}
\centering
{\rotatebox{90}{\textbf{In-Sample}}}
\end{minipage}
\hfill
\begin{minipage}[t]{0.30\linewidth}
\centering
\begin{tikzpicture}
\begin{axis}[
width=\linewidth, height=0.75\linewidth,
grid=both,
xlabel={$c,d$},
ylabel={$H$},
xmin=5, xmax=105,
ymin=0.0, ymax=1.0,
xtick={20,40,60,80,100},
ytick={0,0.2,0.4,0.6,0.8,1.0},
legend to name=sharedlegend_cifar,
legend columns=4,
legend style={font=\small, /tikz/every even column/.append style={column sep=0.3cm}},
]

\addplot+[magenta, thick, solid, mark=pentagon*, error bars/.cd, y dir=both, y explicit] table[row sep=\\, x=c, y=mean, y error=err] {
c   mean  err \\
10  0.00  0.00 \\
20  0.00  0.00 \\
40  0.00  0.00 \\
60  0.00  0.00 \\
80  0.00  0.00 \\
100 0.00  0.00 \\
};
\addlegendentry{BADT}

\addplot+[violet, thick, solid, mark=x, error bars/.cd, y dir=both, y explicit] table[row sep=\\, x=c, y=mean, y error=err] {
c   mean  err \\
10  0.01  0.00 \\
20  0.02  0.00 \\
40  0.02  0.00 \\
60  0.02  0.00 \\
80  0.02  0.00 \\
100 0.06  0.00 \\
};
\addlegendentry{UNSIR}

\addplot+[blue, thick, solid, mark=*, error bars/.cd, y dir=both, y explicit] table[row sep=\\, x=c, y=mean, y error=err] {
c   mean  err \\
10  1.00  0.00 \\
20  1.00  0.00 \\
40  1.00  0.00 \\
60  1.00  0.00 \\
80  1.00  0.00 \\
100 0.92  0.00 \\
};
\addlegendentry{$\nabla \tau$}

\addplot+[red, thick, dashed, mark=square*, error bars/.cd, y dir=both, y explicit] table[row sep=\\, x=c, y=mean, y error=err] {
c   mean  err \\
10  0.27  0.00 \\
20  0.17  0.00 \\
40  0.12  0.00 \\
60  0.11  0.00 \\
80  0.11  0.00 \\
100 0.11  0.00 \\
};
\addlegendentry{SCRUB}

\addplot+[orange, thick, dashdotdotted, mark=diamond*, error bars/.cd, y dir=both, y explicit] table[row sep=\\, x=c, y=mean, y error=err] {
c   mean  err \\
10  1.00  0.00 \\
20  1.00  0.00 \\
40  0.99  0.00 \\
60  0.91  0.00 \\
80  0.76  0.00 \\
100 0.56  0.00 \\
};
\addlegendentry{SSD}

\addplot+[green!60!black, thick, dotted, mark=triangle*, error bars/.cd, y dir=both, y explicit] table[row sep=\\, x=c, y=mean, y error=err] {
c   mean  err \\
10  0.91  0.00 \\
20  0.80  0.00 \\
40  0.58  0.00 \\
60  0.41  0.00 \\
80  0.36  0.00 \\
100 0.35  0.00 \\
};
\addlegendentry{AMN}

\addplot+[black, thick, solid, mark=star, error bars/.cd, y dir=both, y explicit] table[row sep=\\, x=c, y=mean, y error=err] {
c   mean  err \\
10  1.00  0.00 \\
20  0.99  0.00 \\
40  0.98  0.00 \\
60  0.94  0.00 \\
80  0.89  0.00 \\
100 0.76  0.00 \\
};
\addlegendentry{EFFACE}

\end{axis}
\end{tikzpicture}
\end{minipage}
\hfill
\begin{minipage}[t]{0.30\linewidth}
\centering
\begin{tikzpicture}
\begin{axis}[
width=\linewidth, height=0.75\linewidth,
grid=both,
xlabel={$\alpha$},
ylabel={$H$},
xmin=0.04, xmax=0.26,
ymin=0.0, ymax=1.0,
xtick={0.05,0.1,0.15,0.2,0.25},
ytick={0,0.2,0.4,0.6,0.8,1.0},
tick label style={/pgf/number format/fixed},
]

\addplot+[magenta, thick, solid, mark=pentagon*, error bars/.cd, y dir=both, y explicit] table[row sep=\\, x=a, y=mean, y error=err] {
a    mean  err \\
0.05 0.00  0.00 \\
0.10 0.13  0.00 \\
0.15 0.70  0.00 \\
0.20 0.91  0.00 \\
0.25 0.94  0.00 \\
};

\addplot+[violet, thick, solid, mark=x, error bars/.cd, y dir=both, y explicit] table[row sep=\\, x=a, y=mean, y error=err] {
a    mean  err \\
0.05 0.01  0.00 \\
0.10 0.03  0.00 \\
0.15 0.09  0.00 \\
0.20 0.19  0.00 \\
0.25 0.42  0.00 \\
};

\addplot+[blue, thick, solid, mark=*, error bars/.cd, y dir=both, y explicit] table[row sep=\\, x=a, y=mean, y error=err] {
a    mean  err \\
0.05 0.92  0.00 \\
0.10 0.99  0.00 \\
0.15 0.98  0.00 \\
0.20 0.97  0.00 \\
0.25 0.96  0.00 \\
};

\addplot+[red, thick, dashed, mark=square*, error bars/.cd, y dir=both, y explicit] table[row sep=\\, x=a, y=mean, y error=err] {
a    mean  err \\
0.05 0.11  0.00 \\
0.10 0.23  0.00 \\
0.15 0.37  0.00 \\
0.20 0.52  0.00 \\
0.25 0.64  0.00 \\
};

\addplot+[orange, thick, dashdotdotted, mark=diamond*, error bars/.cd, y dir=both, y explicit] table[row sep=\\, x=a, y=mean, y error=err] {
a    mean  err \\
0.05 0.56  0.00 \\
0.10 0.99  0.00 \\
0.15 0.98  0.00 \\
0.20 0.97  0.00 \\
0.25 0.96  0.00 \\
};

\addplot+[green!60!black, thick, dotted, mark=triangle*, error bars/.cd, y dir=both, y explicit] table[row sep=\\, x=a, y=mean, y error=err] {
a    mean  err \\
0.05 0.35  0.00 \\
0.10 0.58  0.00 \\
0.15 0.76  0.00 \\
0.20 0.87  0.00 \\
0.25 0.82  0.00 \\
};

\addplot+[black, thick, solid, mark=star, error bars/.cd, y dir=both, y explicit] table[row sep=\\, x=a, y=mean, y error=err] {
a    mean  err \\
0.05 0.76  0.00 \\
0.10 0.97  0.00 \\
0.15 0.98  0.00 \\
0.20 0.97  0.00 \\
0.25 0.96  0.00 \\
};

\end{axis}
\end{tikzpicture}
\end{minipage}
\hfill
\begin{minipage}[t]{0.30\linewidth}
\centering
\begin{tikzpicture}
\begin{axis}[
width=\linewidth, height=0.75\linewidth,
grid=both,
xlabel={$|\forgetset|$},
ylabel={$H$},
xmin=0, xmax=55,
ymin=0.0, ymax=1.0,
xtick={10,20,30,50},
ytick={0,0.2,0.4,0.6,0.8,1.0},
]

\addplot+[magenta, thick, solid, mark=pentagon*, error bars/.cd, y dir=both, y explicit] table[row sep=\\, x=f, y=mean, y error=err] {
f    mean  err \\
5    0.00  0.00 \\
10   0.00  0.00 \\
20   0.00  0.00 \\
30   0.00  0.00 \\
50   0.01  0.00 \\
};

\addplot+[violet, thick, solid, mark=x, error bars/.cd, y dir=both, y explicit] table[row sep=\\, x=f, y=mean, y error=err] {
f    mean  err \\
5    0.01  0.00 \\
10   0.01  0.00 \\
20   0.01  0.00 \\
30   0.01  0.00 \\
50   0.03  0.00 \\
};

\addplot+[blue, thick, solid, mark=*, error bars/.cd, y dir=both, y explicit] table[row sep=\\, x=f, y=mean, y error=err] {
f    mean  err \\
5    0.92  0.00 \\
10   0.60  0.00 \\
20   0.29  0.00 \\
30   0.17  0.00 \\
50   0.09  0.00 \\
};

\addplot+[red, thick, dashed, mark=square*, error bars/.cd, y dir=both, y explicit] table[row sep=\\, x=f, y=mean, y error=err] {
f    mean  err \\
5    0.11  0.00 \\
10   0.14  0.00 \\
20   0.10  0.00 \\
30   0.07  0.00 \\
50   0.17  0.00 \\
};

\addplot+[orange, thick, dashdotdotted, mark=diamond*, error bars/.cd, y dir=both, y explicit] table[row sep=\\, x=f, y=mean, y error=err] {
f    mean  err \\
5    0.56  0.00 \\
10   0.29  0.00 \\
20   0.10  0.00 \\
30   0.01  0.00 \\
50   0.01  0.00 \\
};

\addplot+[green!60!black, thick, dotted, mark=triangle*, error bars/.cd, y dir=both, y explicit] table[row sep=\\, x=f, y=mean, y error=err] {
f    mean  err \\
5    0.35  0.00 \\
10   0.33  0.00 \\
20   0.25  0.00 \\
30   0.23  0.00 \\
50   0.21  0.00 \\
};

\addplot+[black, thick, solid, mark=star, error bars/.cd, y dir=both, y explicit] table[row sep=\\, x=f, y=mean, y error=err] {
f    mean  err \\
5    0.76  0.00 \\
10   0.60  0.00 \\
20   0.38  0.00 \\
30   0.29  0.00 \\
50   0.17  0.00 \\
};

\end{axis}
\end{tikzpicture}
\end{minipage}



\begin{minipage}[c]{0.03\linewidth}
\centering
{\rotatebox{90}{\textbf{Out-Sample}}}
\end{minipage}
\hfill
\begin{minipage}[t]{0.30\linewidth}
\centering
\begin{tikzpicture}
\begin{axis}[
width=\linewidth, height=0.75\linewidth,
grid=both,
xlabel={$c,d$},
ylabel={$H$},
xmin=5, xmax=105,
ymin=0.0, ymax=1.0,
xtick={20,40,60,80,100},
ytick={0,0.2,0.4,0.6,0.8,1.0},
]

\addplot+[magenta, thick, solid, mark=pentagon*, error bars/.cd, y dir=both, y explicit] table[row sep=\\, x=c, y=mean, y error=err] {
c   mean  err \\
10  0.00  0.00 \\
20  0.00  0.00 \\
40  0.00  0.00 \\
60  0.00  0.00 \\
80  0.00  0.00 \\
100 0.00  0.00 \\
};

\addplot+[violet, thick, solid, mark=x, error bars/.cd, y dir=both, y explicit] table[row sep=\\, x=c, y=mean, y error=err] {
c   mean  err \\
10  0.13  0.00 \\
20  0.05  0.00 \\
40  0.04  0.00 \\
60  0.04  0.00 \\
80  0.04  0.00 \\
100 0.04  0.00 \\
};

\addplot+[blue, thick, solid, mark=*, error bars/.cd, y dir=both, y explicit] table[row sep=\\, x=c, y=mean, y error=err] {
c   mean  err \\
10  0.24  0.00 \\
20  0.26  0.00 \\
40  0.30  0.00 \\
60  0.35  0.00 \\
80  0.45  0.00 \\
100 0.55  0.00 \\
};

\addplot+[red, thick, dashed, mark=square*, error bars/.cd, y dir=both, y explicit] table[row sep=\\, x=c, y=mean, y error=err] {
c   mean  err \\
10  0.38  0.00 \\
20  0.25  0.00 \\
40  0.19  0.00 \\
60  0.18  0.00 \\
80  0.17  0.00 \\
100 0.17  0.00 \\
};

\addplot+[orange, thick, dashdotdotted, mark=diamond*, error bars/.cd, y dir=both, y explicit] table[row sep=\\, x=c, y=mean, y error=err] {
c   mean  err \\
10  0.49  0.00 \\
20  0.52  0.00 \\
40  0.57  0.00 \\
60  0.61  0.00 \\
80  0.52  0.00 \\
100 0.42  0.00 \\
};

\addplot+[green!60!black, thick, dotted, mark=triangle*, error bars/.cd, y dir=both, y explicit] table[row sep=\\, x=c, y=mean, y error=err] {
c   mean  err \\
10  0.16  0.00 \\
20  0.16  0.00 \\
40  0.17  0.00 \\
60  0.24  0.00 \\
80  0.42  0.00 \\
100 0.61  0.00 \\
};

\addplot+[black, thick, solid, mark=star, error bars/.cd, y dir=both, y explicit] table[row sep=\\, x=c, y=mean, y error=err] {
c   mean  err \\
10  0.65  0.00 \\
20  0.78  0.00 \\
40  0.86  0.00 \\
60  0.91  0.00 \\
80  0.97  0.00 \\
100 0.99  0.00 \\
};

\end{axis}
\end{tikzpicture}
\end{minipage}
\hfill
\begin{minipage}[t]{0.30\linewidth}
\centering
\begin{tikzpicture}
\begin{axis}[
width=\linewidth, height=0.75\linewidth,
grid=both,
xlabel={$\alpha$},
ylabel={$H$},
xmin=0.04, xmax=0.26,
ymin=0.0, ymax=1.0,
xtick={0.05,0.1,0.15,0.2,0.25},
ytick={0,0.2,0.4,0.6,0.8,1.0},
tick label style={/pgf/number format/fixed},
]

\addplot+[magenta, thick, solid, mark=pentagon*, error bars/.cd, y dir=both, y explicit] table[row sep=\\, x=a, y=mean, y error=err] {
a    mean  err \\
0.05 0.00  0.00 \\
0.10 0.00  0.00 \\
0.15 0.04  0.00 \\
0.20 0.19  0.00 \\
0.25 0.41  0.00 \\
};

\addplot+[violet, thick, solid, mark=x, error bars/.cd, y dir=both, y explicit] table[row sep=\\, x=a, y=mean, y error=err] {
a    mean  err \\
0.05 0.04  0.00 \\
0.10 0.1  0.00 \\
0.15 0.20  0.00 \\
0.20 0.33  0.00 \\
0.25 0.53  0.00 \\
};

\addplot+[blue, thick, solid, mark=*, error bars/.cd, y dir=both, y explicit] table[row sep=\\, x=a, y=mean, y error=err] {
a    mean  err \\
0.05 0.24  0.00 \\
0.10 0.50  0.00 \\
0.15 0.67  0.00 \\
0.20 0.76  0.00 \\
0.25 0.82  0.00 \\
};

\addplot+[red, thick, dashed, mark=square*, error bars/.cd, y dir=both, y explicit] table[row sep=\\, x=a, y=mean, y error=err] {
a    mean  err \\
0.05 0.17  0.00 \\
0.10 0.31  0.00 \\
0.15 0.49  0.00 \\
0.20 0.65  0.00 \\
0.25 0.75  0.00 \\
};

\addplot+[orange, thick, dashdotdotted, mark=diamond*, error bars/.cd, y dir=both, y explicit] table[row sep=\\, x=a, y=mean, y error=err] {
a    mean  err \\
0.05 0.49  0.00 \\
0.10 0.73  0.00 \\
0.15 0.72  0.00 \\
0.20 0.72  0.00 \\
0.25 0.71  0.00 \\
};

\addplot+[green!60!black, thick, dotted, mark=triangle*, error bars/.cd, y dir=both, y explicit] table[row sep=\\, x=a, y=mean, y error=err] {
a    mean  err \\
0.05 0.16  0.00 \\
0.10 0.40  0.00 \\
0.15 0.62  0.00 \\
0.20 0.77  0.00 \\
0.25 0.85  0.00 \\
};

\addplot+[black, thick, solid, mark=star, error bars/.cd, y dir=both, y explicit] table[row sep=\\, x=a, y=mean, y error=err] {
a    mean  err \\
0.05 0.65  0.00 \\
0.10 0.91  0.00 \\
0.15 0.97  0.00 \\
0.20 0.96  0.00 \\
0.25 0.95  0.00 \\
};

\end{axis}
\end{tikzpicture}
\end{minipage}
\hfill
\begin{minipage}[t]{0.30\linewidth}
\centering
\begin{tikzpicture}
\begin{axis}[
width=\linewidth, height=0.75\linewidth,
grid=both,
xlabel={$|\forgetset|$},
ylabel={$H$},
xmin=0, xmax=55,
ymin=0.0, ymax=1.0,
xtick={10,20,30,50},
ytick={0,0.2,0.4,0.6,0.8,1.0},
]

\addplot+[magenta, thick, solid, mark=pentagon*, error bars/.cd, y dir=both, y explicit] table[row sep=\\, x=f, y=mean, y error=err] {
f    mean  err \\
5    0.00  0.00 \\
10   0.00  0.00 \\
20   0.00  0.00 \\
30   0.00  0.00 \\
50   0.00  0.00 \\
};

\addplot+[violet, thick, solid, mark=x, error bars/.cd, y dir=both, y explicit] table[row sep=\\, x=f, y=mean, y error=err] {
f    mean  err \\
5    0.04  0.00 \\
10   0.05  0.00 \\
20   0.04  0.00 \\
30   0.04  0.00 \\
50   0.06  0.00 \\
};

\addplot+[blue, thick, solid, mark=*, error bars/.cd, y dir=both, y explicit] table[row sep=\\, x=f, y=mean, y error=err] {
f    mean  err \\
5    0.24  0.00 \\
10   0.37  0.00 \\
20   0.08  0.00 \\
30   0.04  0.00 \\
50   0.06  0.00 \\
};

\addplot+[red, thick, dashed, mark=square*, error bars/.cd, y dir=both, y explicit] table[row sep=\\, x=f, y=mean, y error=err] {
f    mean  err \\
5    0.17  0.00 \\
10   0.27  0.00 \\
20   0.22  0.00 \\
30   0.20  0.00 \\
50   0.17  0.00 \\
};

\addplot+[orange, thick, dashdotdotted, mark=diamond*, error bars/.cd, y dir=both, y explicit] table[row sep=\\, x=f, y=mean, y error=err] {
f    mean  err \\
5    0.49  0.00 \\
10   0.15  0.00 \\
20   0.00  0.00 \\
30   0.00  0.00 \\
50   0.00  0.00 \\
};

\addplot+[green!60!black, thick, dotted, mark=triangle*, error bars/.cd, y dir=both, y explicit] table[row sep=\\, x=f, y=mean, y error=err] {
f    mean  err \\
5    0.16  0.00 \\
10   0.15  0.00 \\
20   0.11  0.00 \\
30   0.09  0.00 \\
50   0.10  0.00 \\
};

\addplot+[black, thick, solid, mark=star, error bars/.cd, y dir=both, y explicit] table[row sep=\\, x=f, y=mean, y error=err] {
f    mean  err \\
5    0.65  0.00 \\
10   0.56  0.00 \\
20   0.37  0.00 \\
30   0.27  0.00 \\
50   0.18  0.00 \\
};

\end{axis}
\end{tikzpicture}
\end{minipage}

\begin{minipage}{\linewidth}\centering
\pgfplotslegendfromname{sharedlegend_cifar}
\end{minipage}

\caption{CIFAR100, RepVGG-a2: 5 classes forgetting. (Top): In-sample results. (Bottom): Out-sample results. (Left): $H$ vs. $c=d$. (Middle): $H$ vs. $\alpha$. (Right): $H$ vs. $|\forgetset|$.}
\label{fig:H_vs_c_alpha_cifar_class}

\end{figure*}



\begin{table*}[htbp] 
\caption{20NewsGroups \emph{class}-wise forgetting with $c=d=20$, $\alpha=0.05$, and 1 forgotten class. Coverage/miscoverage results.}
\label{tab:results_20newsgroups_class_f1_c20_a005_coverage}
\centering
\resizebox{\linewidth}{!}{%
\begin{tabular}{llccccccc}
\toprule
Split & Method
& $\ECF_{c}[\retainset]\uparrow$
& $\EmCF_{d}[\forgetset]\uparrow$
& $\ECF_{c}[\calT_r]\uparrow$
& $\EmCF_{d}[\calT_f]\uparrow$
& $\ECF_{c}[\calV_r]\uparrow$
& $\EmCF_{d}[\calV_f]\uparrow$
& $H\uparrow$ \\
\midrule

& OR
& $1.00 \pm 0.00$
& $0.00 \pm 0.00$
& $1.00 \pm 0.00$
& $0.00 \pm 0.00$
& $0.96 \pm 0.00$
& $0.05 \pm 0.00$
& $0.00 \pm 0.00$ \\

\midrule

\multirow{6}{*}{In}

& $\nabla\tau$
& \first{$1.00 \pm 0.00$}
& \second{$0.94 \pm 0.01$}
& \first{$1.00 \pm 0.00$}
& \second{$0.92 \pm 0.00$}
& \first{$1.00 \pm 0.00$}
& \first{$1.00 \pm 0.00$}
& \second{$0.98 \pm 0.00$} \\

& SCRUB
& \first{$1.00 \pm 0.00$}
& \second{$0.94 \pm 0.05$}
& \first{$1.00 \pm 0.00$}
& \third{$0.91 \pm 0.05$}
& \first{$1.00 \pm 0.00$}
& \third{$0.95 \pm 0.06$}
& \third{$0.96 \pm 0.02$} \\

& SSD
& $0.96 \pm 0.00$
& $0.00 \pm 0.00$
& $0.96 \pm 0.00$
& $0.00 \pm 0.00$
& $0.95 \pm 0.00$
& $0.00 \pm 0.00$
& $0.00 \pm 0.00$ \\

& AMN
& \first{$1.00 \pm 0.00$}
& $0.74 \pm 0.02$
& \first{$1.00 \pm 0.00$}
& $0.16 \pm 0.03$
& $0.97 \pm 0.00$
& $0.23 \pm 0.09$
& $0.40 \pm 0.05$ \\

& BADT
& \first{$1.00 \pm 0.00$}
& $0.00 \pm 0.00$
& \first{$1.00 \pm 0.00$}
& $0.00 \pm 0.00$
& $0.97 \pm 0.00$
& $0.07 \pm 0.02$
& $0.00 \pm 0.00$ \\

& EFFACE
& \first{$1.00 \pm 0.00$}
& \first{$0.97 \pm 0.01$}
& \first{$1.00 \pm 0.00$}
& \first{$0.95 \pm 0.01$}
& \first{$1.00 \pm 0.00$}
& \first{$1.00 \pm 0.00$}
& \first{$0.99 \pm 0.00$} \\

\midrule

\multirow{6}{*}{Out}

& $\nabla\tau$
& \first{$1.00 \pm 0.00$}
& \first{$0.95 \pm 0.00$}
& $1.00 \pm 0.00$
& \first{$0.96 \pm 0.00$}
& \first{$1.00 \pm 0.00$}
& \first{$1.00 \pm 0.00$}
& \first{$0.98 \pm 0.00$} \\

& SCRUB
& \first{$1.00 \pm 0.00$}
& $0.82 \pm 0.08$
& $1.00 \pm 0.00$
& \third{$0.82 \pm 0.08$}
& \second{$0.99 \pm 0.00$}
& \third{$0.88 \pm 0.06$}
& \third{$0.91 \pm 0.03$} \\

& SSD
& $0.95 \pm 0.00$
& $0.01 \pm 0.01$
& $1.00 \pm 0.00$
& $0.00 \pm 0.00$
& $0.96 \pm 0.00$
& $0.05 \pm 0.00$
& $0.00 \pm 0.00$ \\

& AMN
& \first{$1.00 \pm 0.00$}
& \third{$0.84 \pm 0.06$}
& $1.00 \pm 0.00$
& $0.15 \pm 0.02$
& $0.98 \pm 0.00$
& $0.33 \pm 0.09$
& $0.44 \pm 0.04$ \\

& BADT
& $0.94 \pm 0.00$
& $0.00 \pm 0.00$
& $1.00 \pm 0.00$
& $0.00 \pm 0.00$
& $0.97 \pm 0.00$
& $0.05 \pm 0.00$
& $0.00 \pm 0.00$ \\

& EFFACE
& \first{$1.00 \pm 0.00$}
& \first{$0.95 \pm 0.00$}
& $1.00 \pm 0.00$
& \first{$0.96 \pm 0.00$}
& \first{$1.00 \pm 0.00$}
& \first{$1.00 \pm 0.00$}
& \first{$0.98 \pm 0.00$} \\

\bottomrule
\end{tabular}%
}
\end{table*}


\begin{figure*}[htpb]
\centering

\begin{minipage}[c]{0.00\linewidth}
\hspace{\linewidth} 
\end{minipage}
\hfill
\begin{minipage}[t]{4cm}
\centering
\textbf{$\alpha=0.05,\; |\forgetset|=1$}
\end{minipage}
\hfill
\begin{minipage}[t]{4cm}
\centering
\textbf{$c=d=20,\; |\forgetset|=1$}
\end{minipage}
\hfill
\begin{minipage}[t]{4cm}
\centering
\textbf{$\alpha=0.05,\; c=d=20$}
\end{minipage}

\vspace{0.2cm}


\begin{minipage}[c]{0.03\linewidth}
\centering
{\rotatebox{90}{\textbf{In-Sample}}}
\end{minipage}
\hfill
\begin{minipage}[t]{0.30\linewidth}
\centering
\begin{tikzpicture}
\begin{axis}[
width=\linewidth, height=0.75\linewidth,
grid=both,
xlabel={$c,d$},
ylabel={$H$},
xmin=3, xmax=22,
ymin=0.0, ymax=1.0,
xtick={5,10,15,20},
ytick={0,0.2,0.4,0.6,0.8,1.0},
legend to name=sharedlegend_20newsgroups,
legend columns=4,
legend style={font=\small, /tikz/every even column/.append style={column sep=0.3cm}},
]

\addplot+[magenta, thick, solid, mark=pentagon*, error bars/.cd, y dir=both, y explicit] table[row sep=\\, x=c, y=mean, y error=err] {
c    mean  err \\
5    0.00  0.00 \\
10   0.00  0.00 \\
15   0.00  0.00 \\
20   0.00  0.00 \\
};

\addplot+[blue, thick, solid, mark=*, error bars/.cd, y dir=both, y explicit] table[row sep=\\, x=c, y=mean, y error=err] {
c   mean  err \\
5   0.00  0.00 \\
10  1.00  0.00 \\
15  1.00  0.00 \\
20  0.98  0.00 \\
};
\addlegendentry{$\nabla \tau$}

\addplot+[red, thick, dashed, mark=square*, error bars/.cd, y dir=both, y explicit] table[row sep=\\, x=c, y=mean, y error=err] {
c   mean  err \\
5   0.00  0.00 \\
10  0.00  0.00 \\
15  0.00  0.00 \\
20  0.96  0.00 \\
};
\addlegendentry{SCRUB}

\addplot+[orange, thick, dashdotdotted, mark=diamond*, error bars/.cd, y dir=both, y explicit] table[row sep=\\, x=c, y=mean, y error=err] {
c   mean  err \\
5   0.00  0.00 \\
10  0.00  0.00 \\
15  0.00  0.00 \\
20  0.00  0.00 \\
};
\addlegendentry{SSD}

\addplot+[green!60!black, thick, dotted, mark=triangle*, error bars/.cd, y dir=both, y explicit] table[row sep=\\, x=c, y=mean, y error=err] {
c   mean  err \\
5   0.89  0.00 \\
10  0.77  0.00 \\
15  0.77  0.00 \\
20  0.40  0.00 \\
};
\addlegendentry{AMN}

\addplot+[black, thick, solid, mark=star, error bars/.cd, y dir=both, y explicit] table[row sep=\\, x=c, y=mean, y error=err] {
c   mean  err \\
5   0.00  0.00 \\
10  0.71  0.00 \\
15  0.92  0.00 \\
20  0.99  0.00 \\
};
\addlegendentry{EFFACE}

\end{axis}
\end{tikzpicture}
\end{minipage}
\hfill
\begin{minipage}[t]{0.30\linewidth}
\centering
\begin{tikzpicture}
\begin{axis}[
width=\linewidth, height=0.75\linewidth,
grid=both,
xlabel={$\alpha$},
ylabel={$H$},
xmin=0.04, xmax=0.26,
ymin=0.0, ymax=1.0,
xtick={0.05,0.1,0.15,0.2,0.25},
ytick={0,0.2,0.4,0.6,0.8,1.0},
tick label style={/pgf/number format/fixed},
]

\addplot+[magenta, thick, solid, mark=pentagon*, error bars/.cd, y dir=both, y explicit] table[row sep=\\, x=a, y=mean, y error=err] {
a    mean  err \\
0.05 0.00  0.00 \\
0.10 0.00  0.00 \\
0.15 0.03  0.00 \\
0.20 0.29  0.00 \\
0.25 0.70  0.00 \\
};

\addplot+[blue, thick, solid, mark=*, error bars/.cd, y dir=both, y explicit] table[row sep=\\, x=a, y=mean, y error=err] {
a    mean  err \\
0.05 0.98  0.00 \\
0.10 0.98  0.00 \\
0.15 0.97  0.00 \\
0.20 0.97  0.00 \\
0.25 0.96  0.00 \\
};

\addplot+[red, thick, dashed, mark=square*, error bars/.cd, y dir=both, y explicit] table[row sep=\\, x=a, y=mean, y error=err] {
a    mean  err \\
0.05 0.96  0.00 \\
0.10 0.98  0.00 \\
0.15 0.96  0.00 \\
0.20 0.94  0.00 \\
0.25 0.91  0.00 \\
};

\addplot+[orange, thick, dashdotdotted, mark=diamond*, error bars/.cd, y dir=both, y explicit] table[row sep=\\, x=a, y=mean, y error=err] {
a    mean  err \\
0.05 0.00  0.00 \\
0.10 0.12  0.00 \\
0.15 0.60  0.00 \\
0.20 0.86  0.00 \\
0.25 0.90  0.00 \\
};

\addplot+[green!60!black, thick, dotted, mark=triangle*, error bars/.cd, y dir=both, y explicit] table[row sep=\\, x=a, y=mean, y error=err] {
a    mean  err \\
0.05 0.40  0.00 \\
0.10 0.71  0.00 \\
0.15 0.90  0.00 \\
0.20 0.95  0.00 \\
0.25 0.96  0.00 \\
};

\addplot+[black, thick, solid, mark=star, error bars/.cd, y dir=both, y explicit] table[row sep=\\, x=a, y=mean, y error=err] {
a    mean  err \\
0.05 0.99  0.00 \\
0.10 0.99  0.00 \\
0.15 0.98  0.00 \\
0.20 0.97  0.00 \\
0.25 0.95  0.00 \\
};

\end{axis}
\end{tikzpicture}
\end{minipage}
\hfill
\begin{minipage}[t]{0.30\linewidth}
\centering
\begin{tikzpicture}
\begin{axis}[
width=\linewidth, height=0.75\linewidth,
grid=both,
xlabel={$|\forgetset|$},
ylabel={$H$},
xmin=0, xmax=12,
ymin=0.0, ymax=1.0,
xtick={1,5,10},
ytick={0,0.2,0.4,0.6,0.8,1.0},
]

\addplot+[magenta, thick, solid, mark=pentagon*, error bars/.cd, y dir=both, y explicit] table[row sep=\\, x=f, y=mean, y error=err] {
f    mean  err \\
1    0.00  0.00 \\
5    0.00  0.00 \\
10   0.02  0.00 \\
};

\addplot+[blue, thick, solid, mark=*, error bars/.cd, y dir=both, y explicit] table[row sep=\\, x=f, y=mean, y error=err] {
f    mean  err \\
1    0.98  0.00 \\
5    0.12  0.00 \\
10   0.09  0.00 \\
};

\addplot+[red, thick, dashed, mark=square*, error bars/.cd, y dir=both, y explicit] table[row sep=\\, x=f, y=mean, y error=err] {
f    mean  err \\
1    0.96  0.00 \\
5    0.15  0.00 \\
10   0.15  0.00 \\
};

\addplot+[orange, thick, dashdotdotted, mark=diamond*, error bars/.cd, y dir=both, y explicit] table[row sep=\\, x=f, y=mean, y error=err] {
f    mean  err \\
1    0.00  0.00 \\
5    0.05  0.00 \\
10   0.00  0.00 \\
};

\addplot+[green!60!black, thick, dotted, mark=triangle*, error bars/.cd, y dir=both, y explicit] table[row sep=\\, x=f, y=mean, y error=err] {
f    mean  err \\
1    0.40  0.00 \\
5    0.16  0.00 \\
10   0.11  0.00 \\
};

\addplot+[black, thick, solid, mark=star, error bars/.cd, y dir=both, y explicit] table[row sep=\\, x=f, y=mean, y error=err] {
f    mean  err \\
1    0.99  0.00 \\
5    0.35  0.00 \\
10   0.16  0.00 \\
};

\end{axis}
\end{tikzpicture}
\end{minipage}



\begin{minipage}[c]{0.03\linewidth}
\centering
{\rotatebox{90}{\textbf{Out-Sample}}}
\end{minipage}
\hfill
\begin{minipage}[t]{0.30\linewidth}
\centering
\begin{tikzpicture}
\begin{axis}[
width=\linewidth, height=0.75\linewidth,
grid=both,
xlabel={$c,d$},
ylabel={$H$},
xmin=3, xmax=22,
ymin=0.0, ymax=1.0,
xtick={5,10,15,20},
ytick={0,0.2,0.4,0.6,0.8,1.0},
]

\addplot+[magenta, thick, solid, mark=pentagon*, error bars/.cd, y dir=both, y explicit] table[row sep=\\, x=c, y=mean, y error=err] {
c    mean  err \\
5    0.00  0.00 \\
10   0.00  0.00 \\
15   0.00  0.00 \\
20   0.00  0.00 \\
};

\addplot+[blue, thick, solid, mark=*, error bars/.cd, y dir=both, y explicit] table[row sep=\\, x=c, y=mean, y error=err] {
c   mean  err \\
5   0.00  0.00 \\
10  1.00  0.00 \\
15  1.00  0.00 \\
20  0.98  0.00 \\
};

\addplot+[red, thick, dashed, mark=square*, error bars/.cd, y dir=both, y explicit] table[row sep=\\, x=c, y=mean, y error=err] {
c   mean  err \\
5   0.00  0.00 \\
10  0.80  0.00 \\
15  0.76  0.00 \\
20  0.91  0.00 \\
};

\addplot+[orange, thick, dashdotdotted, mark=diamond*, error bars/.cd, y dir=both, y explicit] table[row sep=\\, x=c, y=mean, y error=err] {
c   mean  err \\
5   0.00  0.00 \\
10  0.00  0.00 \\
15  0.00  0.00 \\
20  0.00  0.00 \\
};

\addplot+[green!60!black, thick, dotted, mark=triangle*, error bars/.cd, y dir=both, y explicit] table[row sep=\\, x=c, y=mean, y error=err] {
c   mean  err \\
5   0.82  0.00 \\
10  0.71  0.00 \\
15  0.68  0.00 \\
20  0.44  0.00 \\
};

\addplot+[black, thick, solid, mark=star, error bars/.cd, y dir=both, y explicit] table[row sep=\\, x=c, y=mean, y error=err] {
c   mean  err \\
5   0.00  0.00 \\
10  0.92  0.00 \\
15  1.00  0.00 \\
20  0.98  0.00 \\
};

\end{axis}
\end{tikzpicture}
\end{minipage}
\hfill
\begin{minipage}[t]{0.30\linewidth}
\centering
\begin{tikzpicture}
\begin{axis}[
width=\linewidth, height=0.75\linewidth,
grid=both,
xlabel={$\alpha$},
ylabel={$H$},
xmin=0.04, xmax=0.26,
ymin=0.0, ymax=1.0,
xtick={0.05,0.1,0.15,0.2,0.25},
ytick={0,0.2,0.4,0.6,0.8,1.0},
tick label style={/pgf/number format/fixed},
]

\addplot+[magenta, thick, solid, mark=pentagon*, error bars/.cd, y dir=both, y explicit] table[row sep=\\, x=a, y=mean, y error=err] {
a     mean  err \\
0.05  0.00  0.00 \\
0.10  0.00  0.00 \\
0.15  0.00  0.00 \\
0.20  0.14  0.00 \\
0.25  0.27  0.00 \\
};

\addplot+[blue, thick, solid, mark=*, error bars/.cd, y dir=both, y explicit] table[row sep=\\, x=a, y=mean, y error=err] {
a    mean  err \\
0.05 0.98  0.00 \\
0.10 0.98  0.00 \\
0.15 0.98  0.00 \\
0.20 0.96  0.00 \\
0.25 0.94  0.00 \\
};

\addplot+[red, thick, dashed, mark=square*, error bars/.cd, y dir=both, y explicit] table[row sep=\\, x=a, y=mean, y error=err] {
a    mean  err \\
0.05 0.91  0.00 \\
0.10 0.99  0.00 \\
0.15 0.98  0.00 \\
0.20 0.96  0.00 \\
0.25 0.94  0.00 \\
};

\addplot+[orange, thick, dashdotdotted, mark=diamond*, error bars/.cd, y dir=both, y explicit] table[row sep=\\, x=a, y=mean, y error=err] {
a    mean  err \\
0.05 0.00  0.00 \\
0.10 0.00  0.00 \\
0.15 0.00  0.00 \\
0.20 0.00  0.00 \\
0.25 0.15  0.00 \\
};

\addplot+[green!60!black, thick, dotted, mark=triangle*, error bars/.cd, y dir=both, y explicit] table[row sep=\\, x=a, y=mean, y error=err] {
a    mean  err \\
0.05 0.44  0.00 \\
0.10 0.73  0.00 \\
0.15 0.89  0.00 \\
0.20 0.95  0.00 \\
0.25 0.95  0.00 \\
};

\addplot+[black, thick, solid, mark=star, error bars/.cd, y dir=both, y explicit] table[row sep=\\, x=a, y=mean, y error=err] {
a    mean  err \\
0.05 0.98  0.00 \\
0.10 0.98  0.00 \\
0.15 0.97  0.00 \\
0.20 0.95  0.00 \\
0.25 0.93  0.00 \\
};

\end{axis}
\end{tikzpicture}
\end{minipage}
\hfill
\begin{minipage}[t]{0.30\linewidth}
\centering
\begin{tikzpicture}
\begin{axis}[
width=\linewidth, height=0.75\linewidth,
grid=both,
xlabel={$|\forgetset|$},
ylabel={$H$},
xmin=0, xmax=12,
ymin=0.0, ymax=1.0,
xtick={1,5,10},
ytick={0,0.2,0.4,0.6,0.8,1.0},
]

\addplot+[magenta, thick, solid, mark=pentagon*, error bars/.cd, y dir=both, y explicit] table[row sep=\\, x=f, y=mean, y error=err] {
f    mean  err \\
1    0.00  0.00 \\
5    0.00  0.00 \\
10   0.05  0.00 \\
};

\addplot+[blue, thick, solid, mark=*, error bars/.cd, y dir=both, y explicit] table[row sep=\\, x=f, y=mean, y error=err] {
f    mean  err \\
1    0.98  0.00 \\
5    0.16  0.00 \\
10   0.09  0.00 \\
};

\addplot+[red, thick, dashed, mark=square*, error bars/.cd, y dir=both, y explicit] table[row sep=\\, x=f, y=mean, y error=err] {
f    mean  err \\
1    0.91  0.00 \\
5    0.35  0.00 \\
10   0.19  0.00 \\
};

\addplot+[orange, thick, dashdotdotted, mark=diamond*, error bars/.cd, y dir=both, y explicit] table[row sep=\\, x=f, y=mean, y error=err] {
f    mean  err \\
1    0.00  0.00 \\
5    0.00  0.00 \\
10   0.00  0.00 \\
};

\addplot+[green!60!black, thick, dotted, mark=triangle*, error bars/.cd, y dir=both, y explicit] table[row sep=\\, x=f, y=mean, y error=err] {
f    mean  err \\
1    0.44  0.00 \\
5    0.16  0.00 \\
10   0.14  0.00 \\
};

\addplot+[black, thick, solid, mark=star, error bars/.cd, y dir=both, y explicit] table[row sep=\\, x=f, y=mean, y error=err] {
f    mean  err \\
1    0.98  0.00 \\
5    0.28  0.00 \\
10   0.15  0.00 \\
};

\end{axis}
\end{tikzpicture}
\end{minipage}

\begin{minipage}{\linewidth}\centering
\pgfplotslegendfromname{sharedlegend_20newsgroups}
\end{minipage}

\caption{20NewsGroups: 1 class forgetting. (Top): In-sample results. (Bottom): Out-sample results. (Left): $H$ vs. $c=d$. (Middle): $H$ vs. $\alpha$. (Right): $H$ vs. $|\forgetset|$.}
\label{fig:H_vs_c_alpha_20newsgroups_class}

\end{figure*}


\subsection{Limitations of EFFACE}\label[Appendix]{app:limitation} 

We note the following limitations of EFFACE. 

\begin{itemize}[leftmargin=10pt,itemsep=2pt]
  \item Since EFFACE relies on solving an optimization problem over the unlearning set $\ulset$, its generalization cannot be guaranteed for data outside the unlearning set. When the unlearning is in-sample, we notice that the performance of EFFACE increases as it is optimized over a more representative dataset. 
  \item EFFACE uses a specific conformity scoring function for unlearning. In our experiments we used the softmax probability of the label as a scoring function. As described in the discussion under \cref{sec:different-scores}, if the downstream decision maker uses a conformity scoring function that is different in a strange manner from the one used in EFFACE, then EFFACE cannot be guaranteed to induce proper conformal unlearning.
\end{itemize}


\section{Further Sensitivity Analysis}\label[Appendix]{app:sensitivity} 

We present a sensitivity analysis in the following plots. We analyze the sensitivity of EFFACE to various hyperparameters in terms of retained-data coverage $\ECF_{\retainset}$ and $\ECF_{\calV_r}$ and accuracy $A_{\retainset}$ and $A_{\calV_r}$, the forgotten-data miscoverage $\EmCF_(\forgetset)$ and $\EmCF_{\calV_f}$ and accuracy $A_{\forgetset}$ and $A_{\calV_f}$, and the harmonic mean $H$. Experiments were conducted using 1 random seed.

\begin{figure}[htpb]
\centering

\begin{minipage}[t]{0.48\linewidth}
\centering
\begin{tikzpicture}
\begin{axis}[
  width=\linewidth, height=0.8\linewidth,
  grid=both,
  xlabel={$\rho$},
  ylabel={Metric},
  xmode=log,
  ymin=0, ymax=1.1,
  title={Sensitivity to $\rho$},
  legend to name=sharedlegend_news,
  legend columns=3,
  legend style={font=\small, /tikz/every even column/.append style={column sep=0.2cm}},
  tick label style={font=\tiny},
  label style={font=\footnotesize},
]
\addplot+[black, thick, solid, mark=star] coordinates {(0.0001,0.76) (0.001,0.76) (0.01,0.76) (0.1,0.76) (1,0.76)};
\addlegendentry{$H$}
\addplot+[red, thick, solid, mark=square*] coordinates {(0.0001,0.00) (0.001,0.00) (0.01,0.00) (0.1,0.00) (1,0.00)};
\addlegendentry{$A_{\forgetset}$}
\addplot+[blue, thick, solid, mark=*] coordinates {(0.0001,0.95) (0.001,0.95) (0.01,0.95) (0.1,0.95) (1,0.95)};
\addlegendentry{$A_{\retainset}$}
\addplot+[red, thick, dashed, mark=square] coordinates {(0.0001,0.00) (0.001,0.00) (0.01,0.00) (0.1,0.00) (1,0.00)};
\addlegendentry{$A_{\calV_f}$}
\addplot+[blue, thick, dashed, mark=o] coordinates {(0.0001,0.73) (0.001,0.73) (0.01,0.73) (0.1,0.73) (1,0.73)};
\addlegendentry{$A_{\calV_r}$}
\addplot+[orange, thick, solid, mark=diamond*] coordinates {(0.0001,0.83) (0.001,0.83) (0.01,0.83) (0.1,0.83) (1,0.83)};
\addlegendentry{$\EmCF_{\forgetset}$}
\addplot+[cyan, thick, solid, mark=triangle*] coordinates {(0.0001,0.99) (0.001,0.99) (0.01,0.99) (0.1,0.99) (1,0.99)};
\addlegendentry{$\ECF_{\retainset}$}
\addplot+[orange, thick, dashed, mark=diamond] coordinates {(0.0001,0.49) (0.001,0.49) (0.01,0.49) (0.1,0.49) (1,0.49)};
\addlegendentry{$\EmCF_{\calV_f}$}
\addplot+[cyan, thick, dashed, mark=triangle] coordinates {(0.0001,0.97) (0.001,0.97) (0.01,0.97) (0.1,0.97) (1,0.97)};
\addlegendentry{$\ECF_{\calV_r}$}
\end{axis}
\end{tikzpicture}
\end{minipage}
\hfill
\begin{minipage}[t]{0.48\linewidth}
\centering
\begin{tikzpicture}
\begin{axis}[
  width=\linewidth, height=0.8\linewidth,
  grid=both,
  xlabel={$\gamma$},
  xmode=log,
  ymin=0, ymax=1.1,
  title={Sensitivity to $\gamma$},
  tick label style={font=\tiny},
  label style={font=\footnotesize},
]
\addplot+[black, thick, solid, mark=star] coordinates {(0.0001,0.76) (0.001,0.76) (0.01,0.75) (0.1,0.67) (1.0,0.40)};
\addplot+[red, thick, solid, mark=square*] coordinates {(0.0001,0.00) (0.001,0.00) (0.01,0.00) (0.1,0.01) (1.0,0.06)};
\addplot+[blue, thick, solid, mark=*] coordinates {(0.0001,0.95) (0.001,0.95) (0.01,0.95) (0.1,0.95) (1.0,0.86)};
\addplot+[red, thick, dashed, mark=square] coordinates {(0.0001,0.00) (0.001,0.00) (0.01,0.00) (0.1,0.02) (1.0,0.24)};
\addplot+[blue, thick, dashed, mark=o] coordinates {(0.0001,0.73) (0.001,0.73) (0.01,0.74) (0.1,0.76) (1.0,0.75)};
\addplot+[orange, thick, solid, mark=diamond*] coordinates {(0.0001,0.83) (0.001,0.83) (0.01,0.83) (0.1,0.80) (1.0,0.39)};
\addplot+[cyan, thick, solid, mark=triangle*] coordinates {(0.0001,0.99) (0.001,0.99) (0.01,0.99) (0.1,0.99) (1.0,0.97)};
\addplot+[orange, thick, dashed, mark=diamond] coordinates {(0.0001,0.49) (0.001,0.49) (0.01,0.48) (0.1,0.38) (1.0,0.19)};
\addplot+[cyan, thick, dashed, mark=triangle] coordinates {(0.0001,0.97) (0.001,0.97) (0.01,0.97) (0.1,0.97) (1.0,0.96)};
\end{axis}
\end{tikzpicture}
\end{minipage}
\hfill

\vspace{0.3cm}

\begin{center}
\begin{minipage}[t]{0.48\linewidth}
\centering
\begin{tikzpicture}
\begin{axis}[
  width=\linewidth, height=0.8\linewidth,
  grid=both,
  xlabel={$\kappa$},
  xmode=log,
  ymin=0, ymax=1.1,
  title={Sensitivity to $\kappa$},
  tick label style={font=\tiny},
  label style={font=\footnotesize},
]
\addplot+[black, thick, solid, mark=star] coordinates {(0.1,0.26) (1.0,0.58) (5.0,0.76) (10.0,0.81) (20.0,0.85)};
\addplot+[red, thick, solid, mark=square*] coordinates {(0.1,0.19) (1.0,0.00) (5.0,0.00) (10.0,0.01) (20.0,0.00)};
\addplot+[blue, thick, solid, mark=*] coordinates {(0.1,0.78) (1.0,0.90) (5.0,0.95) (10.0,0.97) (20.0,0.97)};
\addplot+[red, thick, dashed, mark=square] coordinates {(0.1,0.42) (1.0,0.01) (5.0,0.00) (10.0,0.00) (20.0,0.00)};
\addplot+[blue, thick, dashed, mark=o] coordinates {(0.1,0.74) (1.0,0.76) (5.0,0.73) (10.0,0.72) (20.0,0.69)};
\addplot+[orange, thick, solid, mark=diamond*] coordinates {(0.1,0.18) (1.0,0.66) (5.0,0.83) (10.0,0.94) (20.0,0.98)};
\addplot+[cyan, thick, solid, mark=triangle*] coordinates {(0.1,0.96) (1.0,0.98) (5.0,0.99) (10.0,1.00) (20.0,1.00)};
\addplot+[orange, thick, dashed, mark=diamond] coordinates {(0.1,0.13) (1.0,0.30) (5.0,0.49) (10.0,0.55) (20.0,0.60)};
\addplot+[cyan, thick, dashed, mark=triangle] coordinates {(0.1,0.95) (1.0,0.96) (5.0,0.97) (10.0,0.98) (20.0,0.98)};
\end{axis}
\end{tikzpicture}
\end{minipage}
\hfill
\begin{minipage}[t]{0.48\linewidth}
\centering
\begin{tikzpicture}
\begin{axis}[
  width=\linewidth, height=0.8\linewidth,
  grid=both,
  xlabel={$c$},
  xmode=log,
  ymin=0, ymax=1.1,
  title={Sensitivity to $c$ ($d=100$)},
  tick label style={font=\tiny},
  label style={font=\footnotesize},
]
\addplot+[black, thick, solid, mark=star] coordinates {(1,0.76) (10,0.76) (30,0.76) (50,0.76) (100,0.76)};
\addplot+[red, thick, solid, mark=square*] coordinates {(1,0.00) (10,0.00) (30,0.00) (50,0.00) (100,0.00)};
\addplot+[blue, thick, solid, mark=*] coordinates {(1,0.95) (10,0.95) (30,0.95) (50,0.95) (100,0.95)};
\addplot+[red, thick, dashed, mark=square] coordinates {(1,0.00) (10,0.00) (30,0.00) (50,0.00) (100,0.00)};
\addplot+[blue, thick, dashed, mark=o] coordinates {(1,0.73) (10,0.73) (30,0.73) (50,0.73) (100,0.73)};
\addplot+[orange, thick, solid, mark=diamond*] coordinates {(1,0.83) (10,0.83) (30,0.83) (50,0.83) (100,0.83)};
\addplot+[cyan, thick, solid, mark=triangle*] coordinates {(1,0.99) (10,0.99) (30,0.99) (50,0.99) (100,0.99)};
\addplot+[orange, thick, dashed, mark=diamond] coordinates {(1,0.49) (10,0.49) (30,0.49) (50,0.49) (100,0.49)};
\addplot+[cyan, thick, dashed, mark=triangle] coordinates {(1,0.97) (10,0.97) (30,0.97) (50,0.97) (100,0.97)};
\end{axis}
\end{tikzpicture}
\end{minipage}
\end{center}

\vspace{0.2cm}
\begin{minipage}{\linewidth}
\centering
\pgfplotslegendfromname{sharedlegend_news}
\end{minipage}

\caption{\footnotesize Sensitivity analysis of EFFACE hyperparameters on CIFAR100, RepVGG-a2 (forgetting 5 classes). We vary one parameter at a time while keeping others fixed. Top row: $\rho$ and $\gamma$. Bottom row: $\kappa$, $c$ (with $d=100$).}
\label{fig:efface_sensitivity}

\end{figure}

In \cref{fig:efface_sensitivity} top left, we vary the condition penalty term $\rho$. In fact, we noticed that changing the set size conditions does not change the results whatsoever. Hence, changing $\rho$ also does not induce any change in the results of any metric. We maintain that in our experiments, the set size conditions were completely irrelevant unless they are chosen to be irreasonably small (e.g. $< 5$ for CIFAR100). However, we keep them in the formulation of EFFACE for generality as in some other applications/dataset the conditions might be more effective.

In \cref{fig:efface_sensitivity} top left, we vary unlearned-model parameters regularization constant $\gamma$. This term controls the penalty on the deviations of the parameters $\theta_u$ from the original parameters $\theta_o$. We cannot find a linear correlation between the metrics and the value of $\gamma$. However, we notice that choosing $\gamma$ close to 1 causes a clear drop in the performance of the model by decreasing the level of miscoverage frequency on the forgotten data. That is expected as choosing a large $\gamma$ value lowers the flexibility of EFFACE in updating the unlearned model's parameters away from the original parameters which makes the unlearning process less effective in achieving the desired coverage/miscoverage objectives. However, large $\gamma$ helps stabilize the unlearning process which results in higher accuracy on the validation subsets (less utility drop of the model).

In \cref{fig:efface_sensitivity} bottom left, we vary the steepness of the surrogate indicator. Recall that the ratio of covered/miscovered points is found using the indicator function on whether their true label is included or excluded from the prediction set. The indicator function is a step function that is non-differentiable. Hence, we use the sigmoid function with steepness $\kappa$ to approximate it. We notice that generally $\kappa=5$ is the optimal value. Decreasing $\kappa$ below 5 makes the surrogate so smooth to capture the decision boundaries created by the indicator function, while increasing $\kappa$ above 5 makes the surrogate so steep to generate smooth gradients for learning.
    
The last subfigure at the bottom right of \cref{fig:efface_sensitivity} shows the effect of varying the set size condition in EFFACE optimization problem, which we noted before was irrelevant in our experiments.


\subsection{Empirical Convergence and Feasibility}
\begin{figure}[htbp]
\centering
    \begin{subfigure}{0.48\textwidth}
        \centering
        \includegraphics[width=1.0\linewidth]{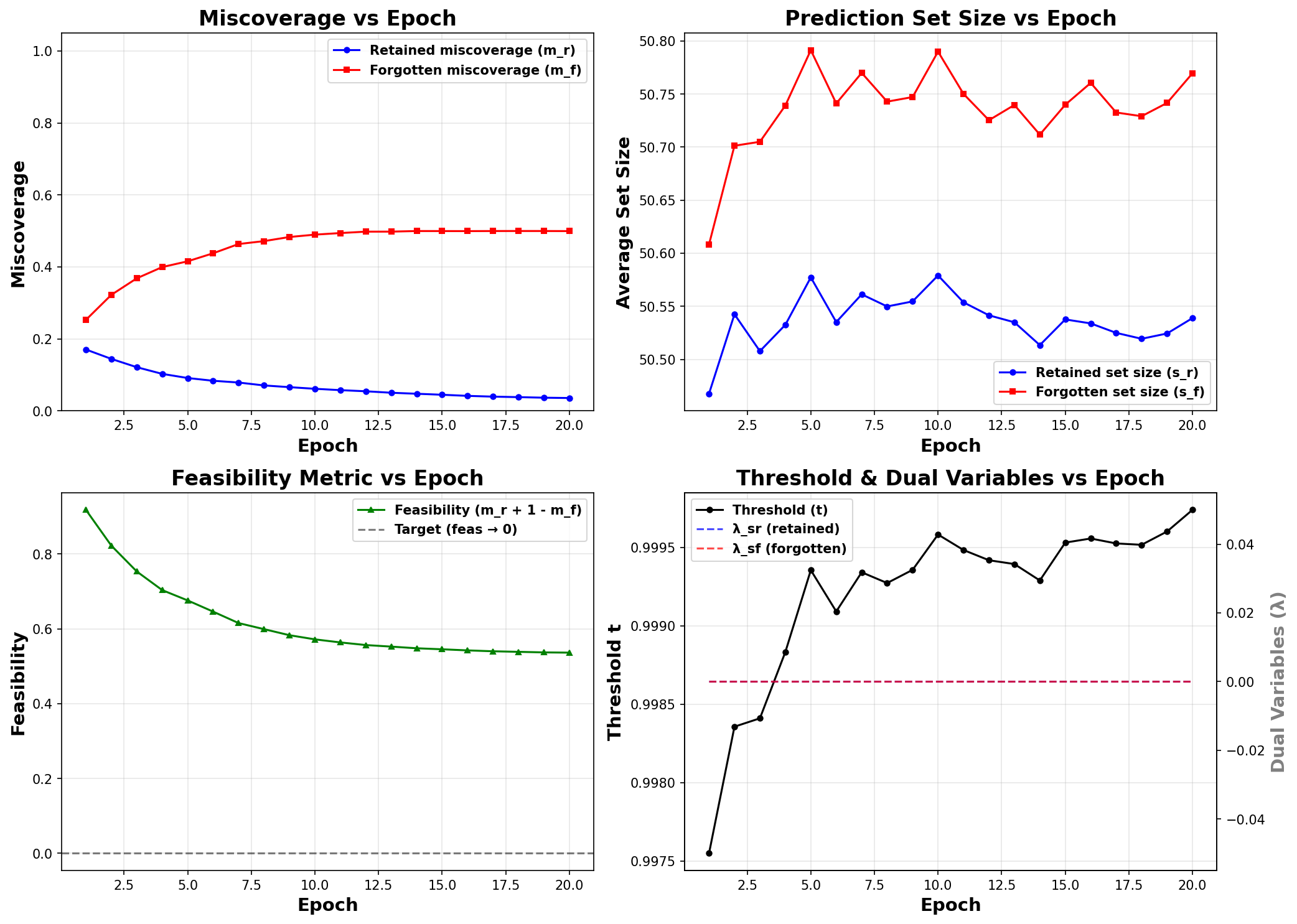}
        \caption{Plots of EFFACE objective convergence.}
        \label{fig:efface_convergance}
    \end{subfigure}
~
    \begin{subfigure}{0.48\textwidth}
         \centering
        \includegraphics[width=1.0\linewidth]{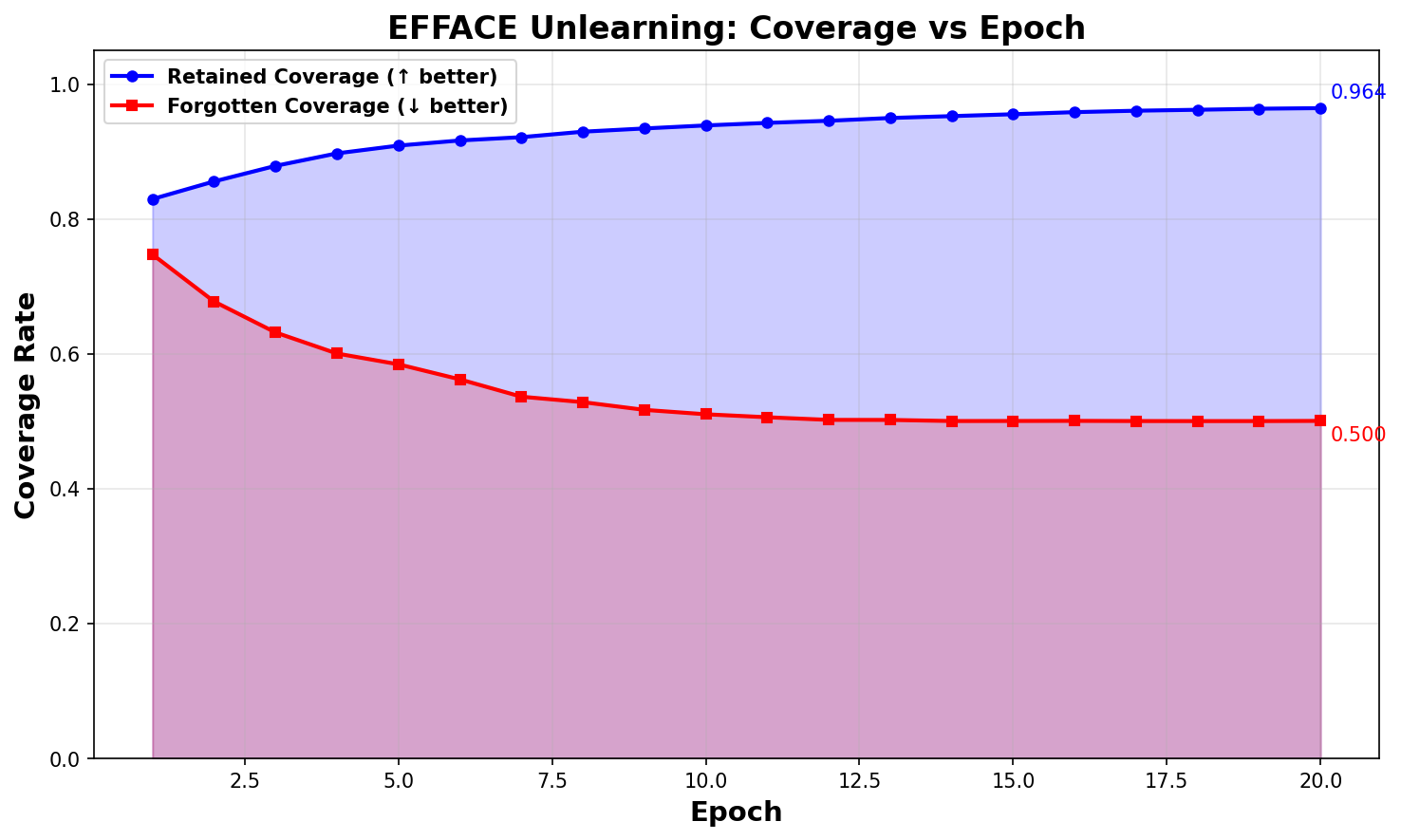}
        \caption{Plot of $1-\hat m _r$ and $1 - \hat m _r$ vs epochs.}
        \label{fig:efface_coverage_plot}
    \end{subfigure}
    \caption{EFFACE empirical convergence and feasibility on ImageNet100 forgetting 10 labels.}
    \label{fig:efface_convergence_and_feasibility}
\end{figure}

To better understand whether EFFACE shows some convergence, we next show how EFFACE's loss behaves against the number of epochs. The results are of ImageNet100 forgetting 10 labels. Notice in \cref{fig:efface_convergance} how the forgotten miscoverage folllows a fast increase at the beginning up to 7 epochs while the retained miscoverage decreases by the same rate. After that, both of miscoverage levels seem to converge to their optimal values. The miscoverage level over the retained set $\alpha$ and that over the forgotten set $\beta$ are related by the probability of the forgotten data. Here, since we are forgetting 10 labels out of 100, we can assume that the probability of getting a new point belonging to the forgotten labels is $\approx$ 0.1. Hence, the forgotten data miscoverage level can be at most 10 times that of the retained data. Since $\alpha=0.05$ in this case, we expect to see what is demonstrated in the figures with forget-data miscoverage $0.5$ and retained-data coverage $0.05$. The same convergence is shown by the feasibility plot which demonstrated the miscoverage level over the retained data minus that over the forgotten data plus 1. We target 0 feasibility when $\beta$ can go up to $1$ (in this case its bounded by $0.5$), meaning full miscoverage over the forgotten data and no miscoverage over the retained data. \Cref{fig:efface_coverage_plot} shows the same tendency but we replace miscoverage with coverage (flipping the plot upside down for easier reference). 

\subsection{Approximate Memory Requirement}
EFFACE requires approximately 4,116 MB of GPU memory when unlearning 10 classes from CIFAR100 on ResNet18, compared to 5,282 MB for $\nabla \tau$, 4,304 MB for SCRUB, 3,054 MB for AMN, 4,024 MB for UNSIR, and 4,032 MB for BADT. This suggests that
EFFACE seems not to use more memory compared to the average memory usage in the considered baselines.

\bibliographystyleSR{IEEEtran}
\bibliographySR{papers,books}

\end{document}